\documentclass[twoside,11pt]{article}

\usepackage{jmlr2e}

\usepackage[leqno]{amsmath}
\usepackage{MnSymbol}
\usepackage{amsfonts}
\usepackage{amsthm}
\usepackage{dsfont}
\usepackage{epsfig}
\usepackage{latexsym}
\usepackage{url}
\usepackage{hyperref}
\usepackage{ifthen}
\usepackage{comment}
\usepackage{algorithm}
\usepackage[noend]{algpseudocode}
\usepackage[shortlabels]{enumitem}
\usepackage{color}
\usepackage{caption}
\usepackage{subcaption}
\usepackage{graphicx}
\graphicspath{{./journal_media/}}

\newcommand{\field}[1]{\mathbb{#1}}
\newcommand{\N}{\field{N}} 
\newcommand{\R}{\field{R}} 
\newcommand{\tends}{{\rightarrow}} 
\newcommand{\E}{\mathbb{E}} 
\renewcommand{\Re}{\R} 

\newcommand{\Xc}{\mathcal{X}}
\newcommand{\Yc}{\mathcal{Y}}

\newcommand{\Pc}{\mathcal{P}}
\newcommand{\Qc}{\mathcal{Q}}
\newcommand{\Oc}{\mathcal{O}}

\newcommand{\Rc}{\mathcal{R}}
\newcommand{\Sc}{\mathcal{S}}
\newcommand{\Ac}{\mathcal{A}}
\newcommand{\Mc}{\mathcal{M}}

\newcommand{\Dc}{\mathcal{D}}

\newcommand{\Hc}{\mathcal{H}}
\newcommand{\Lc}{\mathcal{L}}
\newcommand{\Exp}{\mathds{E}}

\newcommand{\Nat}{\mathbb{N}}

\newcommand{\Alg}{{\rm alg}}
\newcommand{\ass}{\hspace{-1mm} = \hspace{-0.5mm} \mathbf{\cdot} \hspace{0.5mm}}

\newcommand\argmin{\mathop{\mbox{{\rm argmin}}}\limits}
\newcommand\argmax{\mathop{\mbox{{\rm argmax}}}\limits}

\newtheorem{theorem}{Theorem}

\newtheorem{corollary}{Corollary}
\newtheorem*{corollary*}{Corollary}
\newtheorem{lemma}{Lemma}
\newtheorem*{lemma*}{Lemma}
\newtheorem{assumption}{Assumption}
\newtheorem{definition}{Definition}

\newtheorem{remark}{Remark}
\newtheorem{example}{Example}

\numberwithin{equation}{section}
\newcommand{\state}{\mathcal{S}}

\newcommand{\mdp}{\mathcal{M}}

\newcommand{\hist}{\mathcal{H}_{\ell -1}}

\newcommand{\Ind}{\mathds{1}}

\newcommand{\Real}{\mathds{R}}

\errorcontextlines\maxdimen

\makeatletter
    \newcommand*{\algrule}[1][\algorithmicindent]{\makebox[#1][l]{\hspace*{.5em}\thealgruleextra\vrule height \thealgruleheight depth \thealgruledepth}}%
\newcommand*{\thealgruleextra}{}
\newcommand*{\thealgruleheight}{.75\baselineskip}
\newcommand*{\thealgruledepth}{.25\baselineskip}

\newcount\ALG@printindent@tempcnta
\def\ALG@printindent{%
    \ifnum \theALG@nested>0
        \ifx\ALG@text\ALG@x@notext
        \else
            \unskip
            \addvspace{-1pt}
            \ALG@printindent@tempcnta=1
            \loop
                \algrule[\csname ALG@ind@\the\ALG@printindent@tempcnta\endcsname]%
                \advance \ALG@printindent@tempcnta 1
            \ifnum \ALG@printindent@tempcnta<\numexpr\theALG@nested+1\relax
            \repeat
        \fi
    \fi
    }%
\usepackage{etoolbox}
\patchcmd{\ALG@doentity}{\noindent\hskip\ALG@tlm}{\ALG@printindent}{}{\errmessage{failed to patch}}
\makeatother

\newbox\statebox
\newcommand{\myState}[1]{%
    \setbox\statebox=\vbox{#1}%
    \edef\thealgruleheight{\dimexpr \the\ht\statebox+1pt\relax}%
    \edef\thealgruledepth{\dimexpr \the\dp\statebox+1pt\relax}%
    \ifdim\thealgruleheight<.75\baselineskip
        \def\thealgruleheight{\dimexpr .75\baselineskip+1pt\relax}%
    \fi
    \ifdim\thealgruledepth<.25\baselineskip
        \def\thealgruledepth{\dimexpr .25\baselineskip+1pt\relax}%
    \fi
    \State #1%
    \def\thealgruleheight{\dimexpr .75\baselineskip+1pt\relax}%
    \def\thealgruledepth{\dimexpr .25\baselineskip+1pt\relax}%
}

\usepackage{lastpage}
\jmlrheading{20}{2019}{1-\pageref{LastPage}}{5/18; Revised
8/19}{8/19}{18-339}{Ian Osband, Benjamin Van Roy, Daniel J. Russo and Zheng Wen}

\ShortHeadings{Deep Exploration via Randomized Value Functions}{Osband, Van Roy, Russo, Wen}
\firstpageno{1}

\begin{document}

\title{Deep Exploration via Randomized Value Functions}

\author{\name Ian Osband \email iosband@google.com \\
    \addr DeepMind 
\AND
   \name Benjamin Van Roy \email bvr@stanford.edu \\
   \addr Stanford University
\AND
   \name Daniel J. Russo \email djr2174@gsb.columbia.edu \\
   \addr Columbia University
\AND
   \name Zheng Wen \email zwen@adobe.com \\
   \addr Adobe Research
}

\editor{Peter Auer}
\maketitle


\begin{abstract}%
We study the use of randomized value functions to guide deep exploration in reinforcement learning.
This offers an elegant means for synthesizing statistically and computationally efficient exploration with common practical approaches to value function learning.
We present several reinforcement learning algorithms that leverage randomized value functions and demonstrate their efficacy through computational studies.
We also prove a regret bound that establishes statistical efficiency with a tabular representation.
\end{abstract}



\section{Introduction}

Reinforcement learning might provide the basis for an artificial intelligence
that can manage a wide range of systems and better serve the needs of society.
To date, its potential has primarily been assessed through learning in simulated systems,
where data generation is relatively unconstrained and algorithms are routinely trained
over millions to trillions of episodes.  Real systems, where data collection is
costly or constrained by the physical context, call for a focus on statistical efficiency.
A key driver of statistical efficiency is how the agent explores its environment.

The design of reinforcement learning algorithms that efficiently explore intractably large state spaces remains an important challenge.
Though a substantial body of work addresses efficient exploration, most of this focusses on tabular representations in which the number of parameters learned and the quantity of data required scale with the number of states.
Despite valuable insights that have been generated through design and analysis of tabular reinforcement learning algorithms, they are of limited practical import because, due to the curse of dimensionality, state spaces in most contexts of
practical interest are enormous.
There is a need for algorithms that generalize across states while exploring intelligently to learn to make effective decisions within a reasonable time frame.

In this paper, we develop a new approach to exploration that serves this need.
We build on value function learning, which underlies the most popular and successful approaches to reinforcement learning.
In common value function learning approaches, the agent maintains a point estimate of a function mapping state-action pairs to expected cumulative future reward.
This estimate typically takes a parameterized form, such as a linear combination of features or a neural network, with parameters fit to past observations.
The estimate approximates the agent's prevailing expectation of the true value function, and can be used to guide action selection.
As actions are applied and new observations gathered, parameters are adapted to fit the growing data set.
The hope is that this process quickly converges on a mode in which the agent selects near optimal actions and new observations reinforce prevailing value estimates.

In using the value function estimate to guide actions, the agent could operate according to a greedy policy, which at any given state, applies the action that maximizes estimated value.
However, such a policy does not investigate poorly-understood actions that are assigned
unattractive point estimates.
This can forgo enormous potential value; it is worthwhile to experiment with such an action
since the action could be optimal, and learning that can provide cumulating future benefit over subsequent visits to the state.
Thoughtful exploration can be critical to effective learning.

The simplest and most widely used approaches to exploration perturb greedy actions with random {\it dithering}.
An example is $\epsilon$-greedy exploration, which selects the greedy action with probability $1-\epsilon$ and otherwise selects uniformly at random from all currently available actions.
Dithering induces the experimentation required to learn about actions with unattractive point estimates.
However, such approaches waste much exploratory effort because they do not ``write-off'' actions that are known to be inferior.
This is because exploratory actions are selected without regard to the level of uncertainty associated with value estimates.
Clearly, it is only worth experimenting with an action that is expected to be undesirable if there is sufficient uncertainty surrounding that assessment.
As we will discuss further in Section \ref{se:deep}, this inefficiency can result in learning times that grow exponentially with the number of states.

A more sophisticated approach might only experiment with an action when applying the action will reveal useful information.
We refer to such approaches as {\it myopic}, since they do not account for
subsequent learning opportunities made possible by taking an action.
Though myopic approaches do ``write off'' actions where dithering approaches fail to, as we will discuss in Section \ref{se:deep}, myopic  exploration
can also require learning times that grow exponentially with the number of states or even entirely fail to learn.

Reliably efficient reinforcement learning calls for {\it deep exploration}.
By this we mean that the exploration method does not only consider immediate information gain but also the consequences of an action on future learning.
A deep exploration method could, for example, choose to incur losses over a sequence of actions while only expecting informative observations after multiple time periods.
Dithering and myopic approaches do not exhibit such strategic pursuit of information.

In this paper, we develop a new approach to deep exploration.
The idea is to apply actions that are greedy with respect to a randomly drawn statistically plausible value function.
Roughly speaking, we aim to sample from a proxy of the posterior distribution over value functions.
Such randomized value functions incentivize experimentation with actions of highly uncertain value, since this uncertainty translates into variance in the sampled value estimate.
This randomness often generates positive bias and therefore induces exploration.

There is much more to be said about the design of algorithms that leverage randomized value functions, and we cover some of this ground in Section \ref{se:algorithms}.
It is worth mentioning here, though, that this concept is abstract and broadly applicable, transcending specific algorithms.
Randomized value functions can be synthesized with the multitude of useful algorithmic ideas in the reinforcement learning literature to produce custom approaches for specific contexts.

To provide insight into the efficacy of randomized value functions, in Section \ref{sec: bounds}, we establish a strong bound on the Bayesian regret of a tabular algorithm.
This is not the first result to establish strong efficiency guarantees for tabular reinforcement learning.
However, previous algorithms that have been shown to satisfy similar regret bounds do not extend to contexts involving generalization via parameterized value functions.
In this regard, the approach we present is the first to satisfy a strong regret bound with tabular representations while also working effectively with the wide variety of practical value function learning methods that generalize over states and actions.

Section \ref{se:computation} presents computational results guided by randomized value functions that synthesize efficient exploration with generalization.
Experiments with a family of simple toy examples demonstrate dramatic efficiency gains relative to dithering approaches for exploration and that our randomized approaches are 
compatible with linearly parameterized generalizing value functions.
We also consider a cart-pole balancing problem that requires both deep exploration and generalization.  We address this problem through a combination of randomization
and deep learning, with value functions represented by neural networks.

\section{Literature review}

The Bayes-optimal policy serves as a gold standard for statistically efficient exploration in reinforcement learning (RL).
Given a prior distribution over Markov decision processes, one can formulate a problem to maximize expected cumulative reward by taking an action at each future time contingent on the prevailing posterior distribution.
A policy that attains this maximum is Bayes-optimal, and to do this it must explore judiciously.
Unfortunately, for problems of practical interest, computing a Bayes-optimal policy is intractable; the computational requirements grow exponentially in the problem parameters \citep{2010Szepesvari}.
We introduce an approach based on \textit{randomized value functions} that offers a computationally tractable approach to statistically efficient reinforcement learning.
Exploration via randomized value functions is not generally Bayes-optimal but, as we will argue, offers a practical approach to deep exploration,
which common exploration schemes fail to address, sometimes at enormous cost to statistical efficiency.

There is a substantial body of work on simultaneously computationally and statistically efficient exploration in tabular RL.
This begins with the seminal work of Kearns and Singh \citep{Kearns2002}, which identified the necessity of multi-period exploration strategies -- for which we adopt the term {\it deep exploration} -- to polynomial-time learning and established a polynomial-time learning guarantee for a particular tabular algorithm.
Subsequent papers proposed and analyzed alternative tabular algorithms that carry out deep exploration with varying degrees of efficacy \citep{Brafman2002,Auer2006,Strehl2006,Jaksch2010,Osband2013,Dann2015,azar2017minimax}.
None of these algorithms are Bayes-optimal, but they do bound the level of sub-optimality by some polynomial function of states and/or planning horizon.
By contrast, popular schemes such as $\epsilon$-greedy and Boltzmann exploration can require learning times that grow exponentially in the number of states and/or the planning horizon (see, e.g., \cite{Kakade2003,strehl2007probably}).
We discuss this phenomenon further in Section \ref{se:deep}.

The design and analysis of tabular algorithms has generated valuable insights, but the resultant algorithms are of little practical importance since, for practical problems the state space is typically enormous (due to the curse of dimensionality).
To learn effectively, practical RL algorithms must generalize across states to make effective decisions with limited data.
The literature offers a rich collection of such algorithms (e.g. \cite{Bertsekas1996,Sutton2018,2010Szepesvari,Powell2011} and references therein).
Though algorithms of this genre have achieved impressive outcomes, notably in games such as backgammon \citep{tesauro1995temporal}, Atari arcade games \citep{mnih2015human}, and go \citep{silver2016alphago,silver2017mastering}, they use naive exploration schemes that can be highly inefficient.
Possibly for this reason, these applications required enormous quantities of data.
In the case of \cite{silver2016alphago}, for example, neural networks were trained over hundreds of billions to trillions of simulated games.

The design of reinforcement learning algorithms that efficiently explore intractably large state spaces remains an important challenge.
Model learning algorithms exploit generalization in an underlying model of the environment \citep{Kearns1999,Abbasi-Yadkori2011,Ibrahimi2012,Ortner2012,osband2014model,osband2014near,gopalan2015thompson}.
However, these are typically restricted to simple model classes and become statistically or computationally intractable for problems of practical scale.
Policy learning algorithms, and the closely-related `evolutionary' algorithms identify high-performers among a set of policies \citep{Kakade2003,wierstra2008natural,deisenroth2013survey,plappert2017parameter}.
These algorithms can perform well, particularly when the space of possible optimal policies is parameterized to be small.
However, in a typical problem the space of policies is exponentially large; existing works either entail overly restrictive assumptions or do not make strong efficiency guarantees.

Value function learning has the potential to overcome computational challenges and offer practical means for synthesizing efficient exploration and effective generalization.
A relevant line of work establishes that efficient reinforcement learning with value function generalization reduces to efficient ``knows what it knows'' (KWIK) online regression \citep{Li2010,LiLW08}.
However, it is not known whether the KWIK online regression problem can be solved efficiently.
In terms of concrete algorithms, there is optimistic constraint propagation (OCP) \citep{WenVanroy13}, a provably efficient reinforcement learning algorithm for exploration and value function generalization in deterministic systems, and C-PACE \citep{pazis2013pac}, a provably efficient reinforcement learning algorithm that generalizes using interpolative representations.
These contributions represent important developments, but OCP is not suitable for stochastic systems and is highly sensitive to model misspecification, and generalizing effectively in high-dimensional state spaces calls for methods that extrapolate.

In this paper, we leverage randomized value functions to explore efficiently while generalizing via parameterized value functions.   
Algorithms of the kind we consider, which we refer to collectively as randomized least-squares value iteration (RLSVI), were first introduced in \cite{wen2014efficient}. 
Prior reinforcement learning algorithms that generalize via parameterized value functions require, 
in the worst case, learning times exponential in the number of model parameters and/or the planning horizon.
RLSVI aims to overcome these inefficiencies.  While RLSVI operates in a manner similar to well-known approaches such as least-squares value iteration (LSVI) and SARSA (see, e.g. \cite{Sutton2018}), what fundamentally distinguishes RLSVI is exploration through randomly sampling statistically plausible value functions.  Alternatives such as LSVI and SARSA are typically applied in conjunction with action-dithering schemes such as Boltzmann or $\epsilon$-greedy exploration, which lead to highly inefficient learning.

This paper aims to establish the use of \textit{randomized value functions} as a promising approach to tackling a critical challenge in reinforcement learning: synthesizing efficient exploration and effective generalization.
The only preceding work that advocates exploration through random samples of the value function comes from \cite{DeardenFR98}.
This paper proposes a tabular algorithm that resamples every timestep and so does not perform deep exploration.
A preliminary version of part of this work appeared in a short conference paper \citep{osband2016rlsvi}. While that paper proposed a specific algorithm that is compatible with linear function approximation, this paper develops the concept of deep exploration via randomized value functions in much greater depth and generality. We provide a general template for building algorithms that perform randomized value function learning and propose several specific instantiations of this idea. These algorithms are evaluated in entirely new simulations experiments, including an example in which neural networks are used for function approximation. While a proof sketch for a regret bound was given in \citep{osband2016rlsvi}, this paper gives a full and careful proof and develops new recursive stochastic dominance arguments that simplify the analysis. 
Following \citep{osband2016rlsvi}, several papers have proposed adaptations to neural network function approximation via bootstrap sampling \citep{osband2016deep}, linear final layer approximation \citep{azizzadenesheli2018efficient} or variational inference \citep{lipton2016efficient,fortunato2017noisy}.

The mathematical analysis we present in Section \ref{sec: bounds} establishes a bound on expected regret for a tabular version of RLSVI applied to an episodic finite-horizon problem, where the expectation is taken with respect to a particular uninformative distribution.
We view this result as a sanity check that, although it is designed for exploration with generalization, RLSVI recovers state-of-the-art efficiency guarantees in the simple tabular setting.
Our bound is $\tilde{O}(H \sqrt{S A H L})$, where $S$ and $A$ denote the cardinalities of the state and action spaces, $L$ denotes the number of episodes elapsed, and $H$ denotes the episode duration.
The lower bound of \cite{Jaksch2010} can be adapted to the episodic finite-horizon context to produce a $\Omega(H \sqrt{S A L})$ lower bound that applies to any algorithm.  This differs from our upper bound by a factor of $\sqrt{H}$, though this is not an apples-to-apples comparison, since the lower bound applies to a maximum over Markov decision processes and may not hold for the expectation over Markov decision processes, taken with respect to a prior distribution we posit. Follow up work by \cite{russo2019worst} shows that RLSVI also satisfies worst-case regret bounds in tabular environments.

In recent years there has been significant interest in alternative methods to incentivize exploration.
One popular method uses a density model or ``pseudocount'' to assign a bonus to states that have been visited infrequently \citep{bellemare2016count,tang2016explore}.
These methods can perform well, but only when the generalization of the density model is aligned with the task objective.
Crucially, this generalization is not learned from the task and, unlike the optimal value function, ``counts'' are generated by the agent's choices so there is no single target function to learn.
Further, these approaches add uncertainty bonus that is uncoupled across states, which can lead to a substantial negative impact on statistical efficiency, as discussed in \citep{osband2016posterior,o2017uncertainty}.

Exploration via randomized value functions is inspired by Thompson sampling \citep{Thompson1933,russo2017tutorial}.
In particular, when generating a randomized value function, the aim is to approximately sample from the posterior distribution of the optimal value function.
There are problems where Thompson sampling is in some sense near-optimal \citep{agrawal2012analysis,agrawal2012further,agrawal2013thompson,Russo2013b,Russo2014,gopalan2015thompson}.
Further, the theory suggests that ``well-designed'' upper-confidence-bound-based approaches, which appropriately couple uncertainties across state-action pairs, but are often computationally intractable, are similarly near-optimal (statistically) and competitive with Thompson sampling in such contexts \citep{Russo2013b,Russo2014}.
On the other hand, for some problems with more complex information structures, it is possible to explore much more efficiently than do Thompson sampling or upper-confidence-bound methods \citep{russo2014blearning}.
As such, for some RL problems and value function representations, the randomized value function approaches we put forth will leave substantial room for improvement. 

At a high level, randomized value functions replaces a point estimate of the value function by a distribution of plausible value functions.
Recently, another approach called ``distributional RL'' also suggests replacing a scalar value estimate by a distribution \citep{bellemare2017distributional}.
Although both might reasonably claim to offer a distributional perspective on reinforcement learning, the meaning and utility of the two distributions are quite distinct.
Randomized value functions aim to sample from a distribution that captures the Bayesian uncertainty in the unknown optimal value function; this concentrates around the true value function as more data is gathered.
By contrast, ``distributional RL'' fits a distribution to the realized value under stochastic outcomes.
For efficient exploration of \textit{unknown} rather than stochastic outcomes, it is important to use the correct notion of ``distributional RL''.


\section{Reinforcement learning problem}
\label{sec: rl_problem}

We consider a reinforcement learning problem in which an agent interacts with an unknown environment over a sequence of episodes.
We model the environment as a Markov decision process, identified by a tuple $\Mc = (\Sc, \Ac, \Rc, \Pc, \rho)$.
Here, $\Sc$ is a finite state space, $\Ac$ is a finite action space, $\Rc$ is a reward model, $\Pc$ is a transition model, and $\rho \in \Sc$ is an initial state distribution.
For each $s$, $\rho(s)$ is the probability that an episode begins in state $s$.
For any $s, s' \in \Sc$ and $a \in \Ac$, $\Rc_{s,a,s'}$ is a distribution over real numbers and $\Pc_{s,a}$ is a sub-distribution over states.
In particular, $\Pc_{s,a}(s')$ is the conditional probability that the state transitions to $s'$ from state $s$ and action $a$.
Similarly, $\Rc_{s,a,s'}(dr)$ is the conditional probability that the reward is in the set $dr$.
By {\it sub-distribution}, we mean that the sum can be less than one.
The difference $1 - \sum_{s' \in \Sc} \Pc_{s,a}(s')$ represents the probability that the process terminates upon transition.

We will denote by $r^\ell_t$, $s^\ell_t$, $a^\ell_t$ the state, action, and reward observed at the start of the $t$th time period of the $\ell$th episode.
In each $\ell$th episode, the agent begins in a random state $s^\ell_0 \sim \rho$ and selects an action $a^\ell_0 \in \Ac$.
Given this state-action pair, a reward and transition are generated according to $r^\ell_1 \sim \Rc_{s^\ell_0,a^\ell_0,s^\ell_1}$ and $s^\ell_1 \sim \Pc_{s^\ell_0,a^\ell_0}$.
The agent proceeds until termination, in each $t$th time period observing a state $s^\ell_t$, selecting an action $a^\ell_t$, and then observing a reward $r^\ell_{t+1}$ and transition to $s^\ell_{t+1}$.
Let $\tau_\ell$ denote the random time at which the process terminates, so that the sequence of observations made during episode $\ell$ is
$\Oc_\ell = \left(s_0^\ell, a_0^\ell, r_1^\ell, s_1^\ell, a_1^\ell, \ldots, s^\ell_{\tau_\ell-1}, a^\ell_{\tau_\ell - 1}, r^\ell_{\tau_\ell}\right)$.

We define a {\it policy} to be a mapping from $\Sc$ to a probability distribution over $\Ac$, and denote the set of all policies by $\Pi$.
We will denote by $\pi(a|s)$ the probability that $\pi$ assigns to action $a$ at state $s$.
Without loss of generality, we will consider states and actions to be integer indices, so that $\Sc = \{1,\ldots,|\Sc|\}$ and $\Ac=\{1,\ldots,|\Ac|\}$.
As such, we can define a substochastic matrix whose $(s,s')$th element is $\sum_{a \in \Ac} \pi(a|s) \Pc_{s,a}(s')$.
We make the following assumption to ensure finite episode duration:
\begin{assumption}
\label{as:pointwise-transient}
For all policies $\pi \in \Pi$, if each action $a_t$ is sampled from $\pi(\cdot |s_t)$, then the MDP $\Mc$ almost surely
terminates in finite time.
In other words, $\lim_{t \tends \infty} P_\pi^t = 0$,
where $P_\pi$ is the matrix whose $(s,s')$th element is $\sum_{a \in \Ac} \pi(a|s) \Pc_{s,a}(s')$.
\end{assumption}

For any MDP $\Mc$ and policy $\pi \in \Pi$, we define a value function $V^\pi_{\Mc}: \mathcal{S} \mapsto \Re$ by
$$V^\pi_{\Mc}(s) = \E_{\Mc, \pi}\left[\sum_{t=1}^\tau r_t \ \Big| \ s_0 = s\right],$$
where $r_t$, $s_t$, $a_t$, and $\tau$ denote rewards, states, actions, and termination time of a generic episode,
and the subscripts of the expectation indicate that actions are sampled according to $a_t \sim \pi(\cdot |s_t)$ and transitions and rewards are generated by the MDP $\Mc$.
Further, we define an optimal value function:
$$V^*_\Mc(s) = \max_{\pi \in \Pi} V^\pi_\Mc(s).$$

The agent's behavior is governed by a reinforcement learning algorithm $\Alg$.
Immediately prior to the beginning of episode $L$, the algorithm produces a policy $\pi^L = \Alg(\Sc, \Ac, \Hc_{L-1})$
based on the state and action spaces and the history $\Hc_{L-1} = \left(\Oc_\ell: \ell = 1,\ldots,L-1\right)$ of observations made over previous episodes.
Note that $\Alg$ may be a randomized algorithm, so that multiple applications of $\Alg$ may yield different policies.

In episode $\ell$, the agent enjoys a cumulative reward of $\sum_{t=1}^{\tau_\ell} r^\ell_t$.
We define the {\it regret} over episode $\ell$ to be the difference between optimal expected value and the expected value under algorithm $\Alg$.
This can be written as $\E_{\Mc, \Alg}\left[V^*(s^\ell_0) - V^{\pi^\ell}(s^\ell_0))\right]$, where the subscripts of the expectation indicate that each policy $\pi^\ell$ is produced by algorithm $\Alg$ and state transitions and rewards are generate by MDP $\Mc$.
Note that this expectation integrates over all initial states, actions, state transitions, rewards, and any randomness generated within $\Alg$, while the MDP $\Mc$ is fixed.
We denote \emph{cumulative regret}  over $L$ episodes by
$$\text{Regret}(\Mc, \Alg, L) = \sum_{\ell=1}^L \E_{\Mc, \Alg}\left[V^*(s_0^\ell) - V^{\pi^\ell}(s_0^\ell))\right].$$
We will generally refer to {\it cumulative regret} simply as {\it regret}.\

When used as a measure for comparing algorithms, one issue with regret is its dependence on $\Mc$.
One way of addressing this is to assume that $\Mc$ is constrained to a pre-defined set and to design algorithms with an aim of minimizing worst-case regret over this set.
This tends to yield algorithms that behave in an overly conservative manner when faced with representative MDPs.
An alternative is to aim at minimizing an average over representative MDPs.
The distribution over MDPs can be thought of as a prior, which captures beliefs of the algorithm designer.
In this spirit, we define {\it Bayesian regret}:
$$\text{BayesRegret}(\Alg, L) = \mathbb{E}\left[\text{Regret}(\Mc, \Alg, L)\right].$$
Here, the expectation integrates with respect to a prior distribution over MDPs.

It is easy to see that minimizing regret or Bayesian regret is equivalent to maximizing expected
cumulative reward.
These measures are useful alternatives to expected cumulative reward,
however, because for reasonable algorithms,
$\text{Regret}(\Mc, \Alg, L)/L$ and $\text{BayesRegret}(\Alg, L)/L$ should converge to zero.
When it is not feasible to apply an optimal algorithm, comparing how quickly these values diminish and how that depends on problem parameters can yield insight.

To denote our prior distribution over MDPs, as well as distributions over any other randomness that is realized, we will use a probability space $(\Omega, \mathbb{F}, \mathbb{P})$.
With this notation, the probability that $\Mc$ takes values in a set $\mathbb{M}$ is written as $\mathbb{P}(\Mc \in \mathbb{M})$.
In fact, the probability of any measurable event $\mathcal{E}$ is written as $\mathbb{P}(\mathcal{E})$.


\section{Deep exploration}
\label{se:deep}

Reinforcement learning calls for a sophisticated form of exploration that we refer to as {\it deep exploration}.
This form of exploration accounts not only for information gained upon taking an action but also for how the action may position the agent to more effectively acquire information over subsequent time periods.
We will use the following simple example to illustrate the critical role of deep exploration as well as how common approaches to exploration fall short on this front.

\begin{figure}[htpb]
\centering
\includegraphics[scale=0.5]{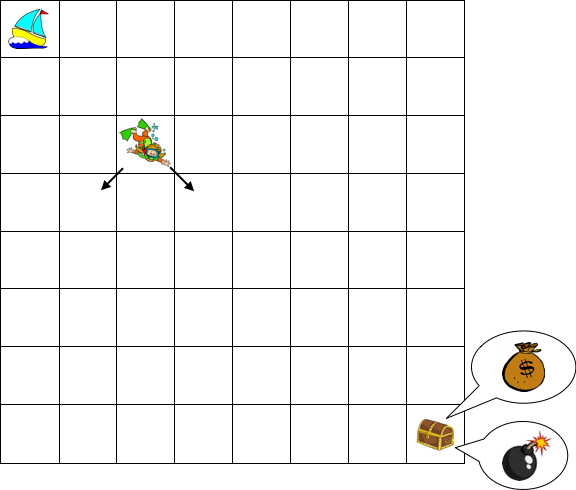}
\caption{Deep-sea exploration: a simple example where deep exploration is critical.}
\label{fig:grid}
\end{figure}

\begin{example}{\bf (Deep-sea exploration)}
\label{ex:grid}
\hspace{0.000001mm} \newline
Consider an MDP $\Mc = (\Sc, \Ac, \Rc, \Pc, \rho)$ with $|\Sc| = N^2$ states,
each of which can be thought of as a square cell in an $N\times N$ grid, as illustrated in Figure \ref{fig:grid}.
The action space is $\Ac = \{1,2\}$.
At each state, one of the actions represents ``left'' and the other
represents ``right,'' with the indexing possibly differing across states.
In other words, for a pair of
distinct states $s,s' \in \Sc$, action $1$ could represent ``left'' at state $s$ and ``right'' at state $s'$.
Any transition from any state in the lowest row leads to termination of the episode.
At any other state, the ``left'' action transitions to the cell immediately to the left, if possible, and below.
Analogously, the ``right'' action transitions to the cell immediately to the right, if possible, and below.
The agent begins every episode in the upper-left-most state (where her boat sits).
Note that, given the dynamics we have described, each episode lasts exactly $N$ time periods.

From any cell along the diagonal, there is a cost of $0.01/N$ incurred each time the ``right'' action is chosen.  No
cost is incurred for the left action.  The only other situation that leads to an additional reward or cost arises
when the agent is in the lower-right-most cell, where there is a chest.  There is an additional
reward of $1$ (treasure) or cost of $1$ (bomb) when the ``right'' action is selected at that cell.  Conditioned
on the $\Mc$, this reward is deterministic, so once the agent discovers whether there is treasure
or a bomb, she knows in subsequent episodes whether she wants to reach or avoid that cell.
In particular, given knowledge of $\Mc$, the optimal policy is to select the ``right'' action 
in every time period if there is treasure and, otherwise, to choose the ``left'' action in every time period.
Doing so accumulates a reward of $0.99$ if there is treasure and $0$ if there is a bomb.
It is interesting to note that a policy that randomly explores by selecting each action with equal 
probability is highly unlikely to reach the chest.
In particular, the probability such a policy reaches that cell in any given episode is $(1/2)^N$.
Hence, the expected number of episodes before observing the chest's content is $2^N$.
Even for a moderate value of $N=50$, this is over a quintillion episodes.

Let us now discuss the agent's beliefs, or state of knowledge, about the MDP $\Mc$, prior to the first episode.
The agent knows everything about $\Mc$ except:
\begin{itemize}
\item Action associations.
At each state, the agent does not know which action index is associated with ``right'' or ''left'', and assigns equal probability to either association.
These associations are independent across states.
\item Reward.
The agent does not know whether the chest contains treasure or a bomb and assigns equal probability to each of these possibilities.
\end{itemize}
Before learning action associations and rewards, the distribution over optimal value at the initial state is given by 
$\mathbb{P}(V^*_\Mc(s_0) = 0.99) = \mathbb{P}(V^*_\Mc(s_0) = 0) = 1/2$.
Because the MDP is deterministic, when an agent transitions from any state, she learns the action associations for that state, and
when the agent selects the ``right'' action at the lower-right-most state, she learns whether there is treasure or a bomb.
\end{example}

Note that the reinforcement learning problem presented in this example is easy to address.  In particular, it is straightforward to 
show that the minimal expected time to learn an optimal policy is achieved by an agent who chooses the ``right'' action whenever she knows which action that is, 
and otherwise, applies a random action, until she discovers the content of the chest, at which point she knows an optimal policy.
This algorithm identifies an optimal policy within $N$ episodes,  since in each episode, the agent learns how to move right from at least one additional cell along the diagonal.
Further, the expected learning time is $(N+1)/2$ episodes, since whenever at a state that has not previously been visited, the agent takes the wrong action with probability $1/2$.
Unfortunately, this algorithm is specialized to Example \ref{ex:grid} and does not extend to other reinforcement learning problems.
For our purposes, this example will serve as a sanity check and context for illustrating flaws and features of algorithms designed for the general reinforcement learning problem.

To facilitate our discussion, it is useful to define a couple of concepts.
The first is that of an {\it optimal state-action value function}, defined by $Q^*_\Mc(s,a) = \mathbb{E}_\Mc\left[r + V^*_\Mc(s')\right]$, where $r$ and $s'$ represent the reward and transition following application of action $a$ in state $s$.
Second, for any $Q:\Sc\times\Ac \mapsto \Re$, the {\it greedy policy} with respect to $Q$ selects an action that maximizes $Q$, sampling randomly among alternatives if there are multiple:
\begin{equation}
\label{eq:greedy_action}
    a \sim \mathtt{unif}\left(\argmax_{\alpha \in \Ac} Q(s,\alpha)\right).  
\end{equation}
Note that the greedy policy with respect to $Q^*_\Mc$ is optimal for the MDP $\Mc$.
This policy depends on the random MDP $\Mc$, and therefore can not be applied in the process of learning.

The first reinforcement learning algorithm we consider is {\bf pure-exploitation} and aims to maximize expected reward in the current episode, ignoring benefits of active exploration.
This algorithm estimates a ``best guess'' MDP $\hat{\Mc}_L$ based upon the data it has gathered up until episode $L$.  To offer a  representative approach, we will take $\hat{\Mc}_L$
to be the MDP with rewards and transition probabilities given by their expectations conditioned on the data.
The pure-exploitation algorithm then follows the policy that would be greedy with respect to
$\hat{Q}_L = Q^*_{\hat{\Mc}_L}$ during episode $L$.
While this algorithm is applicable to any reinforcement learning problem, its behavior in Example \ref{ex:grid} reveals severe inefficiencies.
Note that the algorithm is indifferent about finding the chest, since the expected reward associated with that is $0$.
Further, since moving toward the chest incurs cost, the algorithm avoids that, and therefore never visits the chest.
As such, the algorithm is unlikely to ever learn an optimal policy.

{\bf Dithering} approaches explore by selecting actions that randomly perturb what a pure-exploitation algorithm would do.
As an example, one form of dithering, known as {\it Boltzmann exploration} selects actions according to
\begin{equation}
\label{eq:boltzmann_action}
    a_t \sim \mathtt{multinomial} \left( \frac{\exp\left(\hat{Q}(s_t,\cdot) / \eta\right)}{\sum_{a \in \Ac} \exp\left(\hat{Q}(s_t,a) / \eta\right)} \right).
\end{equation}
Here, $\eta$ represents a ``temperature'' parameter.
As $\eta$ approaches zero, actions become the same as those that would be selected by a pure-exploitation algorithm.
As $\eta$ increases, the selection becomes noisier, eventually converging to a uniform distribution over actions.
In Example \ref{ex:grid}, a dithering algorithm is biased against moving toward the chest because of the associated cost.  
Only the random perturbations can lead the agent to the chest.
As such, the expected learning time is $\Theta(2^N)$.\footnote{A similar observation holds true for another popular dithering strategy: $\epsilon$-greedy. This approach selects random actions according to probability $\epsilon$ and greedy (pure-exploitation) otherwise.}

It is well known that dithering can be highly inefficient, even for bandit learning.
A key shortcoming is that dithering algorithms do not write-off bad actions.
In particular, even when observations make clear that a particular action is not worthwhile, dithering approaches can sample that action.
Despite this understanding, dithering is the most widely used exploration method in reinforcement learning.
The primary reason for this has been lack of computationally efficient approaches that adequately address the complex problems that arise in practical contexts.
This paper aims to fill that need.

Bandit learning can be thought of as a special case of reinforcement learning for which actions bear no delayed consequences.
The bandit learning literature offers sophisticated methods that overcome shortcomings
of dithering.
Such methods write-off bad actions, only selecting an action when it is expected to generate desirable reward or yield useful information or both.
A naive way of applying such an algorithm to a reinforcement learning problem involves selecting an action $a_t$ only if the expected value $\hat{Q}_L(s_t,a_t)$ is large or the observed reward and/or transition are expected to provide useful information.
We call this approach {\bf myopic exploration}, since it incentivizes exploration over a single timestep.
However, applying this approach to Example \ref{ex:grid} would once again avoid moving toward the chest as soon as it had learned the action associations in the initial state.
This is because there is a cost to moving right, but once the action associations are learned, there is no immediate benefit to applying the ``right'' action.
As such, myopic exploration is unlikely to ever learn an optimal policy.

Myopic exploration does not adequately address reinforcement learning because, in reinforcement learning, there is an additional motivation that should not be overlooked:
an action can be desirable even if expected to yield no value or immediate information if the action may place the agent in a state that leads to subsequent learning opportunities.
This is the essence of {\it deep exploration}; the agent needs to consider how actions influence downstream learning opportunities.
Viewed in another way, when considering how to explore, the agent should  probe {\it deep} in his decision tree.

{\bf Optimism} serves as another guiding principle in much of the bandit learning literature and can provide a basis for deep exploration as well.
In Example \ref{ex:grid}, if the agent takes most optimistic plausible view, it would assume that the chest offers treasure rather than a bomb,
so long as this hypothesis has not been invalidated.  In each $L$th episode, the agent follows a greedy policy with respect to 
a value function $Q_L$ that assigns to each state-action pair the maximal expected value under this assumption.  When at a cell 
along the diagonal of the grid, this policy selects the ``right'' action whenever the agent knows which that is.
Hence,  this optimistic algorithm learns the optimal policy within $N$ episodes.

The optimistic algorithm attains its strong performance in Example \ref{ex:grid} through carrying out deep exploration.
In particular, by assuming treasure rather than a bomb, the agent is incentivized to move right whenever it can, since that is the only way to obtain the posited treasure.
This exploration strategy is deep since the agent does not seek only immediate information but also a learning opportunity that will only arise 
after consecutively moving right over multiple time periods.

There are reasonably effective optimistic algorithms that apply to reinforcement learning problems with small (tractably enumerated) state and action spaces.
However, the design of such algorithms that adequately address reinforcement learning problems of practical scale in a computationally tractable manner
remains a challenge.

An alternative approach studied in the bandit learning literature involves randomly
sampled instead of optimistic estimates.
A focus of this paper is to extend this approach -- known as Thompson sampling -- to accommodate deep exploration in complex reinforcement learning problems.
Applied to Example \ref{ex:grid}, this {\bf randomized} approach would sample before each episode a random estimate $\tilde{Q}_L$ from the agent's posterior distribution over $Q^*_\Mc$, conditioned on observations made over previous episodes, or an approximation of this posterior distribution.
Before the agent's first visit to the chest, she assigns equal probability to treasure and a bomb, and therefore, the sample $\tilde{Q}_L$ has an equal chance of being optimistic or pessimistic.
The agent selects actions according to the greedy policy with respect to $\tilde{Q}_L$ and therefore on average explores over half of the episodes in a manner similar to an optimistic algorithm.
As such, the randomized algorithm can expect to learn the optimal policy within $2N$ episodes.

As applied to Example \ref{ex:grid}, there is no benefit to using a randomized rather than optimistic approach.
However, in the face of in complex reinforcement learning problems, the randomized approach can lead to computationally tractable algorithms 
that carry out deep exploration where the optimistic approach does not.

Table \ref{tab:grid} summarizes our discussion of learning times of various exploration methods applied to Example \ref{ex:grid}.
The minimal time required to learn an optimal policy, which is achieved by an agent who moves right whenever she knows how to, is $\Theta(N)$ episodes.
The pure-exploitation algorithm avoids {\it any} active exploration and requires
$\Theta(2^N)$ episodes to learn.
Dithering does not help for our problem.
Though more sophisticated, myopic approaches do not carry out deep exploration, and as such, still require $\Theta(2^N)$ episodes.
Optimistic and randomized approaches require only $\Theta(N)$ episodes.

\begin{table}[htpb]
\centering
 \begin{tabular}{|c|c|}
 \hline
 {\bf exploration method} & {\bf expected episodes to learn} \\
 \hline
 optimal & $\Theta(N)$ \\
 pure exploitation & $\infty$ \\
 myopic & $\infty$ \\
 dithering & $\Theta(2^N)$ \\
 optimistic & $\Theta(N)$ \\
 randomized & $\Theta(N)$ \\
 \hline
 \end{tabular}
 \caption{Expected number of episodes required to learn an optimal policy for Example \ref{ex:grid}.}
\label{tab:grid}
\end{table}


\section{Algorithms}
\label{se:algorithms}

The field of reinforcement learning has produced a substantial body of algorithmic ideas that serve as ingredients to mix, match, and customize in tailoring solutions to specific applications.
Such ideas are well-summarized in the textbooks of \citet{Bertsekas1996} and \citet{Sutton2018}, among others.
The aim of this paper is to contribute to this corpus a new approach to exploration based on randomized value functions, with the intention that this additional ingredient will broadly enable computationally efficient deep exploration.

Much of the literature and most notable applications build on value function learning.
This involves fitting a parameterized value function to observed data in order to estimate the optimal value function.
The algorithms we present will be of this genre.
As a starting point, in Section \ref{se:lsvi}, we will describe least-squares value iteration (LSVI), which is perhaps the simplest of value function learning algorithms.
In Section \ref{se:rlsvi}, we consider modifying LSVI by injecting randomness in a manner that incentivizes deep exploration.
This gives rise to a new class of algorithms, which we will refer to as randomized least-squares value iteration (RLSVI), and which offer computationally tractable means to deep exploration.

LSVI plays a foundational role in the sense that most popular value function learning algorithms can be interpreted as variations designed to improve computational efficiency or robustness to mis-specification of the parameterized value function.
The reinforcement learning literature presents many ideas that address such practical considerations.
In Section \ref{se:variations}, we will discuss how such ideas can be brought to bear in tandem with RLSVI.

\subsection{Value function learning}

Before diving into specific reinforcement learning algorithms, let us discuss general concepts that apply to all of them.
Value function learning algorithms make use of a family $\Qc$ of state-action value functions indexed by $\theta \in \Real^d$.
Each $\Qc_\theta: \Sc\times\Ac \mapsto \Re$ identifies a state-action value function.\footnote{We adopt notation that for all $\Qc$, parameter $\theta_0 = {\bf null}$ indicates $\Qc_{\theta_0} \equiv 0$.}
As a simple example of such a family, consider representing value functions as linear combinations of fixed features.
In particular, if $\phi(s,a) \in \Re^d$ is a vector of features designed to capture salient characteristics of the state-action pair $(s,a)$, it is natural to consider the family of
functions taking the form $\Qc_\theta(s,a) = \theta^\top \phi(s,a)$, with $\theta \in  \Re^d$.

Algorithm \ref{alg:live} ($\mathtt{live}$) provides a template for reinforcement learning algorithms we will consider.
It operates over an endless sequence of episodes, accumulating observations, learning value functions, and applying actions.
We use a Pythonic pseudocode, with an object-oriented division into $\mathtt{agent}$ and $\mathtt{environment}$.
We use
$$\mathtt{transition} =\mathtt{NamedTuple(old\_state, action, reward, new\_state, timestep)},$$
to describe the evolution of the system.
Where convenient, we will alternatively write $\mathtt{transition} = (s_t, a_t, r_t, s'_t, t)$.
We highlight three key methods the $\mathtt{agent}$ must implement:
\begin{itemize}[noitemsep]
    \item $\mathtt{act}$ -- select actions given its internal value estimates, (e.g. greedy action selection).

    \item $\mathtt{update\_buffer}$ -- incorporate observations to its memory buffer, (e.g. append to list).

    \item $\mathtt{learn\_from\_buffer}$ -- update value estimate given the data in the buffer, (e.g. LSVI).
\end{itemize}
The agents that we discuss will be distinguished through their implementation of these methods, which we will now outline.

\begin{algorithm}[!htpb]
\caption{$\mathtt{live}$}
\label{alg:live}
\begin{tabular}{lll}
\textbf{Input:} & \texttt{agent} & methods $\mathtt{act}, \mathtt{update\_buffer}, \mathtt{learn\_from\_buffer}$ \\
                & \texttt{environment} & methods $\mathtt{reset}, \mathtt{step}$ \\
\end{tabular}
\begin{algorithmic}[1]
\For{$\ell$ in $(1,2,\ldots)$}
\State \texttt{agent.learn\_from\_buffer()}
\State \texttt{transition} $\leftarrow$ \texttt{environment.reset()}
\While{\texttt{transition.new\_state} is not {\bf null}}
\State \texttt{action} $\leftarrow$ \texttt{agent.act(transition.new\_state)}
\State \texttt{transition} $\leftarrow$ \texttt{environment.step(action)}
\State \texttt{agent.update\_buffer(transition)}
\EndWhile
\EndFor
\end{algorithmic}
\end{algorithm}

The simplest form of $\mathtt{act}$ is given by the greedy strategy $\mathtt{act\_greedy}$.
An agent that uses this approach will select actions that maximize its estimated state-action value.
If multiple actions attain the maximum, one is sampled uniformly from among them.\footnote{We might also consider action selection via $\mathtt{act\_epsilon\_greedy}$ or $\mathtt{act\_boltzmann}$, as described in Section \ref{se:deep}. As we saw before, these dithering approaches do not perform \textit{deep exploration} and thus can lead to exponentially slower learning.}
Similarly, the simplest form of $\mathtt{update\_buffer}$ is to simply accumulate all observed data $\mathtt{update\_buffer\_queue}$.
Our next two sections will investigate agents that store all observed data and take greedy actions; we investigate the effects of $\mathtt{learn\_from\_buffer}$ and explain why training least-squares value iteration on randomly perturbed versions of the data can offer a computationally tractable means to deep exploration.

\subsection{Least-squares value iteration}
\label{se:lsvi}

Given an MDP $\Mc$, one can apply the value iteration algorithm (Algorithm \ref{alg:vi}) to compute an arbitrarily close approximation to $Q^*$.
The algorithm takes $\Mc$ and a planning horizon $H$ as input and computes $Q^*_H$, the optimal value over the next $H$ time periods of the episode as a function of the current state and action.
The computation is recursive: given $Q^*_h$, the algorithm computes $Q^*_{h+1}$ by taking the expected sum of immediate reward and $Q^*_h$, evaluated at the next state, maximized over actions.
Under Assumption \ref{as:pointwise-transient}, the mapping from $Q^*_h$ to $Q^*_{h+1}$ is a weighted-maximum-norm contraction mapping \citep{Bertsekas1996}, and as such, $Q^*_h$ converges to $Q^*$ at a geometric rate.
Hence, for any $\Mc$ satisfying Assumption \ref{as:pointwise-transient} and sufficiently large $H$, the greedy policy with respect to $Q^*_H$ is optimal.

\begin{algorithm}[H]
\caption{$\mathtt{vi}$}
\label{alg:vi}

\begin{tabular}{lll}
\textbf{Input:} & $\Mc  = (\Sc,\Ac,\Rc,\Pc,\rho)$ & MDP \\
& $H \in \Nat$ & planning horizon \\
\textbf{Output:} & $Q^*_H$ & optimal value function for $H$-period problem
\end{tabular}

\begin{algorithmic}[1]
\State $Q^*_0 \leftarrow 0$
\For{$h$ in $(0,\ldots,H-1)$}
 \State
 $Q^*_{h+1}(s,a) \leftarrow \sum_{s' \in \Sc} \Pc_{s,a}(s') \left(\int r \Rc_{s,a,s'}(dr) +  \max_{a' \in \Ac} Q^*_h(s',a') \right) \ \forall s,a \in \Sc \times \Ac$
\EndFor
\State {\bf return} $Q^*_H$
\end{algorithmic}
\end{algorithm}

Value iteration offers an idealized approach to evaluating $Q^*$, given knowledge of the underlying MDP $\Mc$ and the necessary computational power.
Least-squares value iteration (LSVI) adapts $\mathtt{vi}$ to an RL setting with imperfect statistical knowledge and limited computation.
For a value function family $\Qc = \{\Qc_\theta : \Sc \times \Ac \rightarrow \Real\}$, observed data $\Dc = \{(s_t,a_t,r_t,s_t', t)\}$ and target parameters $\theta^-$ we define the empirical temporal difference (TD) loss:
\begin{equation}
\label{eq:emp_bellman}
    \Lc(\theta ; \theta^-, \Dc) := \sum_{t \in \Dc} \left(r_t + \max_{a' \in \Ac} \Qc_{\theta^-}(s'_t, a') - \Qc_\theta(s_t, a_t) \right)^2.
\end{equation}
Note that, if $\Qc$ spans the true value function $Q^*$ and the data $\Dc$ matches the distribution of $\Mc$ then the minimizer of $\Lc$ matches the solution of Algorithm \ref{alg:vi}; for more information see \citet{Sutton2018}.

Algorithm \ref{alg:learn_lsvi} ($\mathtt{learn\_lsvi}$) describes the $\mathtt{learn\_from\_buffer}$ method for LSVI, whcih approximates the operations carried out by value iteration.
The algorithm successively minimizes the empirical temporal difference loss \eqref{eq:emp_bellman} plus a regularization term:
\begin{equation}
\label{eq:regularize}
    \Rc(\theta ; \theta^p) := \frac{v}{\lambda} \| \theta^p - \theta \|^2_2.
\end{equation}
Here $\theta^p$ can be interpreted as a prior for $\theta$ and $\frac{v}{\lambda}$ determines the strength of the regularization coefficient.
In a linear system these correspond to a prior belief $\theta \sim N(\theta^p, \lambda I)$ with observations $y_t = x_t \theta + z_t$ for $z_t \sim N(0, v)$.
Similarly to $\mathtt{vi}$, $\mathtt{learn\_lsvi}$ computes a sequence of value functions $(\Qc_{\theta_h}: h = 0,\ldots, H)$, reflecting
optimal expected rewards over an expanding horizon.
However, while value iteration computes optimal values using full knowledge of the MDP, LSVI produces estimates based only on observed data.
In each iteration, for each observed transition $(s,a,r,s')$, $\mathtt{learn\_lsvi}$ regresses the sum of immediate reward $r$ and the value
estimate $\max_{a' \in \Ac} \Qc_{\tilde{\theta}_h}(s',a')$ at the next state onto the value estimate $\Qc_{\tilde{\theta}_{h+1}}(s,a)$ for the current state-action pair.

\begin{algorithm}[H]
\caption{$\mathtt{learn\_lsvi}$}
\label{alg:learn_lsvi}

\begin{tabular}{lll}
\textbf{Agent:} & $\Lc(\theta \ass ; \theta^- \ass, \Dc \ass)$ & TD error loss function \\
& $\Rc(\theta \ass; \theta^p \ass)$ & regularization function \\
& $\mathtt{buffer}$ & memory buffer of observations \\
& $\mathtt{prior}$ & prior distribution of $\theta$ \\
& $H \in \Nat$ & planning horizon \\
\textbf{Updates:} & $\tilde{\theta}$ & agent value function estimate
\end{tabular}

\begin{algorithmic}[1]
\State $\tilde{\theta}_0 \leftarrow {\bf null}$
\State Data $\tilde{\Dc} \leftarrow \mathtt{buffer.data()}$
\State Prior parameter $\tilde{\theta}^p \leftarrow \mathtt{prior.mean()}$
\For{$h$ in $(0,\ldots,H-1)$}
\State $\tilde{\theta}_{h+1} \leftarrow \argmin_{\theta \in \Real^D}
    \left( \Lc(\theta ; \tilde{\theta}_h, \tilde{\Dc}) + \Rc(\theta; \tilde{\theta}^p) \right) $

\EndFor
\State update value function estimate $\tilde{\theta} \leftarrow \tilde{\theta}_H$
\end{algorithmic}
\end{algorithm}

In the event that the parameterized value function is flexible enough to represent every function mapping $\Sc\times\Ac$ to $\Re$, it is easy to see that, for any $\overline{\theta}$ and any positive $\lambda$ and $v$, as the observed history grows to include an increasing number of transitions from each state-action pair, value functions $\Qc_{\tilde{\theta}_H}$ produced by LSVI converge to $Q^*_H$.
However, in practical contexts, the data set is finite and the parameterization is chosen to be less flexible in order to enable generalization.
As such, $\Qc_{\tilde{\theta}_H}$ and $Q^*_H$ can differ greatly.

In addition to inducing generalization, a less flexible parameterization is critical for computational tractability.
In particular, the compute time and memory requirements of value iteration scale linearly with the number of states, which, due to the curse of dimensionality, grows intractably large in most practical contexts. LSVI sidesteps this scaling, instead requiring compute time and memory that scale polynomially with the dimension of the parameter vector $\tilde{\theta}$, the number of historical observations, and the time required to maximize over actions at any given state.

An LSVI $\mathtt{agent}$ may also be paired with some dithering strategy for exploration, such as $\epsilon$-greedy or Boltzmann exploration \eqref{eq:boltzmann_action} in place of $\mathtt{act\_greedy}$.
As discussed in Section \ref{se:deep}, randomly perturbing greedy actions -- or dithering -- does not achieve deep exploration and so can lead to exponentially poor performance.
Our next subsection introduces randomized value function estimates as an alternative.

\subsection{Randomized least-squares value iteration}
\label{se:rlsvi}

At a high level, the idea is to randomly sample an imagined optimal parameter $\tilde{\theta}$ according to the probability that it is optimal.
This approach is inspired by Thompson sampling, an algorithm widely used in bandit learning \citep{Thompson1933}.
In the context of a multi-armed bandit problem, Thompson sampling maintains a belief distribution over models that assign mean rewards to arms.
As observations accumulate, this belief distribution evolves according to Bayes rule.
When selecting an arm, the algorithm samples a model from this belief distribution and then selects the arm to which this model assigns largest mean reward.

To address a reinforcement learning problem, one could in principle apply Thompson sampling to value function learning.
This would involve maintaining a belief distribution over candidates for the optimal value function.
Before each episode, we would sample a function from this distribution and then apply the associated greedy policy over the course of the episode.
This approach could be effective if it were practically viable, but distributions over value functions are complex to represent and exact Bayesian inference would likely prove computationally intractable.

Randomized least-squares value iteration (RLSVI) is modeled after this Thompson sampling approach and serves as a computationally tractable method for sampling value functions.
RLSVI was first introduced in \cite{wen2014efficient}, and subsequent work has examined variations of RLSVI and their performance with both linear and nonlinear function approximation \citep{osband2016rlsvi,osband2016deep,osband2016}.
RLSVI does not explicitly maintain and update belief distributions and does not optimally synthesize information, as a coherent Bayesian method would.
Regardless, as we will later establish through computational and mathematical analyses, RLSVI can offer an effective approach to deep exploration.

\subsubsection{Randomization via Gaussian noise}
\label{se:grlsvi}

We first consider a version of RLSVI that induces exploration through injecting Gaussian noise into calculations of the form carried out by LSVI.
To understand the role of this noise, let us first consider a conventional linear regression problem.
Suppose we wish to estimate a parameter vector $\theta \in \Re^D$, with $N(\overline{\theta},\lambda I)$ prior and data $\Dc = ((x_n,y_n) : n=1,\ldots,N)$, where each ``feature vector'' $x_n$ is a row vector with $K$ components and each ``target value'' $y_n$ is scalar.
Given the parameter vector $\theta$ and feature vector $x_n$, the target $y_n$ is generated according to $y_n = x_n \theta + w_n$, where $w_n$ is independently drawn from $N(0,v)$.
Conditioned on $\Dc$, $\theta$ is Gaussian with
\begin{equation}
\label{eq:regression}
\E[\theta | \Dc] = \argmin_{\theta \in \Re^D} \left(\frac{1}{v} \sum_{n=1}^N (y_n - x_n \theta)^2 + \frac{1}{\lambda} \|\overline{\theta} - \theta\|^2\right)
= \left(\frac{1}{v} X^\top X + \frac{1}{\lambda} I\right)^{-1} \left(\frac{1}{v} X^\top y + \frac{1}{\lambda} \overline{\theta}\right)
\end{equation}
and
$${\rm Cov}[\theta | \Dc] = \left(\frac{1}{v} X^\top X + \frac{1}{\lambda} I\right)^{-1},$$
where $X \in \Re^{N \times D}$ is a matrix whose $n$th row is $x_n$ and $y \in \Re^n$ is a vector whose $n$th component is $y_n$.

One way of generating a random sample $\tilde{\theta}$ with the same conditional distribution as $\theta$ is simply to sample from $\tilde{\theta} \sim N(\E[\theta | \Dc], {\rm Cov}[\theta | \Dc])$.
An alternative construction is given by
\begin{equation}
\medmuskip=2mu
\thinmuskip=1mu
\thickmuskip=2mu
\label{eq:regress-sample}
\tilde{\theta} \leftarrow \argmin_{\theta \in \Re^D} \left(\frac{1}{v} \sum_{n=1}^N (y_n + z_n - x_n \theta)^2 + \frac{1}{\lambda} \|\hat{\theta} - \theta\|^2\right)
= \left(\frac{1}{v} X^\top X + \frac{1}{\lambda} I\right)^{-1} \left(\frac{1}{v} X^\top (y+z) + \frac{1}{\lambda} \hat{\theta}\right),
\end{equation}
where $\hat{\theta} \sim N(\overline{\theta}, \lambda I)$ and $z_n \sim N(0,v)$ are sampled independently.
To see why this $\tilde{\theta}$ takes on the same distribution, first note that $\tilde{\theta}$ is Gaussian, since it is affine in $\overline{\theta}$ and $z$.
Further, $\tilde{\theta}$ exhibits the appropriate mean and covariance matrix, since
$$\E[\tilde{\theta} | \Dc] = \left(\frac{1}{v} X^\top X + \frac{1}{\lambda} I\right)^{-1} \left(\frac{1}{v} X^\top (y+\E[z|\Dc]) + \frac{1}{\lambda} \E[\hat{\theta}|\Dc]\right) = \E[\theta | \Dc],$$
and
\begin{eqnarray*}
{\text Cov}[\tilde{\theta} | \Dc]
&=& \left(\frac{1}{v} X^\top X + \frac{1}{\lambda} I\right)^{-1} \left(\frac{1}{v^2} X^\top \E[z z^\top|\Dc] X + \frac{1}{\lambda^2} \E[\hat{\theta} \hat{\theta}^\top|\Dc] \right)
\left(\frac{1}{v} X^\top X + \frac{1}{\lambda} I\right)^{-1} \\
&=& \left(\frac{1}{v} X^\top X + \frac{1}{\lambda} I\right)^{-1} \left(\frac{1}{v} X^\top X + \frac{1}{\lambda} I \right) \left(\frac{1}{v} X^\top X + \frac{1}{\lambda} I\right)^{-1} \\
&=& {\rm Cov}[\theta | \Dc].
\end{eqnarray*}

Equation \eqref{eq:regress-sample} is signficant, since it allows us to understand Bayesian linear regression through a purely computational perspective.
For the linear setting, we see that training a least-squares solution on perturbed versions of the data is equivalent to conjugate Bayesian posterior.
This suggests that, in order to generate approximate posterior samples for $Q^*$ we can replace the least-square computation of Algorithm \ref{alg:learn_lsvi} with an alternative value iteration that trains on randomly perturbed versions of the data.
We call this algorithm \textit{randomized least-squares value iteration}, which we now outline.

\begin{algorithm}[H]
\caption{$\mathtt{learn\_rlsvi}$}
\label{alg:learn_rlsvi}

\begin{tabular}{lll}
\textbf{Agent:} & $\Lc(\theta \ass ; \theta^- \ass, \Dc \ass)$ & TD error loss function \\
& $\Rc(\theta \ass; \theta^p \ass)$ & regularization function \\
& $\mathtt{buffer}$ & memory buffer of observations \\
& $\mathtt{prior}$ & prior distribution of parameters \\
& $H \in \Nat$ & planning horizon \\
\textbf{Updates:} & $\tilde{\theta}$ & agent value function estimate
\end{tabular}

\begin{algorithmic}[1]
\State $\tilde{\theta}_0 \leftarrow {\bf null}$
\State Data $\tilde{\Dc} \leftarrow \mathtt{buffer.sample\_perturbed\_data()}$
\State Prior parameter $\tilde{\theta}^p \leftarrow \mathtt{prior.sample()}$
\For{$h$ in $(0,\ldots,H-1)$}
\State $\tilde{\theta}_{h+1} \leftarrow \argmin_{\theta \in \Real^D}
    \left( \Lc(\theta ; \tilde{\theta}_h, \tilde{\Dc}) + \Rc(\theta; \tilde{\theta}^p) \right) $

\EndFor
\State update value function estimate $\tilde{\theta} \leftarrow \tilde{\theta}_H$
\end{algorithmic}
\end{algorithm}

Note that $\mathtt{learn\_rlsvi}$ is identical to $\mathtt{learn\_lsvi}$ except that the optimization happens over randomized versions of the underlying data and prior.
For correspondence with the Gaussian derivation above we would implement:
{
\medmuskip=2mu
\thinmuskip=1mu
\thickmuskip=2mu
\begin{equation}
    \label{eq:gauss_rlsvi}
    \mathtt{buffer.sample\_perturbed\_data()}
    := [(s_t, a_t, r_t + z_t, s'_t, t) \ \forall t \in \mathtt{buffer}, \ z_t \sim N(0, v)].
\end{equation}
}
We call this version of RLSVI with additive Gaussian noise $\mathtt{learn\_grlsvi}$, indicating the specific choice of data randomization and prior $\theta \sim N(0, \lambda I)$.
In Section \ref{sec: bounds} we prove that this method recovers a polynomial regret bound when used with linear value functions and tabular representation.
This is significant, because LSVI with any dithering action selection scheme can not recover such a bound (Section \ref{se:deep}).
Before we jump to analysis, we provide some intuition for how this simple modification can lead to deep exploration.

\subsubsection{How does RLSVI drive deep exploration?}
\label{sec:how_deep}

To understand the role of injected noise and how this gives rise to deep exploration, let us discuss a simple example, involving an MDP $\Mc$ with four states
$\Sc = \{1,2,3,4\}$ and two actions $\Ac = \{\text{\it up}, \text{\it down}\}$.
Let $\Hc$ be a list of all transitions we have observed, and partition this into sublists
$\Hc_{s,a} = ((\tilde{s},\tilde{a},r,s') \in \Hc : (\tilde{s},\tilde{a}) = (s,a))$,
each containing transitions from a distinct state-action pair.
Suppose that $|\Hc_{(4,\text{\it down})}| = 1$, while for each state-action pair $(s,a) \neq (4,\text{\it down})$, $|\Hc_{s,a}|$ is virtually infinite.
Hence, we are highly uncertain about the expected immediate rewards and transition probabilities at $(4,\text{\it down})$ but can infer these quantities with extreme precision for every other state-action pair.

\begin{figure}[htpb]
\centering
\includegraphics[scale=0.5]{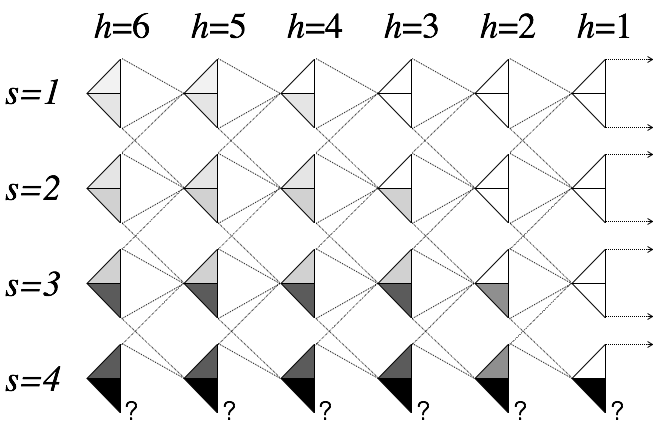}
\caption{Illustration of how $\mathtt{learn\_grlsvi}$ achieves deep exploration.}
\label{fig:rlsvi}
\end{figure}

Given our uncertainty about $\Mc$, $Q^*_H$ for each planning horizon $H$ is a random variable.
Figure \ref{fig:rlsvi} illustrates our uncertainty in these values.
Each larger triangle represents a pair $(s,h)$, where $h$ is the horizon index.  Note that these triangles represent
possible future states, and $h$ represents the number of periods between a visit to the state and the end of the planning horizon.
Each of these larger triangles is divided into two smaller ones, associated with {\it up} and {\it down} actions.
The dotted lines indicate plausible transitions, except at $(4,\text{\it down})$, where we are highly uncertain and any transition is plausible.
The shade of each smaller triangle represents our degree of uncertainty in the value of $Q^*_h(s,a)$.
To be more concrete, take our measure of uncertainty to be the variance of $Q^*_h(s,a)$.

For the case of $h=1$, only immediate rewards influence $Q^*_1$, and as such we are only uncertain about $Q^*_1(4,\text{\it down})$.
Stepping back to $h=2$, in addition to being highly uncertain about $Q^*_2(4,\text{\it down})$, we are somewhat uncertain about $Q^*_2(4,\text{\it up})$ and $Q^*_2(3,\text{\it down})$, since these pairs can transition to $(4,\text{\it down})$ and be exposed to the uncertainty associated with that state-action pair.
We are not as uncertain about $Q^*_2(4,\text{\it up})$ and $Q^*_2(3,\text{\it down})$ as we are about $Q^*_2(4,\text{\it down})$ because from $(4,\text{\it up})$ and $(3,\text{\it down})$, there is reasonable chance that we will never see $(4,\text{\it down})$.
Continuing to work our way leftward in the diagram, it is easy to visualize how uncertainty propagates as $h$ increases.

Let us now turn our attention to the variance of samples $\Qc_{\tilde{\theta}_H}(s,a)$ generated by $\mathtt{learn\_grlsvi}$, which, for reasons we will explain, tend to grow and shrink with the variance of $Q^*_H(s,a)$.
To keep things simple, assume $\lambda=\infty$ and that we use an exhaustive -- or ``tabular'' -- representation of value functions.
In particular, each component of the parameter vector $\theta \in \Re^{|\Sc|\times|\Ac|}$ encodes the value $\Qc_\theta(s,a)$ of a single state-action pair.
This parameterized value function can represent any mapping from $\Sc\times\Ac$ to $\Re$.

Under our simplifying assumptions, it is easy to show that
$$\Qc_{\tilde{\theta}_{h+1}}(s,a) = \frac{1}{|\tilde{\Hc}_{s,a}|} \sum_{(\tilde{s},\tilde{a},r,s',z) \in \tilde{\Hc}_{s,a}} \left(r + \max_{a' \in \Ac} \Qc_{\tilde{\theta}_h}(s',a') + z \right).$$
The right-hand-side is an average of target values.
Recall that, for any $(s,a) \neq (4,\text{\it down})$, $|\tilde{\Hc}_{s,a}|$ is so large that any sample average is extremely accurate, and therefore,
$\Qc_{\tilde{\theta}_{h+1}}(s,a)$ is essentially equal to $\E_\Mc[r_{t+1} + \max_{\alpha \in \Ac} \Qc_{\tilde{\theta}_h}(s_{t+1}, \alpha) | s_t=s,a_t=a]$.
For the distinguished case of $(4,\text{\it down})$, $|\tilde{\Hc}_{4,\text{\it down}}| = 1$, and the average target value may therefore differ substantially from its expectation
$\E[r + \max_{a' \in \Ac} \Qc_{\tilde{\theta}_h}(s',a') | \tilde{\theta}_h, \Mc]$.
Notably, the noise term $z$ does not average out as it does for other state-action pairs and should contribute variance $v$ to the sample $\Qc_{\tilde{\theta}_{h+1}}(4,\text{\it down})$.

Based on this observation, for the case of $h=1$, for $(s,a) \neq (4,\text{\it down})$, $\Qc_{\tilde{\theta}_1}(s,a)$ is virtually equal to $Q^*_1$, while for $\Qc_{\tilde{\theta}_1}(4,\text{\it down})$ exhibits variance of at least $v$.
For $h=2$, $\Qc_{\tilde{\theta}_2}(4,\text{\it down})$ again exhibits variance of at least $v$, but unlike the case of $h=1$, $\Qc_{\tilde{\theta}_2}(4,\text{\it up})$ and $\Qc_{\tilde{\theta}_2}(3,\text{\it down})$ also exhibit non-negligible variance since these pairs can transition to $(4,\text{\it down})$ and therefore depend on the noise-corrupted realization of $\Qc_{\tilde{\theta}_1}(4,\text{\it down})$.
Working leftward through Figure \ref{fig:rlsvi}, we can see that noise propagates and
influences value estimates in a manner captured by the shading in the figure.
Hence, samples $\Qc_{\tilde{\theta}_h}(s,a)$ exhibit high variance where the variance of $Q^*_h(s,a)$ is large.

This relationship drives deep exploration.
In particular, a high variance sample $\Qc_{\tilde{\theta}_H}(s,a)$ will be overly optimistic in some episodes.
Over such episodes, the agent will be incentivized to try the associated action.
This is appropriate because the agent is uncertain about the optimal value $Q^*_H(s,a)$ over the planning horizon.
Note that this incentive is not only driven by uncertainty concerning the immediate reward and transition.
As illustrated in Figure \ref{fig:rlsvi}, uncertainty propagates to offer incentives for the agent to pursue information even if it will require multiple time periods to arrive at an informative observation.
This is the essence of deep exploration.

It is worth commenting on a couple subtle properties of $\mathtt{learn\_grlsvi}$.
First, given $\theta_h$ and $\Hc$, $\theta_{h+1}$ is sampled from a Gaussian distribution.
However, given the inputs to $\mathtt{learn\_grslvi}$, the output $\tilde{\theta}$ is not Gaussian.
This is because at each step $\theta_{h+1}$ depends nonlinearly on $\theta_h$ due to the maximization over actions in the TD loss \eqref{eq:emp_bellman}.
Second, it is important that $\mathtt{learn\_grlsvi}$ uses the same noise samples $z$ in across iterations of the for loop in line 5.
To see why, suppose $\mathtt{learn\_grlsvi}$ used independent noise samples in each iteration.
Then, when applied to the example of Figure \ref{fig:rlsvi}, in some iterations, we would be optimistic about the reward at $(4,\text{\it down})$, while in other iterations, we would be pessimistic about that.
Now consider a sample $\Qc_{\tilde{\theta}_H}(1,{\it up})$ for large $H$.
This sample would be perturbed by a combination of optimistic and pessimistic noise terms influencing assessments at $(4,\text{\it down})$ to the right.
The resulting averaging effect could erode the chances that $\Qc_{\tilde{\theta}_H}(1,{\it up})$ is optimistic.

\subsubsection{Randomization via statistical bootstrap}
\label{se:bootstrap}

With $\mathtt{learn\_grlsvi}$, the Gaussian distribution of noise serves as a coarse model of errors between targets
$r + \max_{a' \in \Ac} \Qc_{\tilde{\theta}_h}(s',a')$ and expectations $\E[r + \max_{a' \in \Ac} \Qc_{\tilde{\theta}_h}(s',a') | \tilde{\theta}_h, \Mc]$.
The statistical bootstrap\footnote{We are overloading the term {\it bootstrap} here.
In reinforcement learning, bootstrapping is commonly used to refer to the calculation of a state-action value estimate based on value estimates at states to which
the agent may transition.
Here, we refer to the {\it statistical} bootstrap of data-based simulation.
The most common form of statistical bootstrap uses the sample data as an approximation to its generating distribution \citep{efron1982jackknife}.}
offers an alternative approach to randomization which may often more accurately capture characteristics of the generating process.
In its classic form, the bootstrap takes a dataset $\Dc$ of size $N$ and generates a sampled dataset $\tilde{\Dc}$ also of size $N$ drawn uniformly with replacement from $\Dc$ \citep{efron1994introduction}.
The bootstrap serves as a form of data-based simulation and, in certain cases, recovers strong convergence guarantees \citep{bickel1981some,fushiki2005bootstrap}.

$\mathtt{learn\_brlsvi}$ is a version of RLSVI that makes use of the bootstrap in place of additive Gaussian noise.
This algorithm follows $\mathtt{learn\_rlsvi}$ (Algorithm \ref{alg:learn_rlsvi}) and implements
{
\medmuskip=2mu
\thinmuskip=1mu
\thickmuskip=2mu
\begin{equation}
    \label{eq:boot_rlsvi}
    \mathtt{buffer.sample\_perturbed\_data()}
    := \mathtt{bootstrap\_sample(buffer.data())}.
\end{equation}
}
\hspace{-2mm} Bootstrap sampling for value function randomization may present several benefits over additive Gaussian noise. First, most bootstrap resampling schemes do not require a `noise variance' as input, which simplifies the algorithm from a user perspective. Related to this point, the bootstrap can effectively induce a state-dependent and heteroskedastic randomization which may be more appropriate in complex environments.
More generally, we can consider bootstrapped RLSVI as a non-parameteric randomization for the value function estimate and this opens a wide range of potential bootstrap variants and prior mechanisms that could be employed with RLSVI \citep{osband2015bootstrapped}. 

\cite{eckles2019bootstrap} were the first to propose using bootstrap samples as an approximation to the posterior samples used in the Thompson sampling algorithm. Unfortunately, bootstrapping does not provide meaningful uncertainty estimates in early periods. If applied without modification, the algorithms in \cite{eckles2019bootstrap} incur regret that scales linearly in the time horizon. Value function estimates in $\mathtt{learn\_brlsvi}$ are instead randomized not just through bootstrap sampling, but through regularizing toward a random prior sample. The effect of the random prior sample vanishes as diverse data is collected, but it is critical to driving exploration in early periods.

\subsection{Practical variants of RLSVI}
\label{se:variations}

In this section, we will present variants of RLSVI designed to address the important practical considerations of computational efficiency and robustness to mis-specification of the parameterized value function.
There are many ideas in the reinforcement learning literature that can be brought to bear for these purposes, and we will by no means cover an exhaustive list.
Rather, we will present a mix of ideas that lead to a particular algorithm that effectively addresses a broad range of complex problems.
This algorithm also serves to illustrate the many degrees of freedom in mixing and matching ingredients from the reinforcement learning literature when randomized value functions are part of the recipe.

\subsubsection{Finite buffer experience replay}
\label{sec: ep}

The use of a buffer of past observations to fit a value function is sometimes referred to as {\it experience replay}.  The algorithms we have presented so far use
an infinite buffer and thus require memory and compute time that grow linearly in the number of observations.
For complex problems that require substantial learning times, such a requirement becomes onerous.  To overcome this, we
can restrict the buffer to some finite size, treating it as a FIFO queue.

Computational requirements aside, there can be other substantial benefits to using a finite buffer.  In particular, the agent may learn to make more
effective decisions within fewer episodes \citep{adam2012experience}.  This is likely due to model mis-specification.  In particular,
if $Q^*$ can not be represented by $\Qc_\theta$ for any $\theta$, it is helpful to restrict attention to the most relevant data when
regressing, as this focusses on minimizing errors at relevant states and actions.
Restricting the buffer to recent observations may serve as a reasonable heuristic here.  Recent work has also demonstrated benefit from
more sophisticated prioritization of data for storage in a finite buffer \citep{schaul2015prioritized}.

\subsubsection{Discounted TD and incremental learning}
\label{sec: discount}

Both LSVI and RLSVI take a planning horizon $H \in \Nat$ as an argument.
These approaches can be wasteful in that they compute separate estimates $\tilde{\theta}_h$ for each $h=1,..,H$.
Further, these algorithms are ``batch,'' in the sense that they require computation over all observed data at the start of each episode; this leads to computational costs that grow with time.
In this subsection we introduce a discounted formulation that admits an incremental computational approach.

Let $\gamma \in (0,1)$ be a discount factor that induces a time preference over future rewards.
A discount $\gamma$ approximates an effective planning horizon $H \simeq \frac{1}{1 - \gamma}$, but affords a solution to the discounted Bellman equation $Q^*_\gamma(s,a) = \Exp_\Mc \left[ R(s, a) + \gamma \max_{a'}Q^*_\gamma(s', a') \right]$ \citep{blackwell1965discounted}.
Inspired by this relationship we define the $\gamma$-discounted empirical TD loss:
\begin{equation}
\label{eq:emp_bellman_gamma}
    \Lc \gamma(\theta ; \theta^-, \Dc) := \sum_{t \in \Dc} \left(r_t + \gamma \max_{a' \in \Ac} \Qc_{\theta^-}(s'_t, a') - \Qc_\theta(s_t, a_t) \right)^2.
\end{equation}
Algorithm \ref{alg:learn_online_lsvi} (\texttt{learn\_online\_lsvi}) presents an incremental variant of LSVI.
Rather than recompute $\tilde{\theta}$ from scratch each episode, $\mathtt{learn\_online\_lsvi}$ updates its previous estimate by gradient descent over a subset of the data.
This algorithm is a form of temporal difference learning \citep{sutton1988learning} and, at a high level, this approach broadly describes famous approaches such as TD-gammon \citep{tesauro1995temporal} and DQN \citep{mnih2015human}.
For more background and discussion of this family of algorithms we refer to \citet{Sutton2018}.

\begin{algorithm}[H]
\caption{$\mathtt{learn\_online\_lsvi}$}
\label{alg:learn_online_lsvi}

\begin{tabular}{lll}
\textbf{Agent:} & $\tilde{\theta}$ & parameter estimate \\
& $\tilde{\theta}^p$ & prior mean of parameter estimate \\
& $\Lc_\gamma(\theta \ass ; \theta^- \ass, \Dc \ass)$ & TD error loss function \\
& $\Rc(\theta \ass; \theta^p \ass)$ & regularization function \\
& $\mathtt{buffer}$ & memory buffer of observations \\
& $\alpha$ & learning rate \\
\textbf{Updates:} & $\tilde{\theta}$ & agent value function estimate
\end{tabular}

\begin{algorithmic}[1]
\State Data $\tilde{\Dc} \leftarrow \mathtt{buffer.sample\_minibatch()}$
\State $\delta \leftarrow \mathtt{buffer.minibatch\_size} \ /\  \mathtt{buffer.size}$
\State $\tilde{\theta} \leftarrow \tilde{\theta} - \alpha \nabla_{\theta \mid \theta=\tilde{\theta}} \left( \Lc_\gamma(\theta ; \tilde{\theta}, \tilde{\Dc}) + \delta \Rc(\theta; \tilde{\theta}^p) \right)$
\end{algorithmic}
\end{algorithm}

\subsubsection{Randomization via ensemble sampling}
\label{sec: parallel_rlsvi}

Algorithm \ref{alg:learn_online_lsvi} (\texttt{learn\_online\_lsvi}) presents an incremental version of LSVI.
However, RLSVI in its purest form calls for an estimate trained on randomly perturbed data at the beginning of each episode $l=1,2,..$; this is not immediately amenable to such an incremental algorithm.
Instead, our solution approximates the effects of RLSVI in an online algorithm via \textit{ensemble sampling}.

\begin{figure}[!h]
\centering
\begin{subfigure}{.5\textwidth}
  \centering
  \includegraphics[width=.95\linewidth]{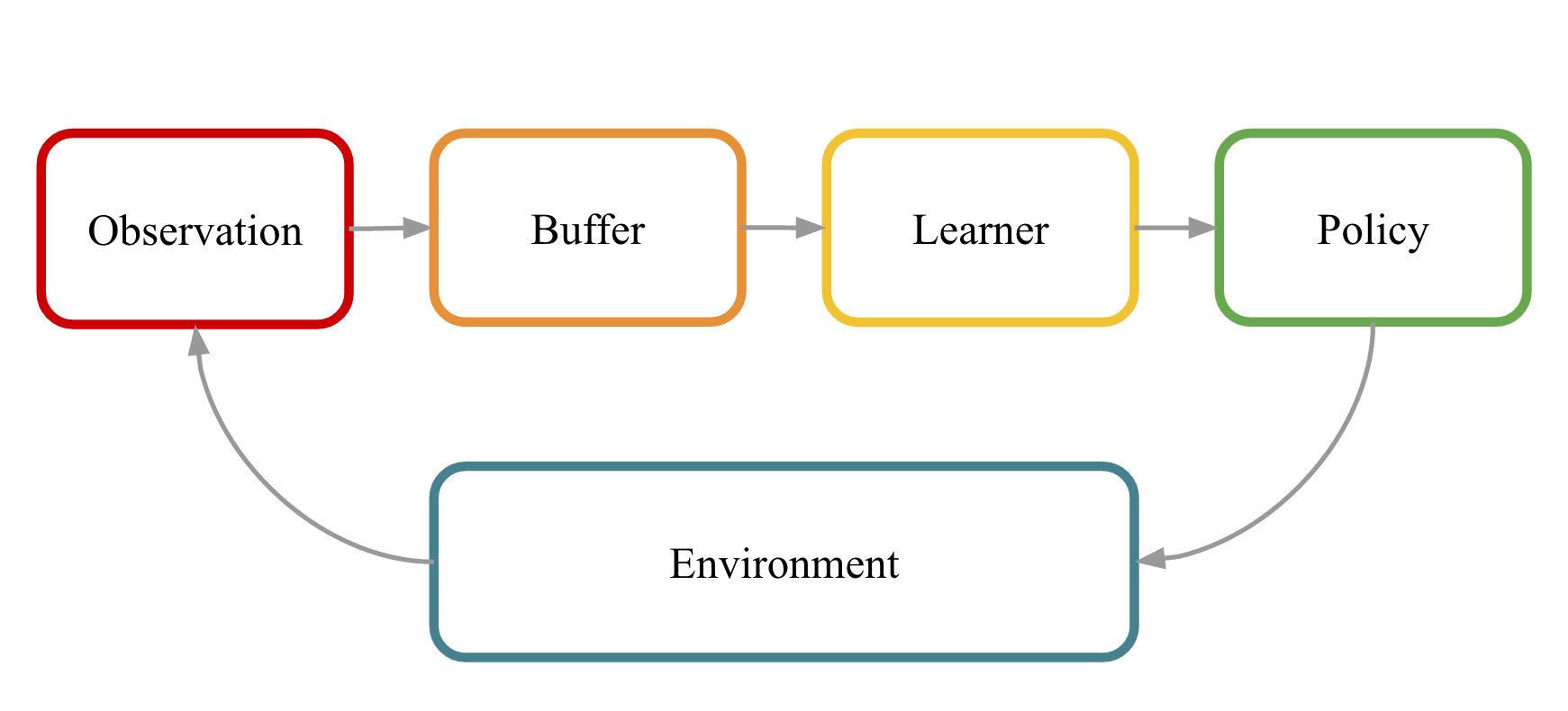}
  \vspace{-2mm}
  \caption{learning a single value function}
  \label{fig:online_lsvi}
\end{subfigure}%
\begin{subfigure}{.5\textwidth}
  \centering
  \includegraphics[width=.95\linewidth]{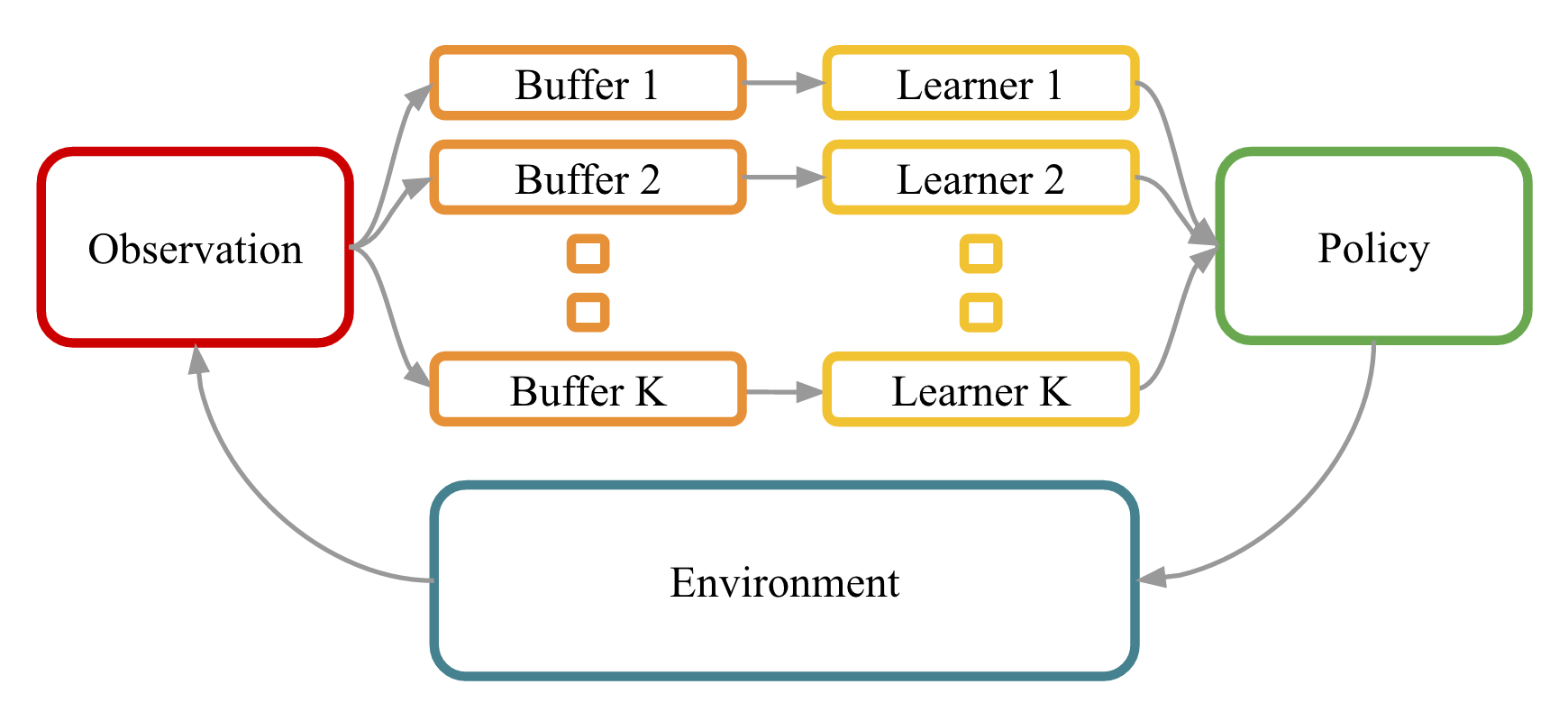}
  \vspace{-2mm}
  \caption{learning multiple value functions in parallel}
  \label{fig:online_rlsvi}
\end{subfigure}
\caption{\small RLSVI via ensemble sampling, each member produced by LSVI on perturbed data.}
\label{fig:online_models}
\end{figure}

Ensemble sampling approximates the distribution induced by RLSVI through an ensemble of $K \in \Nat$ estimates trained in parallel $(\tilde{\theta}_1, .., \tilde{\theta}_K)$ \citep{osband2015bootstrapped,NIPS2017_6918}.
At the start of any episode $l$, we can approximate the distribution of $\tilde{\theta}$ under RLSVI via a sample of $\tilde{\theta}_k$ for $k \sim {\rm Unif}(1,..,K)$.
Figure \ref{fig:online_models} presents an illustration of such a system.

One method to maintain $K$ estimates in parallel is to implement an ensemble memory buffer $\mathtt{ensemble\_buffer}$.
An ensemble buffer should function similarly to $\mathtt{buffer}$, but maintain $K$ distinct perturbations of the observed data $\tilde{\Dc}_k$, each associated with their appropriate ensemble value estimate $\tilde{\theta}_k$.
We use a Pythonic notation that $\mathtt{ensemble\_buffer[k]}$ is a $\mathtt{buffer}$ for each $\mathtt{k=1,..,K}$; although in a practical implementation it will be sensible to share appropriate parts of the memory requirements.
For an online variant of $\mathtt{learn\_grlsvi}$ (additive Gaussian noise) we can store $K$ distinct samples of additive noise, as described in Algorithm \ref{alg:ensemble_update_gaussian}.
Similarly, for an online approximation to bootstrap sampling we might use ``double-or-nothing'' sampling per Algorithm \ref{alg:ensemble_update_bootstrap} \citep{owen2012bootstrapping}.

Algorithm \ref{alg:learn_ensemble_rlsvi} ($\mathtt{learn\_ensemble\_rlsvi}$) presents the ensemble RLSVI algorithm.
An agent that employs $\mathtt{learn\_ensemble\_rlsvi}$ maintains ensemble parameter estimates  $\tilde{\theta}_1, .., \tilde{\theta}_K$ with random prior samples $\tilde{\theta}^p_1, .., \tilde{\theta}^p_K$.
The regulatization effects of $\Rc(\theta; \tilde{\theta}^p_k)$ can play an important role in exploration.
Although we have suggested a particular form in \eqref{eq:regularize}, we would consider alternative approaches based on prior observations as studied in {\it off-policy learning} \citep{precup2001off} and {\it transfer learning} \citep{taylor2009transfer}.

We note that, when used off-policy, minimizing the TD loss may lead to unstable learning and even cause the value function estimate to diverge \citep{tsitsiklis1997analysis}.
This instability can be exacerbated by algorithms such as \texttt{learn\_ensemble\_rlsvi}, which may lead to value estimates computed from data more off-policy than a single value estimate.
In this context, it may be beneficial to replace the naive TD loss with an alternative designed for off-policy learning \citep{sutton2009fast,munos2016safe}.

\begin{algorithm}[!htpb]
\caption{$\mathtt{ensemble\_buffer.update\_gaussian\_noise}(\cdot)$}
\label{alg:ensemble_update_gaussian}

\begin{tabular}{lll}
\textbf{Input:} & $\mathtt{transition}$ & $(s_t, a_t, r_t, s'_t, t)$ \\
\textbf{Agent:} & $v$ & noise variance \\
\textbf{Updates:} & $\mathtt{ensemble\_buffer}$ & replay buffer of $K$-parallel perturbed data
\end{tabular}

\begin{algorithmic}[1]
\For{$k$ in $(1,\ldots,K)$}
\State $\mathtt{ensemble\_buffer[k].enqueue(}(s_t, a_t,  r_t + z^k_t, s'_t, t))$ where $z^k_t \sim N(0, v)$
\EndFor
\end{algorithmic}
\end{algorithm}

\begin{algorithm}[!htpb]
\caption{$\mathtt{ensemble\_buffer.update\_bootstrap}(\cdot)$}
\label{alg:ensemble_update_bootstrap}

\begin{tabular}{lll}
\textbf{Input:} & $\mathtt{transition}$ & $(s_t, a_t, r_t, s'_t, t)$ \\
\textbf{Updates:} & $\mathtt{ensemble\_buffer}$ & replay buffer of $K$-parallel perturbed data
\end{tabular}

\begin{algorithmic}[1]
\For{$k$ in $(1,\ldots,K)$}
\If{$m^k_t \sim {\rm Unif}(\{0, 1\}) = 1$}
\State $\mathtt{ensemble\_buffer[k].enqueue(}(s_t, a_t,  r_t, s'_t, t))$
\EndIf
\EndFor
\end{algorithmic}
\end{algorithm}

\begin{algorithm}[!ht]
\caption{$\mathtt{learn\_ensemble\_rlsvi}$}
\label{alg:learn_ensemble_rlsvi}

\begin{tabular}{lll}
\textbf{Agent:} & $\tilde{\theta}_1, .., \tilde{\theta}_K$ & ensemble parameter estimates \\
& $\tilde{\theta}^p_1, .., \tilde{\theta}^p_K$ & prior samples of parameter estimates \\
& $\Lc_\gamma(\theta \ass ; \theta^- \ass, \Dc \ass)$ & TD error loss function \\
& $\Rc(\theta \ass; \theta^p \ass)$ & regularization function \\
& $\mathtt{ensemble\_buffer}$ & replay buffer of $K$-parallel perturbed data \\

& $\alpha$ & Learning rate \\
\textbf{Updates:} & $\tilde{\theta}$ & agent value function estimate
\end{tabular}

\begin{algorithmic}[1]
\For{$k$ in $(1,\ldots,K)$}
\State Data $\tilde{\Dc}_k \leftarrow \mathtt{ensemble\_buffer[k].sample\_minibatch()}$
\State $\delta \leftarrow \mathtt{buffer.minibatch\_size} \ /\  \mathtt{buffer.size}$
\State $\tilde{\theta}_{k} \leftarrow \tilde{\theta}_{k} - \alpha \nabla_{\theta \mid \theta=\tilde{\theta}_k} \left( \Lc_\gamma(\theta ; \tilde{\theta}_k, \tilde{\Dc}_k) + \Rc(\theta; \tilde{\theta}^p_k) \right) $
\EndFor
\State update $\tilde{\theta} \leftarrow \tilde{\theta}_{j}$ for $j \sim {\rm Unif}({1,..,K})$
\end{algorithmic}
\end{algorithm}


\section{Regret bound}
\label{sec: bounds}

This section provides  a regret analysis of RLSVI in a particularly simple type of decision problem (Section \ref{sec: rl_problem}).
We consider an RLSVI $\mathtt{agent}$ with an infinite buffer, greedy actions and $\mathtt{learn\_grlsvi}$ (Algorithm \ref{alg:learn_rlsvi} with additive Gaussian noise \eqref{eq:gauss_rlsvi}) and a tabular representation.\footnote{A tabular representation for RLSVI means that $\Qc_{\theta} = \theta \in \Real^{|\Sc| \times |\Ac|}$ and $\Qc_{\theta}(s,a) = \theta_{s,a}$.}
The bound we establish applies to a tabular time-inhomogeneous MDP with transition kernel drawn from a Dirichlet prior.
This stylized setting provides rigorous confirmation that RLSVI is capable of performing provably efficient deep exploration in tabular environments.
In addition, we hope this analysis provides a framework for establishing more general guarantees -- for example those applying to RLSVI with linearly parameterized value functions.
Several intermediate lemmas used in the analysis hold under much less restrictive assumptions, and could be useful beyond the setting studied here.

\subsection{Formulation of a time-inhomogenous MDP}

We consider a class of \emph{finite-horizon time-inhomogeneous MDPs}.
This can be formulated as a special case the paper's general formulation as follows.
Assume the state space factorizes as
$\state= \state_0 \cup \state_1 \cup \state_2\cup \cdots \cup \state_{H-1}$
where  the state always advances from some state $s_t \in \state_{t}$ to $s_{t+1} \in \state_{t+1}$ and the process terminates with probability $1$ in period $H$.
For notational convenience, we assume each set $\state_0,...,\state_{H-1}$ contains an equal number of elements.
This is stated formally in the next assumption, which is maintained for all statements in this section.

\begin{assumption}[Finite-horizon time-inhomogeneous MDP]
\label{ass: finite_inhomog}
\hspace{0.00001mm} \newline
The state space factorizes as
$\state= \state_0 \cup \state_1 \cup \state_2\cup \cdots \cup \state_{H-1}$
where $|\state_0|=\cdots = |\state_{H-1}| <\infty$.
For any MDP  $\mdp=(\Sc, \Ac, \Rc, \Pc, \rho)$,
\[
\sum_{s' \in \state_{t+1}} \Pc_{s,a}(s') =1 \qquad \forall t\in \{0,...,H-2\}, \, s\in \state_t, \, a\in \Ac,
\]
and
\[
\sum_{s'\in \state} \Pc_{s,a}(s')=0 \qquad \forall s\in \state_{H-1}, \, a\in \Ac.
\]
\end{assumption}

Each state $s \in \Sc_{t}$ can be written as a pair $s=(t, x)$ where $t \in \{0,...,H-1\}$ and $x \in \Xc = \{1,...,|\Sc_{0}| \}$.
Similarly, a policy $\pi : \Sc\to \Ac$ can be viewed as a sequence $\pi=(\pi_0,...,\pi_{H-1})$ where $\pi_{t}: x \mapsto \pi((t,x))$.
Our notation can be specialized to this time-inhomogenous problem, writing transition probabilities as $\Pc_{t, x, a}(x') \equiv \Pc_{(t, x), a}((t+1, x'))$ and reward probabilities as $\Rc_{t, x, a, x'}(r) \equiv \Rc_{(t, x), a, (t+1, x')}(r) $.
For consistency, we also use different notation for the optimal value function, writing
\[
V_{\Mc, t}^{\pi}(x) \equiv V_{\Mc}^{\pi}((t,x))
\]
and define $V_{\Mc, t}^{*}(x) := \max_{\pi} V_{\Mc, t}^{\pi}(x)$.
Similarly, we can define the state-action value function under the MDP at timestep $t\in \{0,...,H-1\}$ by
\[
Q^*_{\Mc, t}(x,a) = \E[ r_{t+1} +  V_{\Mc, t+1}^{*}(x_{t+1})  \mid \Mc, x_{t}=x, a_{t}=a] \qquad \forall x\in \Xc, a\in \Ac.
\]
This is the expected reward accrued by taking action $a$ in state $x$ and proceeding optimally thereafter.

Upon choosing an action, the algorithm observes a pair $o=(x', r)$ consisting of a state transition and a reward.
We will refer to this pair $o$ as an \emph{outcome} of the decision.
Assumptions about the distribution of rewards and state-transitions can be more compactly written as an assumption about outcome distributions.
We study the regret of RLSVI under the following Bayesian model for the MDP $\Mc$.
This assumption is not required for some of the results in this section, and we will specify when it is needed.

\begin{assumption}[Independent Dirichlet prior for outcomes]
\label{assumption: prior}
\hspace{0.00001mm} \newline
Rewards take values in $\{0,1\}$ and so the cardinality of the outcome space is $|\Xc \times \{0,1\}| =2|\Xc|.$ 
For each, $(t,x,a) \in \{0,...,H-2\} \times \Xc \times \Ac$, the outcome distribution is drawn from a Dirichlet prior
\[
\Pc^{O}_{t,x,a}(\cdot)\sim {\rm Dirichlet}(\alpha_{0,t,x,a})
\]
for $\alpha_{0,t, x, a} \in \Re^{2|\Xc|}_{+}$ and each $\Pc^{O}_{t,x,a}$ is drawn independently across $(t,x,a)$.
Assume there is $\beta \geq 3$ such that $\Ind^T \alpha_{0, t,a,x}=\beta$ for all $(t,x,a)$.
\end{assumption}

\subsection{Bayesian regret bound}

The following theorem is the main result of this section, and establishes a polynomial bound on the Bayesian regret of RLSVI.
\begin{theorem}[Bayesian regret bound for RLSVI]
\label{thm: regret}
\hspace{0.0001mm} \newline
Consider an RLSVI $\mathtt{agent}$ with an infinite buffer, greedy actions and $\mathtt{learn\_grlsvi}$ with tabular representation.\footnote{\texttt{learn\_grlsvi} is \texttt{learn\_rlsvi} (Algorithm \ref{alg:learn_rlsvi}) with additive Gaussian reward noise of variance $v$ \eqref{eq:gauss_rlsvi}.\\ A tabular representation for RLSVI means that $\Qc_{\theta} = \theta \in \Real^{|\Sc| \times |\Ac|}$ and $\Qc_{\theta}(s,a) = \theta_{s,a}$.} 
Under Assumption \ref{assumption: prior} with $\beta\geq 3$, if this version of RLSVI is applied with planning horizon $H$, and parameters $v=3H^2$, $\bar{\theta}= H \Ind$ and $v/\lambda = \beta$, then for all $L\in \mathbb{N}$,
\begin{equation}\label{eq: first regret bound}
        {\rm BayesRegret}({\rm RLSVI}_{\bar{\theta}, v, \lambda}, L) \leq 6H^2 \sqrt{\beta |\Xc||\Ac| L \log_{+}(1+|\Xc||\Ac|HL)}\log_{+}\left(1+ \frac{L}{|\Xc||\Ac|}\right),
    \end{equation}
    and
  {
  \begin{eqnarray}\label{eq: second regret bound}
     \hspace{20pt} {\rm BayesRegret}({\rm RLSVI}_{\bar{\theta}, v, \lambda}, L) &\leq& 5\beta H^3 |\Xc|||\Ac| \sqrt{\log_{+}(1+|\Xc| |\Ac| HL) }\log_+\left(1+ \frac{L}{|\Xc||\Ac|}\right) \\ \nonumber
    && + 2 H^2 \sqrt{6|\Xc| |\Ac|L \log(|\Xc||\Ac|) }
    \end{eqnarray}
  }
    where $\log_{+}(x) = \max\{1,\log(x)\}$.
\end{theorem}
Let us focus on the first bound given in equation \eqref{eq: first regret bound}. The parameter $\beta$ governs the relative strength of prior mean $\bar{\theta}$ in the $Q$-functions sampled by RLSVI.
We typically think of $\beta$ as a constant, reflecting situations with weak prior knowledge of the optimal value function that does not grow with variables $H, S, A, L$.
In this case, this regret bound is $\tilde{O}(H^2 \sqrt{|\Xc| |\Ac| L})$ where $\tilde{O}$ ignores poly-logarithmic factors.
Note that since $|\Sc_0|=...=|\Sc_{H-1}|=|\Xc|$ then $|\Sc| = |\Xc|H$ and for $T=LH$ denoting the number of periods,
\[
{\rm BayesRegret}({\rm RLSVI}_{\bar{\theta}, v, \lambda}, L)    = \tilde{O}(H \sqrt{|\Sc||\Ac| T}).
\]
This bound reveals that RLSVI requires a number of episodes that is just linear in the number of states to reach near optimal performance.
Indeed, it is possible to guarantee cumulative Bayesian regret less than $L\epsilon$ with a value of $L$ that scales with $|\Xc| / \epsilon^2$.
In general, at least order $|\Xc|^2$ samples are required to learn the transition kernel $\Pc_{t,x,a}$.
Therefore, for large $|\Xc|$ we prove that RLSVI learns to make near-optimal decisions using fewer samples than would be required to learn the transition dynamics of the MDP.

It is interesting to compare this Bayesian regret bound with bounds that have been established for other tabular reinforcement learning algorithms.
 The results of \cite{Bartlett2009} and \cite{Jaksch2010} are not directly comparable to the bound established for RLSVI, as they develop bounds on minimax, rather than Bayesian regret, and study classes of MDPs satisfying recurrence assumptions, rather than episodic MDPs.
 However, it is worth noting that because these algorithms attempt to represent each transition probability $\Pc_{t,x,a}(x')$ accurately, applying their analysis to our problem yields a regret bound of $\tilde{O}(H^2 |\Xc| \sqrt{|\Ac| L}) $,  which has is larger dependence on $|\Xc|$.

The second bound given in equation \eqref{eq: second regret bound} reveals the dependence of regret on $\beta$ more precisely. This bound is $\tilde{O}(\beta H^3 |\Xc| |\Ac| + H^2\sqrt{|\Xc||\Ac| L })$.
The first term in this regret bound can roughly be thought of as a bound on the regret incurred throughout an initial phase of the algorithm, during which it gathers data that overwhelms the prior mean.
When the number of episodes $L$ is large, the dominant term is the second one, which is $\tilde{O}(H^2\sqrt{|\Xc||\Ac| L })$ and has no dependence on $\beta$.

\subsection{Stochastic Bellman operators}
\label{subsec: stochastic Bellman operators}

 Any state-action value function $Q\in \Re^{|\Xc||\Ac|}$ induces a value function $V(x)=\max_{a\in \Ac}Q(x,a)$ that maps each state to a real number.
 To simplify the analysis, it is useful to introduce nonstandard notation for the value function over outcomes $o=(r,x)$.
\begin{definition}[Induced value function]
\label{def: induced_value}
\hspace{0.00001mm} \newline
For a state-action value function $Q\in \Re^{|\Xc||\Ac|}$   define the corresponding value function $V_Q \in \Re^{2|\Xc|}$ over outcomes by $V_Q(r,x') := r+\max_{a \in \Ac} Q(x',a)$ for all $x'\in \Xc$ and $r\in \{0,1\}$.
\end{definition}
It is useful to also keep notation for the empirical distribution over observed outcomes. Let
\[
D_{\ell-1}(t,x,a)=\{(r_{t+1}^k, x_{t+1}^k) : k < \ell, x_t^k=x, a_{t}^{k} = a \}
\]
be the set of data observed up to episode $\ell$ when action $a$ was chosen in $(t,x)$, and set $n_{\ell}(t, x,a) = |D_{\ell-1}(t,x,a) |$ to be number of past observations of the triple $(t,x,a)$.
For ease of notation we will write $y$ for the timestep, state, and action $y := (t,x,a)$.
Denote by $\hat{P}^{O}_{\ell, y}(r',x')$ the empirical distribution over outcomes $(r',x')$ in the dataset $D_{\ell-1}(y)$.

This section introduces the Bellman operator underlying the MDP $\Mc$ and a notion of a Bellman operator that underlies the recursion defining RLSVI.
Due to the randomness in $\Mc$ under Assumption \ref{assumption: prior} and the Gaussian noise added by RLSVI iterations both of these can be viewed as \emph{stochastic Bellman operators}, as applying one of these operators to a state action value function $Q\in \Re^{|\Xc||\Ac|}$ generates a random state-action value function as output.

\paragraph{True Bellman Operator.}
For $Q:\Xc \times \Ac \to \mathbb{R}$ the true Bellman operator at timestep $t$ applied to $Q$ is defined by
\begin{eqnarray*}
    F_{\Mc, t}Q(x,a) &=& \E[ r_{t+1} + \max_{a'\in \Ac} Q(x_{t+1}, a')  \mid \Mc, x_{t}=x, a_{t}=a] \\
    & = & \E[V_{Q}(r_{t+1}, x_{t+1}) \mid \Mc, x_t=x, a_t =a] \\
    &=& V_{Q}^T \Pc^{O}_{t,x,a}.
\end{eqnarray*}
Applying $F_{\Mc, t}$ backward in time produces a sequence of  optimal state-action value functions satisfying  $Q^*_{\Mc, H}= 0$ and the Bellman equation $Q^*_{\Mc, t}=F_{\Mc, t}Q^*_{\Mc, t+1}$ for $t<H$.
Under Assumption \ref{assumption: prior}, this can be viewed as a randomized Bellman operator due to the randomness in the MDP $\Mc$.

Under Assumption \ref{assumption: prior}, the posterior transition probabilities are distributed as
\[
\Pc_{y}^{O}(\cdot)| \hist \sim {\rm Dirichlet}(\alpha_{\ell,y})
\]
where
\begin{equation}
\label{eq: p_rlsvi}
\alpha_{\ell, y} = \alpha_{0,y}
+ n_\ell(y) \hat{P}_{\ell, y}^{O} \in \Re^{2|\Xc|}
\end{equation}
for any triple $y=(t,x,a)$.
These determine the posterior mean of $\Pc_{y}^{O}$ as a weighted linear combination of the prior and the empirical observations:
\[
\E[\Pc_{y}^{O} \mid \hist]=  \frac{\alpha_{0,y}+n_\ell(y) \hat{P}^{O}_{\ell, y}}{\beta+ n_\ell(y)}.
\]

\paragraph{Bellman operator of RLSVI.}
In episode $\ell$, we can  define a notion of a Bellman operator underlying the recursion of RLSVI. Define
\begin{gather*}
F_{\ell, t}Q(x,a) := \sigma^{2}_{\ell}(t,x,a) \left(\frac{\bar{\theta}_{t,x,a}}{\lambda}+\frac{1}{v}\left(\sum_{\substack{(r, x')\in\\D_{\ell-1}(t,x,a)}} r +\max_{a' \in \Ac} Q(x', a')\right)\right) + w_{\ell}(t,x,a) \\
\sigma^{2}_{\ell}(t,x,a)= \left( \frac{1}{\lambda}+\frac{n_{\ell}(t,x,a)}{v} \right)^{-1} = \frac{v}{n_{\ell}(t,x,a)+v/\lambda}  \\
w_{\ell}(t,x,a) \mid \hist \sim N(0, \sigma_{\ell}^2(t,x,a))
\end{gather*}
where $w_{\ell}(y)/\sigma_{\ell}(y) \sim N(0,1)$ is drawn independently across episodes $\ell$ and triples $y=(t,x,a)$.

In episode $\ell$ RLSVI generates a sequence of state-action value functions $Q_{\ell,1},...,Q_{\ell,H}$  where $Q_{\ell, H} = 0 \in \Re^{|\Xc||\Ac|}$ consists of all zeros and for all $t< H$, $Q_{\ell,t}= F_{\ell, t}Q_{\ell, t+1}$.
RLSVI chooses actions greedily with respect to this sequence of state-action value functions.
Note that because of the Gaussian sampling noise, the action $\arg\max_{a\in \Ac} Q_{\ell}(x,a)$ is unique with probability one for any $x$ and $\ell$.
 Therefore the policy applied by RLSVI in an episode is completely determined by the state-action value functions it samples.

We can also express the RLSVI Bellman update in a simple way using the empirical distribution $\hat{P}^{O}_{\ell, y}$ over past outcomes resulting from $y=(t,x,a)$. We have
\[
\sum_{\substack{(r, x')\in D_{\ell-1}(y)}} \left(r +\max_{a' \in \Ac} Q(x', a')\right)= n_{\ell}(y) V_{Q}^T \hat{P}^{O}_{\ell,y}.
\]
Direct calculation gives the following alternate expression
\begin{equation}
\label{eq: RLSVI Bellman}
  F_{\ell, t}Q(x,a)= \frac{(v/\lambda)\bar{\theta}+n_{\ell}(y)V^T_Q \hat{\Pc}^O_{\ell,y}}{(v/\lambda)+n_{\ell}(y)}+w_{\ell}(y) \qquad \forall y=(t,x,a).
\end{equation}
This shows that the Bellman update of RLSVI differs from the empirical Bellman update
$V_{Q}^T \hat{\Pc}^O_{\ell,y}$ in two ways:
there is slight regularization toward the prior mean $\bar{\theta}$, and more importantly, RLSVI adds independent Gaussian noise to each update.

\subsection{Optimism and regret decompositions}
The next lemma forms a crucial element of the proof.

\begin{lemma}[Planning Error to On Policy Bellman Error]
\label{lem: planning to Bellman}
\hspace{0.000001mm} \newline
Let $Q_0, Q_1, Q_2,...,Q_{H}\in \Re^{|\Xc| |\Ac|}$ be any sequence with $Q_{H}=0\in \Re^{|\Xc| |\Ac|}$ and take $\pi=(\pi_0, \pi_1,..., \pi_{H-1})$ to be a policy with $\pi_t(x) \in \arg\max_{a\in \Ac} Q_t(x,a)$ for all $x$.
Then for any MDP $\Mc$ and initial state $x\in \Xc$,
\[
Q_0(x, \pi_{0}(x))- V^{\pi}_{\Mc,0}(x) = \E_{\Mc, \pi}\left[ \sum_{t=0}^{H-1} (Q_{t}-F_{\Mc, t}Q_{t+1})(x_t, a_t)   \mid x_0=x \right].
\]
\end{lemma}
\begin{remark}
    To interpret this lemma, consider an algorithm that  generates a sequence of state-action value functions $Q_0,...,Q_H\in \Re^{|\Xc||\Ac|}$ and chooses actions greedily with respect to this sequence.
    We can interpret $Q_{0}(x, \pi_{0}(x))$ to be the algorithm's estimate of the value of following this greedy policy throughout the episode from a starting state $x$, while $V^{\pi}_{\Mc,0}(x)$ denotes the true expected value.
    One can interpret $Q_t - F_{\Mc}Q_{t+1}$ as the error in Bellman's equation at stage $t$.
    The right hand side of of the equation in Lemma \ref{lem: planning to Bellman} measures Bellman error on policy, i.e. at the states and actions that the agent is expected to sample by following the policy throughout the episode.
    This lemma says that the prediction $Q_{0}(x, \pi_{0}(x))$ can be far from the true value function only when on policy Bellman error is large.
\end{remark}

Lemma \ref{lem: planning to Bellman} is a powerful tool for studying the regret of optimistic algorithms.
The regret of the policy $\pi$ in Lemma \ref{lem: planning to Bellman} incurred in a single episode can always be decomposed as
\[
V_{\Mc,0}^*(x)-V_{\Mc,0}^{\pi}(x)  =  \left( \max_{a\in \Ac}Q_{\Mc,0}^*(x,a)-\max_{a \in \Ac} Q_{0}(x,a) \right) +\left(\max_{a \in \Ac} Q_{0}(x,a) -  V_{\Mc,0}^{\pi}(x) \right),
\]
where we have used the fact that $V_{\Mc,0}^*(x) = \max_{a\in \Ac} Q_{\Mc,0}^*(x,a)$.
The second term in this decomposition can be rewritten using  Lemma \ref{lem: planning to Bellman}. In particular, for any sequence $Q_{0},...,Q_H \in \Re^{|\Xc||\Ac|}$ with $Q_H=0$ and policy $\pi = (\pi_0,...,\pi_{H-1})$ under which actions $\pi_{t}(x)= \arg\max_{a\in \Ac} Q_t(x,a)$ are chosen greedily with respect these $Q$--functions, regret can be decomposed as follows:
\begin{eqnarray}\nonumber
    V_{\Mc,0}^*(x)-V_{\Mc,0}^{\pi}(x)=
    &  \max_{a\in \Ac} Q^{*}_{\Mc,0}(x,a)-\max_{a\in \Ac}Q_{0}(x,a) & (\text{pessimism of } Q_0) \\\label{eq: regret decomposition}
    +& \E_{\Mc, \pi}\left[ \sum_{t=0}^{H-1} (Q_{t}-F_{\Mc,t }Q_{t+1})(x_t, a_t) \mid x_0=x\right] & (\text{on policy Bellman error}).
\end{eqnarray}
If the function $Q_0$ is optimistic at an initial state $x$, in the sense that $\max_{a} Q_0(x,a) \geq \max_{a} Q_{\Mc,0}^*(x,a)$, then regret in the episode is bounded by on policy Bellman error under $(Q_0,...,Q_H)$.

One can apply this regret decomposition to study RLSVI by taking $(Q_0,...,Q_H)$ to be the sequence $(Q_{\ell,0},...,Q_{\ell,H})$ generated by RLSVI in some episode $\ell$. On policy Bellman error can be simplified further by plugging in $Q_{\ell,t}= F_{\ell, t} Q_{\ell, t+1}$.
The next corollary of Lemma \ref{lem: planning to Bellman} then follows by taking expectations on both sides of equation \eqref{eq: regret decomposition}.

\begin{corollary}[Optimistic regret bounds]
\label{cor: RLSVI regret decomposition}
\hspace{0.0001mm} \newline
For any episode $\ell \in \N$, if
\begin{equation}\label{eq: optimism in expectation}
\E \left[ \max_{a\in \Ac} Q_{\ell,0}(x_{0}^{\ell},a) \right] \geq \E\left[\max_{a\in \Ac}Q_{\Mc,0}^*(x_{0}^{\ell},a)\right]
\end{equation}
then
\[
\E\left[ V_{\Mc,0}^*(x^{\ell}_0)-V_{\Mc,0}^{\pi_{\ell}}(x^{\ell}_0) \right] \leq   \E\left[ \sum_{t=0}^{H} (F_{\ell, t} Q_{\ell, t+1} -F_{\Mc,t }Q_{\ell,t+1})(x^{\ell}_t, a^{\ell}_t) \right].
\]
\end{corollary}

Corollary \ref{cor: RLSVI regret decomposition} forms the core of our analysis.
The next section establishes that \eqref{eq: optimism in expectation} holds in every episode $\ell\in \mathbb{N}$.
We then complete the proof by bounding the cumulative on policy Bellman error throughout $L$ episodes.

\subsection{Stochastic optimism}
Our goal is to show equation \ref{eq: optimism in expectation} holds when RLSVI is applied with appropriate parameters.
We will instead prove that under Assumption \ref{assumption: prior} the stronger condition that
\[
\E \left[ \max_{a\in \Ac} Q_{\ell,0}(x_{0}^{\ell},a)  \mid \hist \right] \geq \E\left[\max_{a\in \Ac}Q_{\Mc,0}^*(x_{0}^{\ell},a) \mid \hist \right]
\]
holds for any history $\hist$.
By the tower property of conditional expectation, this clearly implies equation \eqref{eq: optimism in expectation}.

As highlighted in Subsection \ref{subsec: stochastic Bellman operators},  both $Q_{\ell,0}= F_{\ell, 0}\cdots F_{\ell, H-1}0$ and $Q^*_{\Mc,0}=F_{\Mc, 0}\cdots F_{\Mc, H-1}0$ are calculated through recursive backward application of stochastic Bellman operators.
The distributions of $Q_{\ell,0}$ and $Q_{\Mc, 0}^*$ generated in this fashion is complicated and difficult to study directly.
Instead, we study properties of the stochastic Bellman operators themselves.
We establish a strong sense in which $F_{\ell,t}$ generates random $Q$-functions $F_{\ell,t}Q$ that are optimistic compared to those generated by applying $F_{\Mc,t}$ to $Q$.
We then show this optimism is preserved under recursive application of the stochastic Bellman operators, which will imply the optimism of the final iterate $Q_{\ell,0}$.
This strong notion of optimism is defined below.

\begin{definition}[Stochastic optimism]
\label{def: stoch_opt}
\hspace{0.0001mm} \newline
A random variable $X$ is stochastically optimistic with respect to another random variable $Y$, written $X \succeq_{SO} Y$, if  for all convex increasing functions $u:\Re \to \Re$
\begin{equation}\label{eq: stochastic optimism}
\E[u(X)] \geq \E[u(Y)].
\end{equation}
\end{definition}

This definition closely mirrors that of ``second order stochastic dominance'', which is widely used in decision theory \citep{hadar1969rules}.
A random payout $X$ is second order stochastically dominant with respect to $Y$ if \eqref{eq: stochastic optimism} holds for all \emph{concave} increasing function $u$.
This means that any rational \emph{risk-averse} agent prefers $X$ to $Y$, while $X \succeq_{SO} Y$ implies that any rational \emph{risk-loving} agent prefers $X$ to $Y$.
Intuitively, this requirement means that draws of $X$ generate payouts that are larger and noisier than $Y$.
Our goal then is to show if RLSVI is applied with appropriate parameters, it generates iterates that are larger and noisier than the true $Q-functions$.

\begin{example}[Stochastic optimism in Gaussian random variables]
\label{ex: gauss_stoch_opt}
\hspace{0.0001mm} \newline
    If $X\sim N(\mu_X, \sigma^2_X)$ and $Y\sim N(\mu_Y, \sigma^2_Y)$ then $X\succeq_{SO} Y$ if and only if $\mu_X\geq \mu_Y$ and $\sigma^2_X \geq \sigma^2_Y$.
\end{example}

The following observation is key to our analysis.
\begin{lemma}[Preservation of optimism under convex operations]
\label{lem: convex operations}
\hspace{0.0001mm} \newline
  For any two collections $(X_1,...,X_n)$ and $(Y_1,...,Y_n)$ of independent random variables with $X_i \succeq_{SO} Y_i$ for each $i\in \{1,...n\}$ and any convex increasing function $f:\Re^n \to \Re$,
    \[
    f(X_1,...,X_n) \succeq_{SO} f(Y_1...,Y_n).
    \]
\end{lemma}

Two special cases of Lemma \ref{lem: convex operations} imply that the partial ordering of stochastic optimism is preserved under convolution and maximization.
In particular, for any independent random variables $(X,Y,Z)$ if $X \succeq_{SO} Y$ we can conclude\footnote{This follows from Lemma \ref{lem: convex operations} by looking at the pairs $(X,Z)$ and $(Y,Z)$ and taking $f:\Re^2 \to \Re$ to be $f(x_1,x_2)=x_1+x_2$.}
$X+Z \succeq_{SO} Y+Z$.
For two pairs of independent random variables $(X_1, X_2)$ and $(Y_1, Y_2)$ with $X_1 \succeq_{SO} Y_1$ and $X_2 \succeq_{SO} Y_2$,
\[
\max\{X_1, X_2 \} \succeq_{SO} \max\{ Y_1, Y_2 \}.
\]
This implies the following monotonicity property of the Bellman operator $F_{\ell, t}$ underlying RLSVI.
This will later enable us to show that if initial iterates of RLSVI are stochastically optimistic, then this optimism is preserved under recursive application of the stochastic Bellman operators $F_{\ell,0}\cdots F_{\ell, H-1}$.

\begin{lemma}[Monotonicity]
\label{lem: monotonicity}
\hspace{0.0001mm} \newline
Fix two random $Q$ functions $Q_1, Q_2 \in \Re^{|\Xc||\Ac|}$.
Suppose that conditioned on $\hist$, for each $i=1,2$ the entries of $Q_i(x,a)$ are drawn independently across $x,a$, and drawn independently of the RLSVI noise terms $w_{\ell}(t,x,a)$.
Then
\[
Q_1(x,a) \mid \hist  \succeq_{SO} Q_2(x,a) \mid \hist \qquad \forall (x,a) \in \Xc\times \Ac \]
implies
\[
F_{\ell, t} Q_1(x,a) \mid \hist \succeq_{SO} F_{\ell, t} Q_2(x,a) \mid \hist \qquad \forall (x,a) \in \Xc\times \Ac,\, t\in \{0,...,H-1\}.
\]
\end{lemma}

\begin{proof}
    Conditioned on $\mathcal{H}_{\ell-1}$,
    \[
    F_{\ell,t} Q(x,a)= \frac{\sigma^{2}_{\ell}(t,x,a)\bar{\theta}_{t,x,a}}{\lambda}+ \frac{\sigma^{2}_{\ell}(t,x,a)}{v} \left(
    \sum_{\substack{(r, x')\in\\D_{\ell-1}(t,x,a)}} r +\max_{a' \in \Ac} Q(x', a')\right) + w_{\ell}(t,x,a)
    \]
    is a convex function of $(Q(x',a'))_{x'\in \Xc, a'\in \Ac}$ convolved with the independent noise term $w_{\ell}(t,x,a)$.
    The result therefore follows by Lemma \ref{lem: convex operations}.
\end{proof}

Consider the random variable $Y=P^T V$ where $V \in \Re^n$ and $P\sim {\rm Dirichlet}(\alpha)$.
Then, $Y$ has mean $V^T \alpha / \Ind^T \alpha$.
The size of its fluctuations depends on how concentrated $P$ is around its mean, captured by the pseudocount $\Ind^T\alpha=\sum_{i=1}^{n} \alpha_i$, and spread of the elements in $V$, captured by ${\rm Span}(V) \equiv \max_{i} V_i- \min_{j} V_j$.
The next lemma shows that a Gaussian random variable large enough mean and variance is stochastically optimistic with respect to $Y$.
This result is established in the appendix. 

\begin{lemma}[Gaussian vs Dirichlet optimism]
    \label{lem: Dir Norm}
    \hspace{0.0001mm} \newline
    Let $Y = P^T V$ for $V \in \Re^n$ fixed and $P \sim {\rm Dirichlet}(\alpha)$ with $\alpha \in \Re^n_+$ and $\sum_{i=1}^n \alpha_i \ge 3$.
    Let $X \sim N(\mu, \sigma^2)$ with
    $ \mu \geq \frac{\sum_{i=1}^n \alpha_i V_i}{\sum_{i=1}^n \alpha_i},
    \ \sigma^2 \geq 3 \left(\sum_{i=1}^n \alpha_i \right)^{-1} {\rm Span}(V)^2$, then $X \succeq_{SO} Y$.
\end{lemma}

With Lemma \ref{lem: Dir Norm} in place, we can now establish a sense in which the Bellman operator underlying RLSVI is stochastically optimistic relative to the true Bellman operator. Recall definition \ref{def: induced_value}, which defines the value over outcomes $(r, x')$ under $Q$ by $V_{Q}(r, x')\equiv r+ \max_{a'\in \Ac} Q(x',a')$.

\begin{lemma}[Stochastically optimistic operators]
\label{lem: optimism of RLSVI operator}
\hspace{0.0001mm} \newline
Suppose Assumption \ref{assumption: prior} holds and RLSVI is applied with parameters $(\bar{\theta}, v, \lambda)$ satisfying $(v / \lambda) = \beta$.
Then for any episode $\ell$ with history $\hist$, time $t\in \{0,...,H-1\}$, and pair $(x,a)\in \Xc \times \Ac$,
\[
    F_{\ell, t}Q(x,a) \mid \hist \succeq_{SO} F_{\Mc, t} Q(x,a) \mid \hist
\]
for any fixed $Q \in \Re^{|\Xc||\Ac|}$ such that $v \geq 3 {\rm Span}(V_Q)^2$ and $\max_{x\in \Xc} V_{Q}(x)\leq \min_{t,x,a}\overline{\theta}_{t,x,a}$.
\end{lemma}

\begin{remark}
When $Q(x,a)\geq 0$ for all $(x,a)$, ${\rm Span}(V_Q) \leq \| V_Q\|_{\infty} \leq \| Q\|_{\infty}+1$.
Therefore it suffices that $v \geq 3(1+\| Q\|_{\infty})^2$ and $\min_{y}\bar{\theta}_{y} \geq \| Q \|_{\infty}+1$.
\end{remark}
\begin{proof}
Recall the Bellman update of $Q$ under the true MDP $\Mc$ is
 \[
 F_{\Mc,t} Q(x,a) = V_{Q}^T \Pc^{O}_{t,x,a}
\]
For each $y=(t,x,a)$, $\Pc^{O}_{y}| \hist \sim {\rm Dirichlet}(\alpha_{\ell,y})$ with
$
\alpha_{\ell, y} = \alpha_{0,y}
+ n_\ell(y) \hat{P}_{\ell, y}^{O} \in \Re^{2|\Xc|}.
$
Similarly, for each $y=(t,x,a)$, plugging in $\beta=v/\lambda$ we have
\[
F_{\ell, t} Q(x,a) \mid \hist \sim N( \mu_y, \sigma^2_y)
\]
where
\begin{eqnarray*}
\mu_y \equiv \frac{\beta\bar{\theta}_y+n_{\ell}(y)V^T_Q\hat{\Pc}^O_{\ell,y}}{\beta+n_{\ell}(y)} &\quad& \sigma^2_y \equiv  \frac{v}{n_{\ell}(y)+\beta}.
\end{eqnarray*}
The result follows from Lemma \ref{lem: Dir Norm} if we establish  $\sigma^2_{y} \geq (\Ind^T \alpha_{\ell, y})^{-1} {\rm Span}(V_Q)^2$ and $\mu_{y} \geq V_Q^T \alpha_{\ell, y} / \Ind^T\alpha_{\ell, y}$.
We have
\[
 \frac{3 \cdot {\rm Span}(V_Q)^2}{\Ind^T \alpha_{\ell, y}} = \frac{3 \cdot {\rm Span}(V_Q)^2}{\beta + n_{\ell}(y)}  \leq \sigma_{y}^2
\]
because of the assumption that $v \geq 3 \cdot {\rm Span}(V_Q)^2$. Next we have
\[
\frac{ V_Q^T \alpha_{\ell, y}}{\Ind^T\alpha_{\ell, y}}  = \frac{ V_Q^T\alpha_{0,y}  +n_{\ell}(y)V_Q^T \hat{\Pc}^{O}_{y}  }{\beta + n_{\ell}(y)} \leq  \frac{\beta \max_{x\in \Xc}V_Q(x) + n_{\ell}(y) V_Q^T \hat{\Pc}^{O}_{y}}{\beta +n_{\ell}(y)} \leq \mu_y
\]
because of the assumption that $V_{Q}(x)\leq \min_{y} \bar{\theta}_{y}$ for all $x$.
\end{proof}
Lemmas \ref{lem: monotonicity} and \ref{lem: optimism of RLSVI operator} together imply the stochastic optimism of the state-action value functions $Q_{\ell,0}$ generated by RLSVI.
\begin{corollary}\label{cor: optimism of RLSVI}
    If Assumption \ref{assumption: prior} holds and RLSVI is applied with parameters $(\bar{\theta}, v, \lambda)$ satisfying $(v / \lambda) = \beta$, $v \geq 3 H^2$ and $\min_{y} \bar{\theta}_{y} \geq H$,
    \[
    Q_{\ell,0}(x,a) \mid \hist \succeq_{SO} Q^*_{\Mc, 0}(x,a) \mid \hist
    \]
    for any history $\Hc_{\ell-1}$ and state-action pair $(x,a)\in \Xc\times \Ac$.
\end{corollary}
\begin{proof}
To reduce notation, we prove this for episode $\ell=1$, but the proof follows identically for general $\ell$ by conditioning on the history $\hist$ at every step. Recall that $Q_{1,0}=F_{1,0} F_{1,1}\cdots F_{1, H-1} 0$ and $Q_{\Mc,0}^* = F_{\Mc,0} F_{\Mc,1}\cdots F_{\Mc, H-1} 0$.

By Lemma \ref{lem: optimism of RLSVI operator},
\[
(F_{1, H-1} 0)(x,a) \succeq_{SO} (F_{\Mc, H-1} 0)(x,a)  \qquad \forall x,a.
\]
Proceeding by induction, suppose for some $t\leq H-1$
\[
\left(F_{1,t+1} F_{1,t+2}\cdots F_{1, H-1} 0\right)(x,a)  \succeq_{SO} \left(F_{\Mc,t+1} F_{\Mc,t+2}\cdots F_{\Mc, H-1} 0\right)(x,a)  \qquad \forall x,a.
\]
Combining this with Lemma \ref{lem: monotonicity} shows
\begin{eqnarray*}
F_{1,t}\left(F_{1,t+1} F_{1,t+2}\cdots F_{1, H-1} 0\right)(x,a)  &\succeq_{SO}&  F_{1,t}\left(F_{\Mc,t+1} F_{\Mc,t+2}\cdots F_{\Mc, H-1} 0\right)(x,a) \\
&\succeq_{SO} & F_{\Mc, t}\left(F_{\Mc,t+1} F_{\Mc,t+2}\cdots F_{\Mc, H-1} 0\right)(x,a)
\end{eqnarray*}
where the final step uses Lemma \ref{lem: optimism of RLSVI operator} combined with the fact that for any $t\in \{0,..,H-1\}$,
\[
Q \equiv F_{\Mc,t+1} F_{\Mc,t+2}\cdots F_{\Mc, H-1} 0
\]
satisfies $3 \cdot {\rm Span} (V_{Q})\leq 3H \leq v$ and $Q\leq \bar{\theta}$.
\end{proof}

\subsection{Analysis of on-policy Bellman error: proof of Theorem \ref{thm: regret}}
The proof relies on the following bound.
For standard Gaussian random variables $X_1,...,X_n$, a basic Gaussian maximal inequality implies $\E [ \max_{i} X_i ] \leq \sqrt{2\log(n)}$. The next lemma is a generalization of this result, which can be seen by taking $J=\arg\max_{j} X_j$. This lemma is implied by Proposition A.1. of \citep{russo2015much}.
\begin{lemma}
Let $(X,J)$ be jointly distributed random variables where $X \in \mathbb{R}^n$ follows a multivariate Gaussian distribution with $X_j \sim N(0,\sigma_j^2)$ and $J\in \{1,...n\}$ is a random index.
Then
\[
\E[X_J] \leq \sqrt{2\log(n)\E[\sigma_{J}^2]}.
\]
\end{lemma}
Applying this leads to two bounds that are used in our analysis. The first bounds the noise terms $w_{\ell}(t, x_t, a_t)$ of RLSVI at the state and action visited by RLSVI, and the second bounds the norm of the value function sampled by RLSVI.
\begin{corollary}\label{cor: bound on expected noise}
    For each $t \leq H$ and $\ell \leq L$
    \[
    \E[ w_{\ell}(t, x_t, a_t)] \leq \sqrt{2\log(|\Ac||\Xc|)\E[\sigma_{\ell}(t, x_t, a_t)^2 ]}.
    \]
\end{corollary}

\begin{corollary}\label{cor: bound on sampled value function}
 If RLSVI is applied with parameters $(\lambda, v, \bar{\theta})$ with $v/\lambda=\beta\geq3 $ , $v=3H^2$ and $\bar{\theta}=H\Ind$,
\[
\E[ \max_{\ell\leq L, t< H} \|V_{Q_{\ell,t+1}}\|_{\infty}] \leq 2H+H^2\sqrt{2\log(1+|\Xc| |\Ac| HL) }.
\]
\end{corollary}
A proof of this corollary is provided in the appendix. We now complete the regret analysis of RLSVI and establish Theorem \ref{thm: regret}.
\begin{proof}
Set
\[
\Delta_{\ell} = V_{\Mc, 0}^*(x^{\ell}_0)-V_{\Mc, 0}^{\pi_{\ell}}(x^{\ell}_0).
\]
By Corollary \ref{cor: RLSVI regret decomposition} and Corollary \ref{cor: optimism of RLSVI},
\[
\E\left[ \sum_{\ell=1}^{L} \Delta_{\ell}\right] \leq
   \E\left[ \sum_{\ell=1}^{L}\sum_{t=0}^{H-1} (F_{\ell, t} Q_{\ell, t+1} -F_{\Mc,t }Q_{\ell,t+1})(x^{\ell}_t, a^{\ell}_t) \right].
\]
The posterior-mean Bellman update of $Q$ under $\Mc$ is
\[
\E[F_{\Mc, t}Q(x,a) | \hist] = V_Q^T \E[\Pc^{O}_{t,x,a} | \hist].
\]
Recall as well that for each $y=(t,x,a)$, $\Pc^{O}_{y}| \hist \sim {\rm Dirichlet}(\alpha_{\ell,y})$ with
\[
\alpha_{\ell, y} = \alpha_{0,y}
+ n_\ell(y) \hat{P}_{\ell, y}^{O} \in \Re^{2|\Xc|}.
\]
Since the prior over $\Pc^{O}_{t,x,a}(\cdot)$ is distributed independently  across states and actions $(t,x,a)$, and $(t,x,a)$ cannot be visited prior to period $t$ in any episode, we have also that
\[
\Pc^{O}_{t,x,a} | \hist, x^{\ell}_{0}, a_{0}^{\ell},..,x^{\ell}_{t}, a^{\ell}_{t} \sim {\rm Dirichlet}(\alpha_{\ell,y}).
\]
As a result
\begin{eqnarray*}
\E[ F_{\Mc, t}Q(x,a) | \hist, x^{\ell}_{0}, a_{0}^{\ell},..,x^{\ell}_{t}, a^{\ell}_{t} ] &=&\E[ F_{\Mc, t}Q(x,a) \mid \hist]\\
 &=& \frac{V_Q^T\alpha_{0,y} + n_{\ell}(y) V_{Q}^T \hat{P}^{O}_{\ell,y} }{\beta + n_{\ell}(y) } \\
 &\geq& \frac{-\beta\|V_Q\|_{\infty}}{\beta + n_{\ell}(y) } +\frac{n_{\ell}(y)V_{Q}^T \hat{P}^{O}_{\ell,y} }{\beta + n_{\ell}(y) }.
\end{eqnarray*}
By equation \eqref{eq: RLSVI Bellman}, we find
\[
F_{\ell, t}Q(x,a)- \E[ F_{\Mc, t}Q(x,a) | \hist, x^{\ell}_{1}, a_{1}^{\ell},..,x^{\ell}_{t}, a^{\ell}_{t} ] \leq \frac{\beta(\|\bar{\theta}\|_{\infty}+\|V_{Q}\|_{\infty})}{\beta+n_{\ell}(y)}  + w_{\ell}(y).
\]
Then,
\begin{eqnarray*}
\E\left[ \Delta_{\ell} \right] &\leq& \E\left[ \sum_{t=0}^{H-1} (F_{\ell, t}Q_{\ell,t+1}-F_{\Mc,t }Q_{\ell,t+1})(x^{\ell}_t, a^{\ell}_t)\right] \\
&=& \E\left[ \sum_{t=0}^{H-1} ((F_{\ell, t}Q_{\ell,t+1})(x^{\ell}_t, a^{\ell}_t) - \E[F_{\Mc,t }Q_{\ell,t+1}(x^{\ell}_t, a^{\ell}_t) \mid \hist, x^{\ell}_{1}, a_{1}^{\ell},..,x^{\ell}_{t}, a^{\ell}_{t} ])\right]  \\
&\leq& \E\left[ \sum_{t=0}^{H-1}  \frac{\beta (\|\bar{\theta}\|_{\infty}+\|V_{Q_{\ell, t+1}}\|_{\infty})}{\beta+n_{\ell}(t, x^{\ell}_t, a^{\ell}_t)}+ w_{\ell}(t, x^{\ell}_t, a^{\ell}_t)  \right]
\end{eqnarray*}
where the second inequality uses that $Q_{\ell, t+1}$ and $F_{\Mc, t}$ are independent conditioned on $\Hc_{\ell t}$. Summing over episodes $\ell\in \{1,..,L\}$ implies
\[
\E \sum_{\ell=1}^{L} \Delta_{\ell} \leq \E \left[\beta \left(\|\bar{\theta}\|_{\infty}+ \max_{\ell\leq L, t< H} \|V_{Q_{\ell,t+1}}\|_{\infty}\right) \sum_{t<H, \ell\leq L} \frac{1}{\beta+n_{\ell}(t, x^{\ell}_t, a^{\ell}_t)} + \sum_{\ell\leq L, t\leq H} w_{\ell}(t, x^{\ell}_t, a^{\ell}_t) \right].
\]
Each term can be bounded separately.
By Corollary \ref{cor: bound on expected noise}
\begin{eqnarray*}
\E \sum_{\ell\leq L, t< H} w_{\ell}(t, x^{\ell}_t, a^{\ell}_t) \leq \E \sum_{\ell\leq L, t< H} \sigma_{\ell}(t, x^{\ell}_t, a^{\ell}_t) \sqrt{2\log(|\Xc||\Ac|)} &=& \E \sum_{\ell\leq L, t< H} \sqrt{\frac{2v \log(|\Xc||\Ac|)}{\beta+n_{\ell}(t, x^{\ell}_t, a^{\ell}_t)}} \\
&\overset{(a)}{\leq}& 2\sqrt{2v H^2 |\Xc|\Ac|L \log(|\Xc||\Ac|) }  \\
& = & 2H^2 \sqrt{6|\Xc|\Ac|L \log(|\Xc||\Ac|) }
\end{eqnarray*}
where the second to last inequality is proved in Lemma \ref{lem: bound on sums}, provided below.
The other term can be bounded as,
\begin{eqnarray*}
&& \E \left[\beta \left(\|\bar{\theta}\|_{\infty}+ \max_{\ell\leq L, t\leq H} \|V_{Q_{\ell,t+1}}\|_{\infty}\right) \sum_{t\leq T, \ell\leq L} \frac{1}{\beta+n_{\ell}(t, x^{\ell}_t, a^{\ell}_t)}\right] \\
&\overset{(b)}{\leq}& \beta\left(\|\bar{\theta}\|_{\infty}+\E \left[ \max_{\ell\leq L, t\leq H} \|V_{Q_{\ell,t+1}}\|_{\infty}\right]\right) H|\Xc||\Ac| \log\left(1+ \frac{L}{|\Xc||\Ac|}\right) \\
& \overset{(c)}{\leq} & \beta\left(H+2H+H^2\sqrt{2\log(1+|\Xc| |\Ac| HL) }\right) H|\Xc||\Ac| \log\left(1+ \frac{L}{|\Xc||\Ac|}\right) \\
& \leq & 5\beta H^3 |\Xc|||\Ac| \sqrt{\log_{+}(1+|\Xc| |\Ac| HL) }\log\left(1+ \frac{L}{|\Xc||\Ac|}\right)
\end{eqnarray*}
where the bound on the sum in inequality (b) is from Lemma \ref{lem: bound on sums} (proof in the Appendix), and inequality (c) applies Corollary \ref{cor: bound on sampled value function}.

\begin{lemma}\label{lem: bound on sums}
   If $\beta\geq 2$ then with probability 1,
  \[
  \sum_{\ell\leq L}\sum_{t\leq H}  \frac{1}{\beta+n_{\ell}(t, x^{\ell}_t, a^{\ell}_t)} \leq H|\Xc||\Ac| \log\left(1+\frac{L}{|\Xc||\Ac|}\right)
  \]
  and
  \[
  \sum_{\ell\leq L}\sum_{t\leq H}  \sqrt{\frac{1}{\beta+n_{\ell}(t, x^{\ell}_t, a^{\ell}_t)}} \leq 2\sqrt{H^{2}|\Xc||\Ac|L}.
  \]
\end{lemma}
Together, the calculations above yield the regret bound
\[
\E \sum_{\ell=1}^{L} \Delta_{\ell} \leq 5\beta H^3 |\Xc|||\Ac| \sqrt{\log_{+}(1+|\Xc| |\Ac| HL) }\log\left(1+ \frac{L}{|\Xc||\Ac|}\right) + 2H^2 \sqrt{6|\Xc|\Ac|L \log(|\Xc||\Ac|) }.
\]
Unfortunately, this alone does not yield the desired bound of order $\tilde{O}(H^2 \sqrt{|\Xc| |\Ac| L})$.
To complete the proof, we consider two cases. First suppose $L\geq 25\beta H^2 |\Xc| |\Ac|$.
Then
\begin{eqnarray*}
\E \sum_{\ell=1}^{L} \Delta_{\ell} & \leq & H^2\sqrt{ |\Xc|\Ac|L \log_{+}(1+|\Xc||\Ac|HL) }\left(2\sqrt{6}+ 5\beta H\sqrt{|\Xc||\Ac|/L} \log\left(1+\frac{L}{|\Xc||\Ac|}\right) \right)    \\
&\leq & H^2 \sqrt{|\Xc||\Ac| L \log_{+}(1+|\Xc||\Ac|HL)}\left(2\sqrt{6}+\sqrt{\beta}\log_{+}\left(1+ \frac{L}{|\Xc||\Ac|}\right)\right)\\
&\overset{(*)}{\leq} & 6H^2 \sqrt{\beta |\Xc||\Ac| L \log_{+}(1+|\Xc||\Ac|HL)}\log_{+}\left(1+ \frac{L}{|\Xc||\Ac|}\right)
\end{eqnarray*}
which is the desired bound.
When $L \leq 25H^2 |\Xc| |\Ac|$, we use the naive bound
\[
\E \sum_{\ell=1}^{L} \Delta_{\ell} \leq HL \leq H \sqrt{L}\sqrt{ 25H^2|\Xc||\Ac|} = 5H^2 \sqrt{\beta |\Xc| |\Ac| L},
\]
which is also less than the term in $(*)$.
This completes the proof of Theorem \ref{thm: regret}.
\end{proof}


\section{Computational studies}
\label{se:computation}

In Section \ref{sec: bounds} we established formal guarantees for a tabular version of RLSVI.  This result serves as a sanity check, demonstrating
that RLSVI carries out efficient deep exploration, but the tabular nature and prior structure of the setting limits the scope of our theoretical results.
Perhaps most importantly, the results do not apply when parameterized representations are used to generalize across states and actions.
In this section, we present computational results that offer further assurances.  In particular, we discuss results from a series of experiments designed to
enhance insight into the workings of RLSVI beyond the scope of our theoretical analysis.
The focus of these experiments is to improve understanding, rather than to solve challenging problems.
Nevertheless, we believe that observations from these didactic examples will prove valuable toward the design of practical systems that
require the synthesis of efficient deep exploration with effective generalization.

\subsection{Deep-sea exploration}
\label{sec:compute_deep}

We begin our computational experiments with an empirical study of the ``deep-sea exploration'' problem from Example \ref{ex:grid}.
This offers a simple illustration of the importance of deep exploration.
Although the associated MDP has only $N^2$ states, dithering schemes require a number of episodes that grows exponentially in $N$
to effectively explore the environment.  Deep exploration approaches, on the other hand, can effectively explore the environment within
a sub-exponential number of episodes.  Our results verify the efficacy randomized value functions and that RLSVI
carries out deep exploration.





\subsubsection{Tabular representation}
\label{sec:compute_tabular}

We begin with an investigation into RLSVI with a tabular representation.
Our goal will be to study the behavior of RLSVI in a simple setting similar to that addressed by Theorem \ref{thm: regret}.
To do this, we generate random ``deep-sea'' environments according to Example \ref{ex:grid} and empirically evaluate
performance over many simulations.

We apply RLSVI \texttt{agent} with infinite buffer, greedy actions and \texttt{learn\_grlsvi}\footnote{\texttt{learn\_grlsvi} is \texttt{learn\_rlsvi} (Algorithm \ref{alg:learn_rlsvi}) with additive Gaussian reward noise of variance $v$ \eqref{eq:gauss_rlsvi}} with a tabular representation.
Specifically, each component of the parameter vector $\theta \in \Real^{|\Sc \times \Ac|}$ provides a value estimate for one state-action pair.
We set the tuning parameters to $v = H^2 / 25$, $\overline{\theta} = 0$, and $\lambda = v$.
Note that, compared to the setting specified in Theorem \ref{thm: regret} we rescaled $v$ by a constant in order to accelerate learning in the deterministic deep-sea environment.
Further our prior parameter $\overline{\theta}$ is not optimistic.
This choice is not particularly important in terms of performance on this task, in fact setting $\overline{\theta}=H \Ind$ leads to almost identical results.
The reason we do not rely on an optimistic prior is to highlight the practical efficacy of RLSVI even without optimistic prior.
This will be important in later sections where we study RLSVI in domains with generalization, for which the notion of an universally `optimistic prior' does not carry over from the tabular setting \citep{osband2016rlsvi}.

We compare the performance of RLSVI against two well-studied reinforcement learning algorithms specifically designed to explore efficiently with
tabular representations: UCRL2 \citep{Jaksch2010} and PSRL \citep{Osband2013}.
We similarly modify UCRL2 and PSRL that accelerate learning in the deterministic deep-sea environment\footnote{Specifically, we update the confidence sets for PSRL and UCRL2 as if each observed transition $(s,a,r,s')$ occured identically $10$ times repeatedly. We also further rescale the confidence sets for UCRL2 to be $10$ times smaller than prescribed by the analysis.}.
For each of the algorithms our modifications reduce learning times but do not affect rates at which learning times scale with problem size.

\begin{figure}[!h]
\centering
\includegraphics[width=0.6\textwidth]{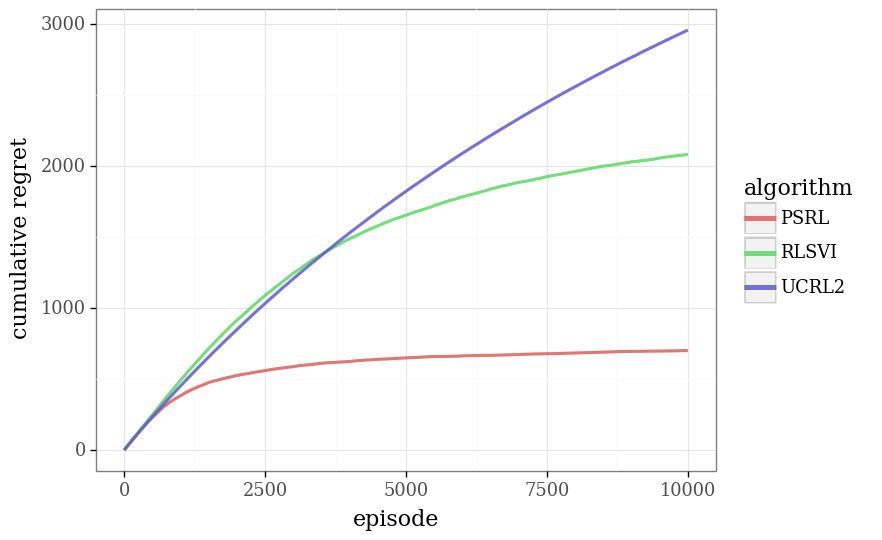}
\vspace{-3mm}
\caption{\small RLSVI is competitive with algorithms designed for tabular exploration ($N=10$).}
\label{fig:tabular_regret}
\end{figure}

Figure \ref{fig:tabular_regret} plots the average regret realized by RLSVI (specifically, \texttt{learn\_grlsvi}), UCRL2 and PSRL, over five seeds with a bomb and five seeds with treasure.
Among the algorithms, PSRL offers the lowest level of regret, followed by RLSVI, and then UCRL2.
Hence, RLSVI is competitive with these algorithms, which are designed to yield efficient exploration with tabular representations.

One natural question is how this performance scales with the size $N$ of the problem.
To answer this we study the ``learning time,'' defined to be the first episode where the average regret per episode is less
than $0.5$.  Formally,
\begin{equation}
  \label{eq: learn_time}
  {\rm Learning\ time}(\mathcal{M}^*, {\rm alg}) := \min \left\{L > 1 \left\mid \frac{{\rm Regret}(\mathcal{M}^*, {\rm alg}, L)}{L} \le 0.5 \right. \right\},
\end{equation}
This quantity is random, as it depends on the realization of $\mathcal{M}^*$.
For an algorithm with regret bound ${\rm Regret}(\mathcal{M}^*, {\rm alg}, L) \le \sqrt{B L}$ we would expect the learning time to be $\tilde{O}(B)$.

\begin{figure}[!htpb]
\centering
\includegraphics[width=0.6\textwidth]{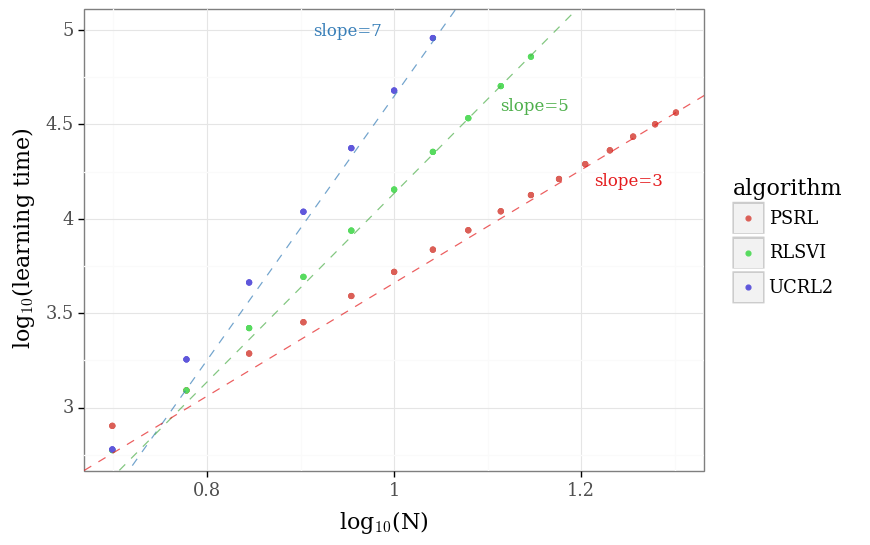}
\caption{Scaling with tabular learning.}
\label{fig:tabular_scale}
\end{figure}

The results of Theorem \ref{thm: regret} suggests an $\tilde{O}(\sqrt{H^3 S AL})$ \textit{average} scaling when the environment is drawn from a symmetric Dirichlet distribution.
We can contrast this to existing performance guarantees for UCRL2 which, when adapted to this setting, provide a $\tilde{O}(\sqrt{H^3 S^2 AL})$ regret bound.
For the deep-sea problem, $H=N$, $S = N^2$ and $A=2$, and the bounds therefore suggests that learning times scale as $\tilde{O}(N^5)$ for RLSVI and $\tilde{O}(N^7)$ for UCRL2.
Figure \ref{fig:tabular_scale} shows that observed performance to a large degree matches performance suggested by these theoretical results.  The best known bound for PSRL also
suggests a $\tilde{O}(N^5)$ scaling. However, recent work suggests that this bound is loose \citep{osband2016posterior,osband2016lower}, and the associated plot in Figure \ref{fig:tabular_scale} strengthens the case for that hypothesis.

\subsubsection{Linearly parameterized value functions}
\label{sec:compute_coherent}

Section \ref{sec:compute_tabular} presents evidence of the efficacy of RLSVI with a tabular representation.
However, the value of RLSVI lies in its ability to function well with parameterized value functions that generalize across states and actions.
Model-based algorithms such as UCRL2 or PSRL do not accommodate this form of generalization.

In this subsection, we continue our investigation of the ``deep-sea'' environment, but now using linear parameterized representations.
To do this, we generate a random subspace of dimension $D$ that is specifically designed to include the true optimal value function of the deep-sea environment irrespective of whether there is treasure or a bomb.
We then generate a random basis of $D$ unit vectors $\phi_{1}, \ldots, \phi_{D} \in \Re^{|\mathcal{S}| |\mathcal{A}|}$ that span the space.
Each vector $\phi_{d}$ can be thought of as representing a feature that assigns a numerical value to each state-action pair.
As such, the representation can be thought of as a linear combination of features, with a dimensional parameter vector $\theta \in \Re^{D}$ encoding feature weighs: $\tilde{Q}_{\theta } = \sum_{d=1}^D \theta_{d} \phi_{d}$.

\begin{figure}[!htpb]
\centering
\includegraphics[width=0.5\textwidth]{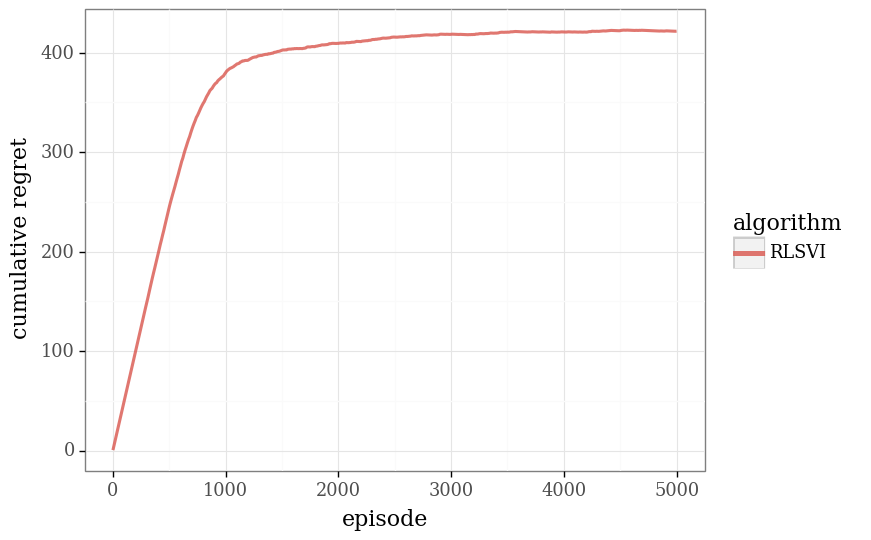}
\vspace{-4mm}
\caption{Regret with $N=50$ and $M=50$}
\label{fig:coherent_reg}
\end{figure}

To facilitate efficient computation in the deep-sea problem we restrict these $D$-dimensional features so that each is nonzero only at
state-action pairs corresponding to one row of the grid of states.
We only consider values of $D$ that are multiples of $N$, and for each row assign $M=D/N$ features to generate nonzero values.
With this representation, RLSVI learns a separate $M$-dimensional representation for each row, avoiding a costly dimension $D$ inversion.
Figure \ref{fig:coherent_reg} plots realized regret generated by \texttt{learn\_grlsvi} with $\lambda = 100$, $v = 0.01$, and $N=50$.
Once again, we simulate this problem for five random seeds with treasure and five random seeds with a bomb and report the average regret.
These results demonstrate that per-episode regret vanishes much faster than any dithering method, which would expect at least $2^{50} \simeq 10^{15}$ episodes to even reach the chest!


\begin{figure}[!h]
\centering
\begin{subfigure}{.5\textwidth}
  \centering
  \includegraphics[width=.9\linewidth]{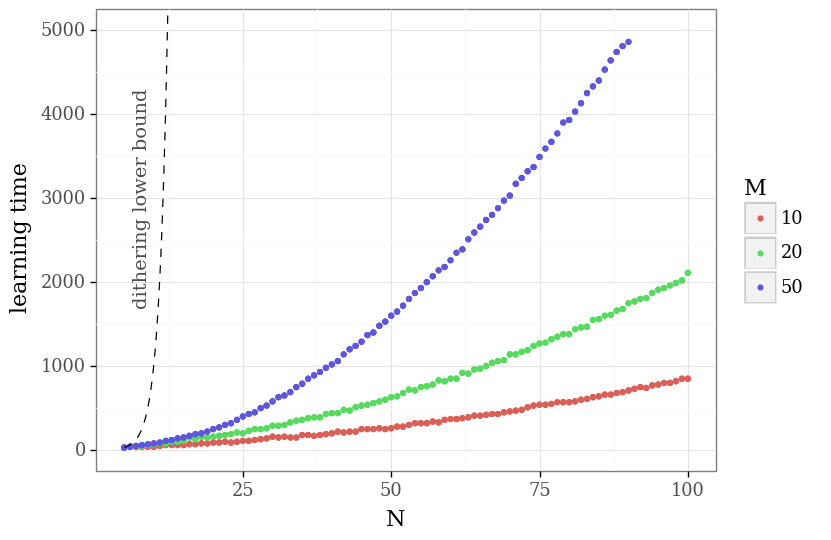}
  \vspace{-2mm}
  \caption{Raw values}
  \label{fig:scale_len}
\end{subfigure}%
\begin{subfigure}{.5\textwidth}
  \centering
  \includegraphics[width=.9\linewidth]{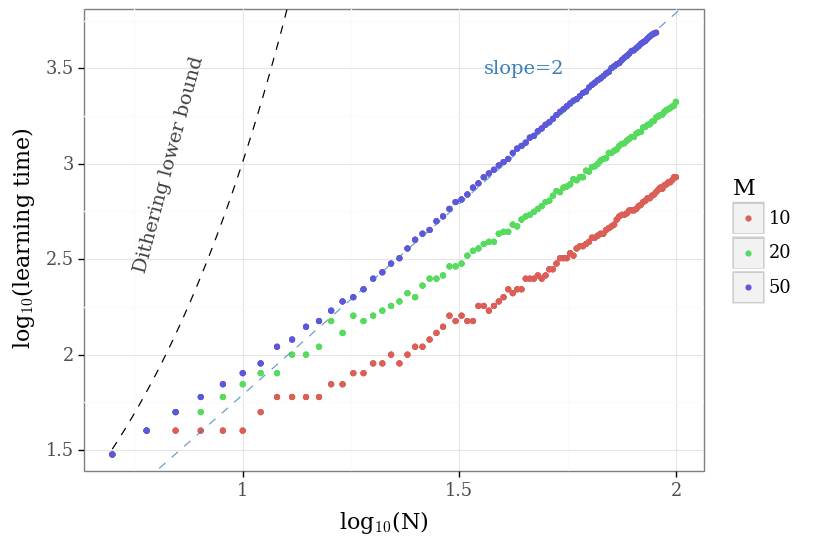}
  \vspace{-2mm}
  \caption{Log-log plot shows scaling}
  \label{fig:scale_len_ln}
\end{subfigure}
\caption{Effect of problem size $N$ on learning time.}
\label{fig:coherent_scale_len}
\end{figure}

It is the cases with treasure rather than a bomb that bind regret and learning times.
For this reason we will only present results associated with the former case from here on to save on computation.
Figure \ref{fig:coherent_scale_len}(a) plots learning times as a function of $N$
for different numbers of features $M$ per row.
As one would expect, the learning time increases with the number of features.
Importantly, this scaling with chain length $N$ is graceful and grows much more slowly than even the lower bound for dithering methods $O(2^N)$.
Figure \ref{fig:coherent_scale_len}(b) plots the same data on a log-log scale to highlight this sub-exponential growth.
We can see empirically that the slope on this scale is approximately two, implying that learning time scales approximately quadratically in $N$.

\begin{figure}[!htpb]
\centering
\includegraphics[width=0.6\textwidth]{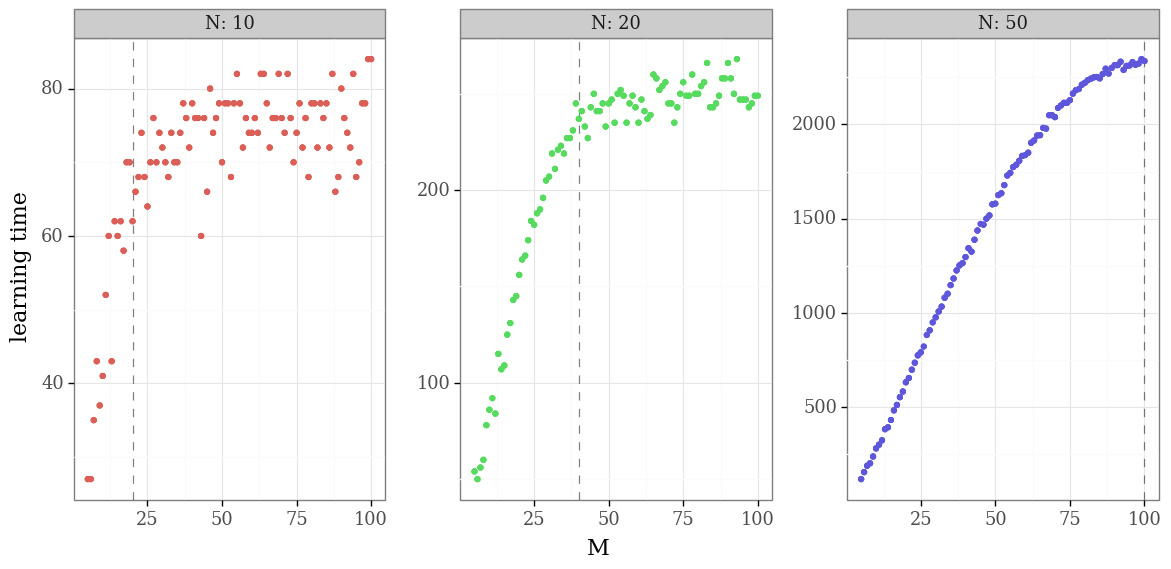}
\vspace{-4mm}
\caption{Scaling with number of features.}
\label{fig:coherent_feat}
\end{figure}

For another perspective on scaling, Figure \ref{fig:coherent_feat} presents plots of learning times as a function of the number of features $M$, for several values of $N$.
In each case, the learning time appears to grow linearly in the number of features up until some threshold and then increase much more slowly beyond this point.
The vertical dotted lines in Figure \ref{fig:coherent_feat} appear at $M=2N$.
Empirically, this seems to be the point beyond which the incremental learning time incurred with additional features is small.
Intuitively, one might speculate that this is reasonable because $2N$ is equal to
the maximum number of states-action pairs which can be observed in any time period.
Beyond this point, additional features must be linearly dependent.



\subsubsection{Misspecified representations}
\label{sec:compute_agnostic}

Let us now consider a more realistic setting in which the value function representation is mis-specified in the sense that
$Q^*$ is not equal to $\tilde{Q}_\theta$ for any vector $\theta$.  We experiment with a setting completely analogous to that
of the previous section, except we add to each feature vector $\phi_d$ a random vector $\eta_d \in \Re^{|\mathcal{S}| |\mathcal{A}|}$.
The random vector $\eta_d$ is nonzero only only at state-action pairs associated with the feature $\phi_d$.  Each nonzero
noise component is sampled from  $N(0, \psi I)$.  Hence, we make use of a representation of the form
$\tilde{Q}_\theta = \sum_{d=1}^D \theta_d (\phi_d + \eta_d)$.  As the parameter $\psi$ increases, the representation
becomes increasingly misspecified.

\begin{figure}[!htpb]
\centering
\includegraphics[width=0.5\textwidth]{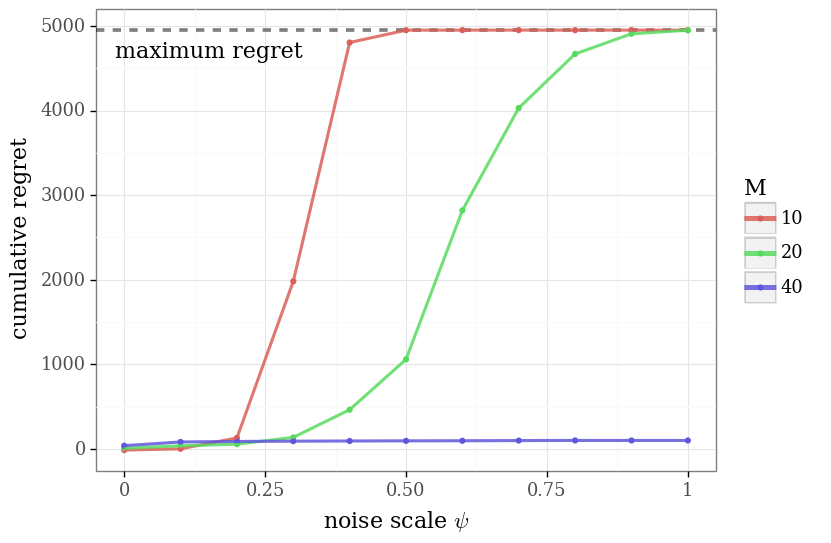}
\vspace{-4mm}
\caption{Robustness to misspecification with $N=20$.}
\label{fig:agnostic}
\end{figure}

Figure \ref{fig:agnostic} plots cumulative regret of \texttt{learn\_grlsvi} with varying numbers of features and degrees of misspecification over 5000 episodes.
Our results are the average of 20 seeds for each value of the noise scale $\psi$.
These results indicate that RLSVI remains robust to some degree of misspecification.
However, at some point the model-misspecification becomes too severe as the value of $\psi$ increases depending upon the number of basis functions $M$.
The power of the representation increases with the number of features, and this enables RLSVI to tolerate larger values of $\psi$.
In the case $M \ge 2N$ random basis functions will span the true value function with high probability.
As expected, we observe that for $M=40, N=20$ RLSVI performs similarly well irrespective of $\psi$.

\subsubsection{Parameter tuning}

The computational results we have present in Sections \ref{sec:compute_coherent} and \ref{sec:compute_agnostic} make use of particular settings for the prior and noise variance parameters $\lambda = 100, v = 0.01$.
In this section, we study the dependence of results on these parameter settings.
The deep-sea problem we have considered is in some sense degenerate because each problem instance is deterministic.
In order to offer a more representative set of results pertaining to variance parameter tuning, we will also consider a modified version of the deep sea problem where all reward observations are corrupted by some $N(0,1)$ noise.

\begin{figure}[!htpb]
\centering
\vspace{-2mm}
\includegraphics[width=0.7\textwidth]{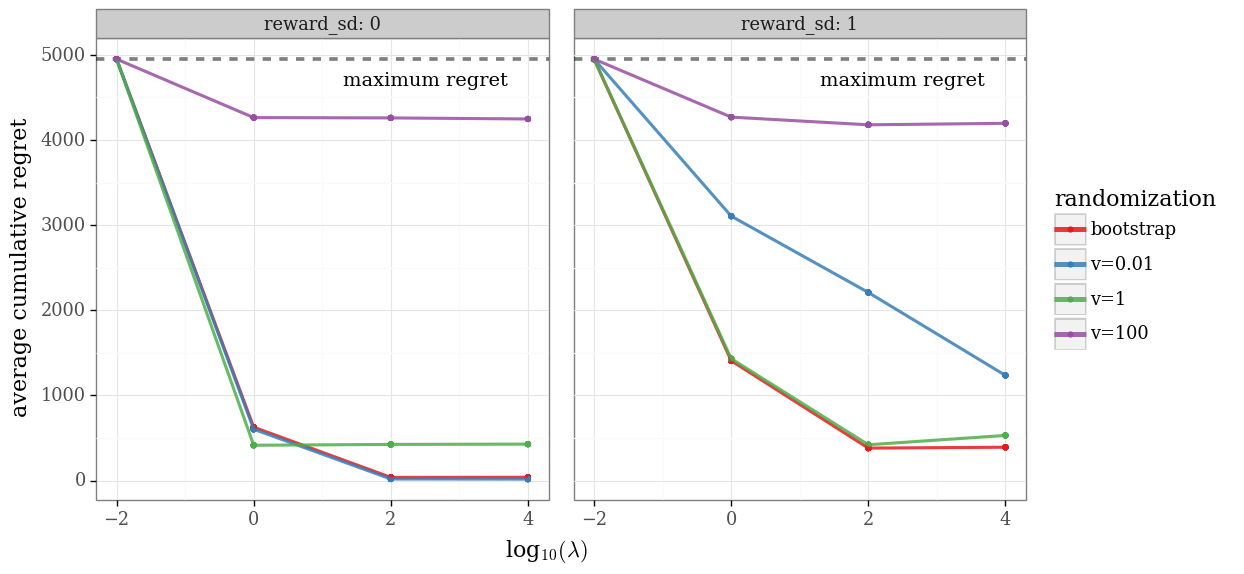}
\vspace{-4mm}
\caption{Robustness to prior and noise variance parameters.}
\vspace{-2mm}
\label{fig:lambda_param}
\end{figure}

Figure \ref{fig:lambda_param} plots the cumulative average regret after 5000 episodes over 10 random seeds for various choices of prior and noise variance with $N=20$ and $M=10$.
In both settings with and without stochastic rewards, and for all choices of noise randomization, we can see that prior variance which is too small can prohibit learning.
In this problem, where our prior $\overline{\theta} = 0$ is not informative, choosing even very large $\lambda$ does not degrade performance.

\begin{figure}[!htpb]
\centering
\vspace{-2mm}
\includegraphics[width=0.7\textwidth]{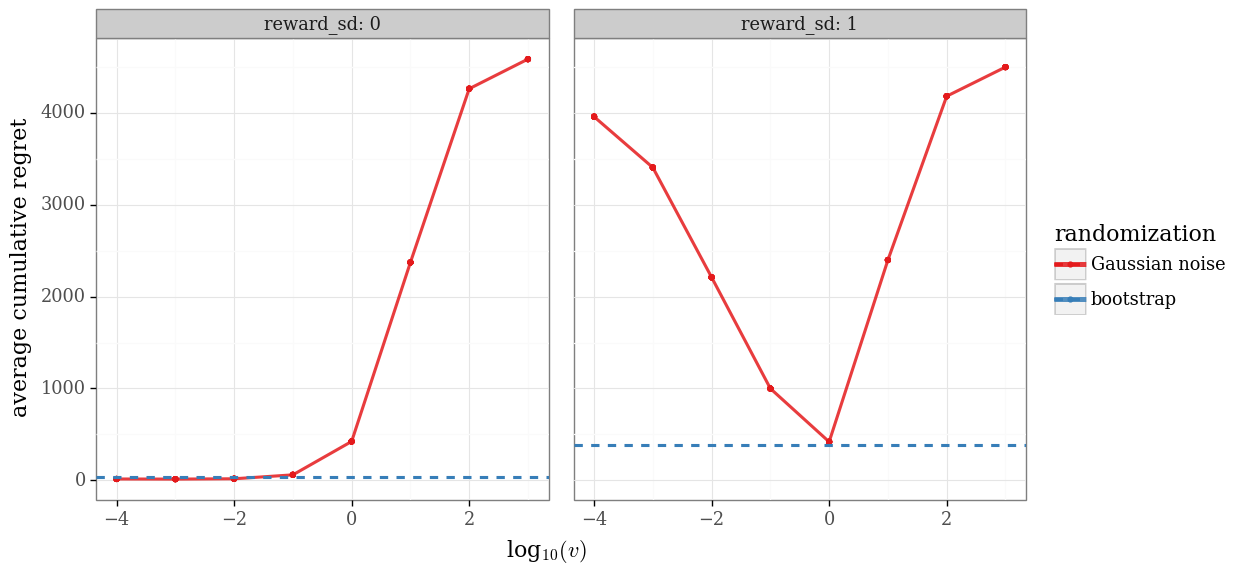}
\vspace{-4mm}
\caption{Bootstrap is competitive with the best choice of $v$ across levels of noise.}
\vspace{-2mm}
\label{fig:v_param}
\end{figure}

Figure \ref{fig:v_param} takes the same data as Figure \ref{fig:lambda_param} but investigates the sensitivity of RLSVI to the noise randomization, for the choice $\lambda=100$.
We see that choice of the best-performing noise variance $v$ is largely dependent upon the scale of the noise in the actual environment.
When the underlying environment is deterministic there is no benefit to adding noise and low values of $v$ perform best.
However, when the environment is stochastic choosing $v$ on the order of the variance of the noise in the environment is necessary to not fall victim to unlucky observations.
In both settings, bootstrapping performs competitively with the ex-ante ``best'' choice of $v$ but does not need to be specified in advance.

\begin{figure}[!htpb]
\centering
\includegraphics[width=0.7\textwidth]{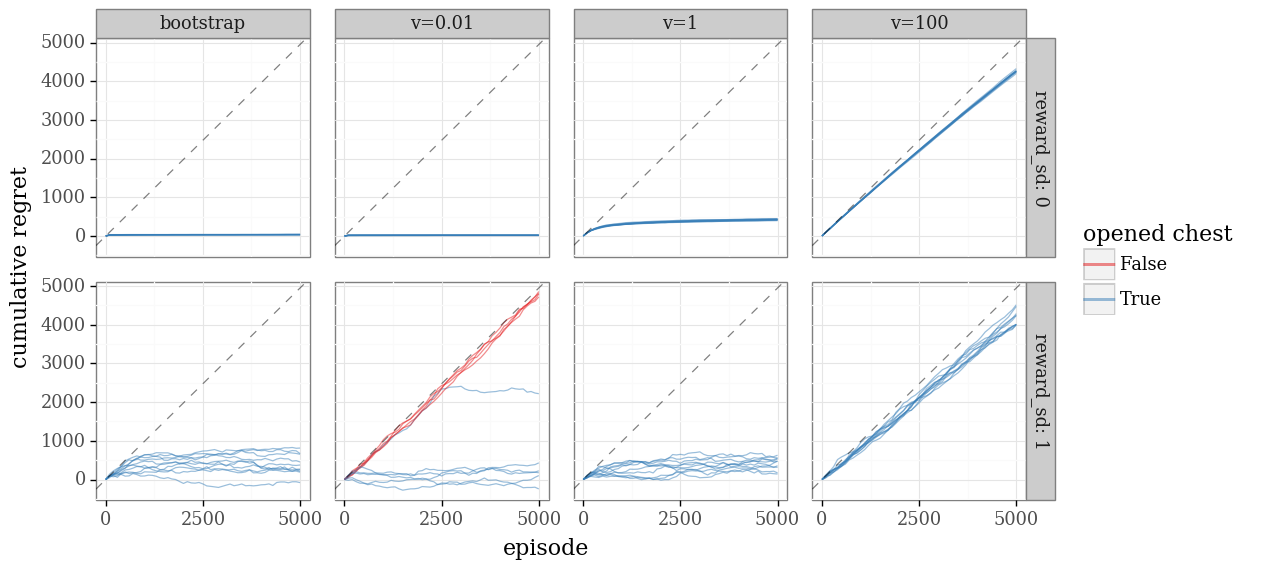}
\vspace{-4mm}
\caption{Higher $v$ is more robust to stochastic environments but learns more slowly.}
\label{fig:reg_param}
\end{figure}

To gain some more intuition for this parameter tuning we take this same data as Figure \ref{fig:v_param} and  present the realized regret by random seed in Figure \ref{fig:reg_param}.
We see $v$ smaller than the noise in the problem can lead to premature and sub-optimal convergence that never opens the chest (and so leads to linear regret).
Choices of $v$ which are too large lead to slower learning and more exploration, but do not lead to linear regret.
We note that RLSVI with $v > 0$ does not seem to significantly degrade in performance for stochastic rewards with variance up to $v$.
Once again, we see that randomization by bootstrap is competitive with the best ex-ante choice of $v$ but with one fewer parameter to tune.

\begin{figure}[!h]
\centering
\begin{subfigure}{.5\textwidth}
  \centering
  \includegraphics[width=.6\linewidth]{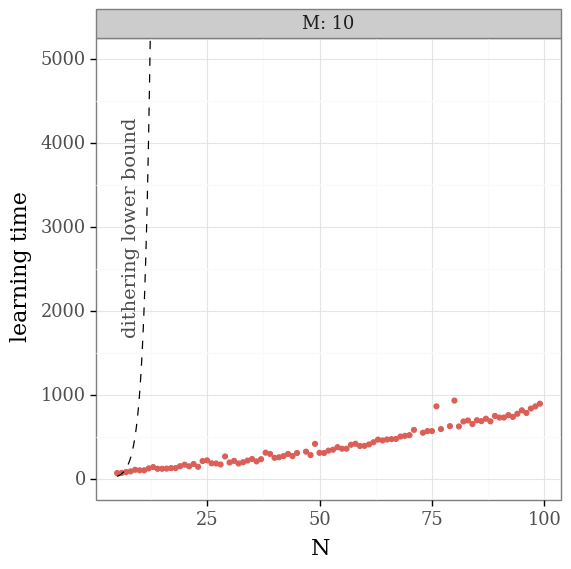}
  \vspace{-2mm}
  \caption{Scaling with $N$}
  \label{fig:boot_len}
\end{subfigure}%
\begin{subfigure}{.5\textwidth}
  \centering
  \includegraphics[width=.6\linewidth]{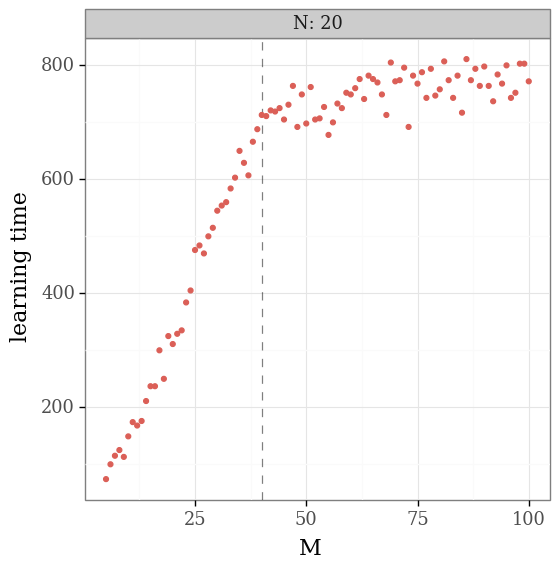}
  \vspace{-2mm}
  \caption{Scaling with number of features.}
  \label{fig:boot_feat}
\end{subfigure}
\caption{Performance of the bootstrap scales similarly to \texttt{learn\_grlsvi}.}
\label{fig:coherent_boot}
\end{figure}

The bootstrap learns the ``right'' noise variance, and beyond that, can even learn how it should vary over states and actions.
Further, Figure \ref{fig:coherent_boot} plot learning times from applying the \texttt{learn\_brlsvi} in the same settings to which \texttt{learn\_grlsvi} was applied to generate Figures \ref{fig:coherent_reg} and \ref{fig:coherent_scale_len}.
These results suggest that learning times of \texttt{learn\_brlsvi} scale similarly to those of \texttt{learn\_grlsvi}.
In general, our results suggest that bootstrapping offers a natural approach to setting an appropriate form of randomizing noise without prior knowledge.

\subsection{Deep exploration with deep learning}

The experiments of Section \ref{sec:compute_deep} are designed to highlight several key properties of RLSVI in a simple setting.
These results demonstrate that RLSVI can successfully synthesize efficient exploration with generalization.
However, the context was a ``toy'' example in that the underlying system involved a tractable number of states.
Further, the algorithms we evaluate in that section are not practical for large-scale learning problems for two reasons.
First, they recompute the whole history of data each episode and so have computational costs that grow with the amount of data collected.
Second, their performance is highly reliant upon an accurate linear basis for the value function whereas many of the recent breakthroughs in the field have come from so-called ``deep'' RL that uses deep neural networks for function approximation in RL.

In this section we present computational results for deep exploration with a practical variant \texttt{learn\_ensemble\_rlsvi} together with neural network function approximation.
We begin with more experiments on the ``deep sea'' problem and show that a parallel RLSVI strategy can recover performance qualitatively similar to full batch resampling, but at a dramatically lower computational cost.
Next we investigate the scaling properties of RLSVI with nonlinear neural network representations and find that the performance can successfully synthesize exploration with generalization in this setting.
Finally, we apply these findings to a difficult task in continuous control.
We find that \texttt{learn\_ensemble\_rlsvi} successfully demonstrates deep exploration together with complex nonlinear generalization.

\subsubsection{RLSVI via ensemble sampling}
\label{sec:compute_parallel}

In this section we apply \texttt{learn\_ensemble\_rlsvi} (Algorithm \ref{alg:learn_ensemble_rlsvi}) for $K=1,5,10, 20, 40$ and with an $\mathtt{ensemble\_buffer}$ that stores the most recent $10^{5}$ transitions.
For $\mathtt{update}$ we use $\mathtt{update\_bootstrap}$ (Algorithm \ref{alg:ensemble_update_bootstrap}) to approximate a ``double or nothing'' online bootstrap \citep{owen2012bootstrapping}.
We use a discounted TD loss with $\gamma=0.99$, learning rate $\alpha=10^{-3}$ and minibatch size of $128$.
For our value function family $\Qc$ we consider two-layer MLP with 50 rectified linear units in each layer.

In place of explicit prior regularization $\Rc$ we evaluate $\tilde{Q}_{\theta_k} = f^{\rm MLP}_{\theta_k} + f^{\rm MLP}_{\theta^0_k}$ where $f^{\rm MLP}$ is a 2-layer MLP and the parameters $\theta_k, \theta^0_k$ are sampled independently from Glorot initialization and henceforth $\theta^0_k$ is held fixed \citep{glorot2010understanding,osband2018prior}.
We found that this randomization plus SGD training provide sufficient regularization for deep learning without use of weight decay \citep{zhang2016understanding,bartlett2017spectrally}.
Using two separate networks, one with fixed weights, is useful so that the SGD training cannot easily learn to ignore the state input and learn $Q=0$ as a degenerate global solution.
More detail on this specific prior mechanism for deep neural networks is available in \citet{osband2018prior}.

\begin{figure}[!htpb]
\centering
\includegraphics[width=0.6\textwidth]{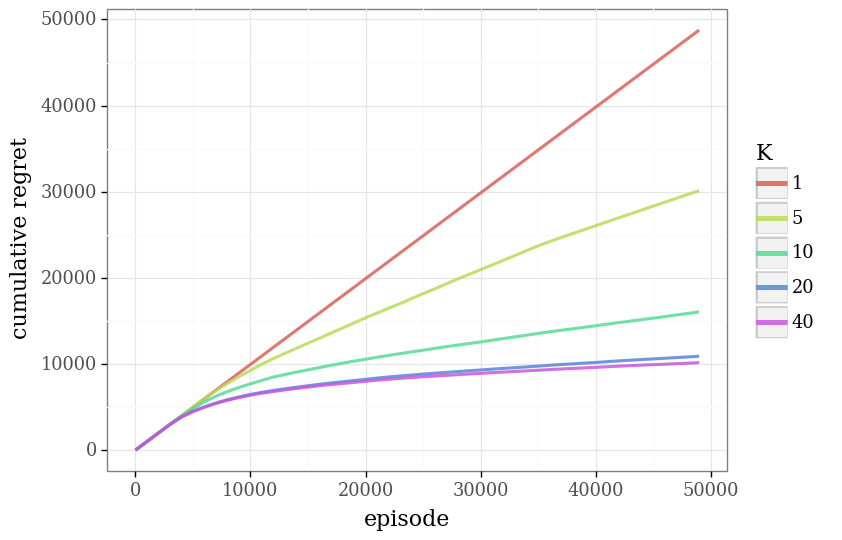}
\vspace{-4mm}
\caption{Investigating ensemble size $K$ for \texttt{learn\_ensemble\_rlsvi} on deep sea $N=25$.}
\label{fig:ensemble_25}
\end{figure}

We apply this algorithm to ``deep sea'' problem with a raw pixel representation in $[0,1]^{N \times N}$ where the diver's position is given by a $1$ and all other entries zero.
This problem is by nature tabular and does not necessitate complex representation via neural network however we use it as a simple example to investigate the importance of ensemble size $K$.
Figure \ref{fig:ensemble_25} presents the performance for $N=25$ averaged over 20 seeds.
We see that even a relatively small number of parallel estimates $K$ can direct deep exploration and that, for a problem of this size, the marginal benefits seem to plateau around $K=20$.
The computational savings of this parallel approach can be quite significant.
We compare the cost of $K=20$ parallel $Q$-networks each with $O(1)$ computation per episode against a naive \texttt{learn\_brlsvi} that computes $H=25$ value functions each with $\Omega(L)$ computation per episode.

\begin{figure}[!htpb]
\centering
\hspace{45mm}
\includegraphics[width=0.9\textwidth]{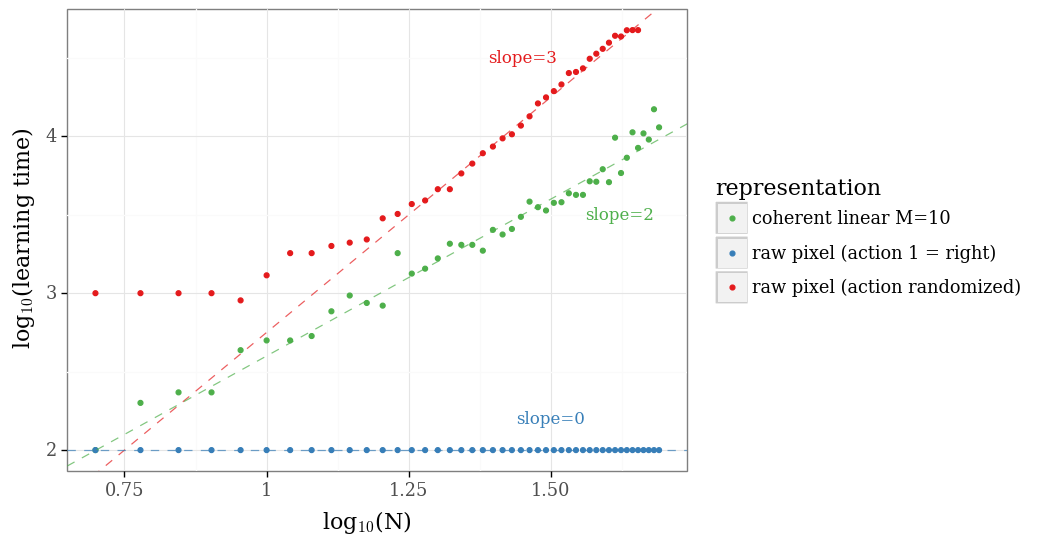}
\vspace{-4mm}
\caption{Log-log plot shows empirical scalings with different feature representations.}
\label{fig:ensemble_scale}
\end{figure}

Our next set of experiments investigates the scalability of \texttt{learn\_ensemble\_rlsvi} with different feature representations.
We repeat the experiment of Figure \ref{fig:ensemble_25} with $K=20$ varying $N=5,..,50$ under three separate representations and averaged over $20$ seeds.
First we consider the raw pixel representation as above; this is effectively a tabular problem.
Next, we consider the informative linear basis of Section \ref{sec:compute_coherent} with $M=10$; this learning can be expedited through generalization.
Finally, we consider the raw pixel representation but alter the problem formulation so that the action 1 is always ``go right''; the optimal value function for this setting takes a particularly simple form and an optimal policy can easily be happened upon through random weight initialization and without any learning.
Figure \ref{fig:ensemble_scale} shows that RLSVI with neural network architecture can exploit these feature representations where they are present, but defaults to an approximately tabular learning approach when they are not.

\subsubsection{Cartpole swing up}
\label{sec:compute_cartpole}

In this section we consider the classic ``cartpole'' problem of a cart attached to a pole on a frictionless rail.
We modify the problem so that, as in previous sections, deep exploration is crucial to finding rewarding states and thus learning the optimal policy.
However, unlike ``deep sea'' the underlying dynamics are not governed by a small finite MDP.
The cart is of mass $M=1$ and the pole is mass $m=0.1$ and length $l=1$, with acceleration due to gravity $g=9.8$.
At each timestep the agent can apply a horizontal force $F_t$ to the cart.
The dynamics for this system are given by a second order differential equation in $x_t$, the horizontal position of the cart and $\theta_t$, the angle of the pole from vertically upright at $\theta = 0$,
\begin{eqnarray}
\label{eq:cp_dynamics}
  \tau_t = \frac{F_t + \frac{l}{2} \dot{\theta_t}^2 \sin(\theta_t)}{m + M}, \
  \ddot{\theta_t} = \frac{g \sin(\theta_t) - \cos(\theta_t)\tau_t}{\frac{l}{2} \left(\frac{4}{3} - \frac{m}{m+M} \cos(\theta_t)^2 \right)}, \ 
  \ddot{x} = \tau - \frac{m \frac{l}{2} \ddot{\theta_t} \cos(\theta_t)}{m + M}.
\end{eqnarray}

Unlike the traditional cartpole problem, where the agent begins with the pole stood upright and must learn to balance it; our agent begins each episode with the pole hanging down and has to learn to swing it up.
Concretely we interact with the environment through the state $s_t := (\theta_t, \dot{\theta}_t, x_t, \dot{x}_t, t) \in \Real^5$.
Each episode begins with $s_0 = (\pi, 0, 0, 0) + w$ for $w_i \sim {\rm Unif}([-0.05, 0.05])$ i.i.d. in each component.
We discretize the evolution of \eqref{eq:cp_dynamics} with timescale $\Delta t = 0.01$ and present the choice of actions $F_t \in \{-10, 0, 10\}$ for all $t$.
The reward structure of this task is specifically designed to necessitate deep exploration; each timestep the agent pays a cost $\frac{|F_t|}{1000}$ for its action but can receive a reward of $1$ if the pole is balanced upright and steady in the middle\footnote{Reward $+1$ received if $\cos(\theta) > 0.95$ with all of $|\dot{\theta}|, |x|, |\dot{x}| \le 1$.}.
The ends of the rail at $x=-5, 5$ are rigid and immovable; an episode ends whenever $t > 10$.

\begin{figure}[!htpb]
\centering
\includegraphics[width=0.65\textwidth]{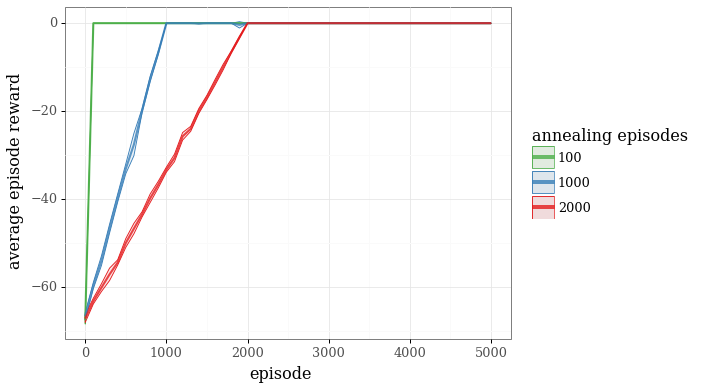}
\vspace{-4mm}
\caption{DQN with $\epsilon$-greedy exploration simply learns to stay motionless.}
\label{fig:cartpole_dqn}
\end{figure}

Figure \ref{fig:cartpole_dqn} presents results for DQN and a 50-50-MLP with rectified linear units with linear annealing $\epsilon$-greedy dithering from $1$ to $0$ over varying number of episodes.
Irrespective of annealing schedule, DQN is unable to gather informative data since it does not perform deep exploration.
In this environment, dithering strategies for exploration are insufficient to gain information beyond the locally-attractive policy to remain motionless and receive a reward of zero.

\begin{figure}[!htpb]
\centering
\includegraphics[width=0.65\textwidth]{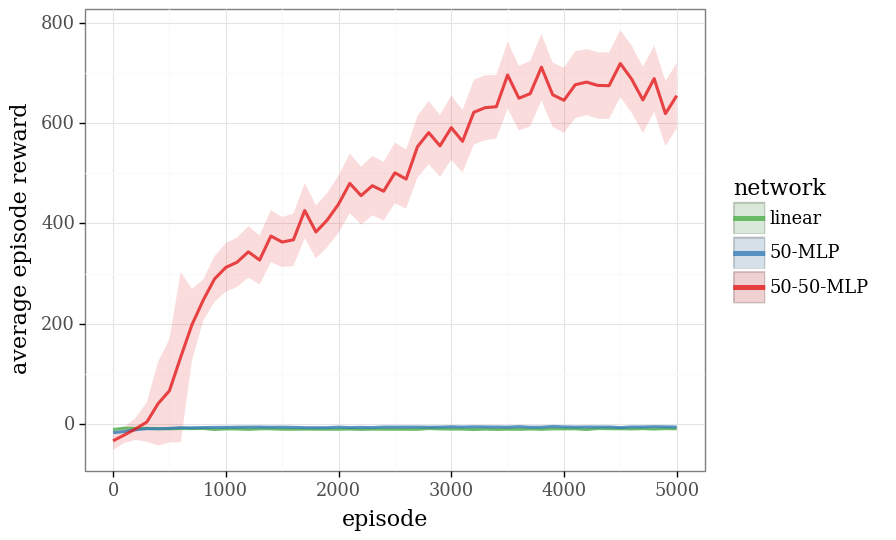}
\vspace{-4mm}
\caption{RLSVI with 2-layer neural network is able to learn a near-optimal policy.}
\label{fig:cartpole_boot}
\end{figure}

Figure \ref{fig:cartpole_boot} presents the average episodic reward for an ensemble approach to RLSVI with $K=20$ and the same algorithmic approach as Section \ref{sec:compute_parallel}.
We note that, unlike DQN with $\epsilon$-greedy exploration, RLSVI is able to learn a successful swing up policy with an identical network architecture.
In addition, we note that RLSVI implemented with linear or single-layer $Q$-value functions is unable to learn a successful swing up policy.
This demonstrates the importance of \textit{both} deep exploration and deep representation learning in order for a successful application of deep RL in this setting.
The results of Figures \ref{fig:cartpole_dqn} and \ref{fig:cartpole_boot} are averaged over 20 seeds, with confidence intervals at 1 standard error of the mean.
We present visualization of this performance \url{https://youtu.be/ia72VyW5MfI}.

The computational results we present in this paper are tailored to be simple and interpretable, with a clear focus on the importance of deep exploration and the compatibility of this approach with linear and nonlinear value function learning.
Related work investigates scaling up this approach in several arcade games including Tetris, Angry Birds, Atari 2600 as well as a model of recommendation systems \citep{osband2016rlsvi,ibarra2016angrier,osband2016deep}.
We consciously choose to keep our empirical investigation concise and the key results sanitary, but look forward to pushing the boundaries of large-scale applications of ``deep RL'' via randomized value functions in future work.


\section{Closing remarks}

Much of the applied RL literature focuses on simulated systems and learning a good final policy, potentially over billions or trillions of episodes.
Assessed in this manner, performance is driven largely by the investment of computational resources and simulation time; not just how effectively a reinforcement learning algorithm makes decisions and interprets observations.
In many real systems, data collection is costly or constrained by the physical context, and this calls for a focus on statistical efficiency.
In these contexts it may be more appropriate, for example, to evaluate algorithms over a fixed number of episodes.

Exploration is a key driver of statistical efficiency.
As discussed in Section \ref{se:deep}, there can be an exponentially large difference in data requirements between an agent that explores via dithering, as has commonly been done in past applications of reinforcement learning, and an agent that  carries out deep exploration.
In this paper, we have developed randomized value functions as a concept that enables efficient deep exploration in conjunction with value function learning methods commonly used in reinforcement learning.

\subsection*{Acknowledgements}

This work was generously supported by a research grant from Boeing, a Marketing Research Award from Adobe, and Stanford Graduate Fellowships, courtesy of PACCAR, Burt and Deedee McMurty, and ST Microelectronics.  We thank Emma Brunskill, Hamid Reza Maei, and Rich Sutton for helpful discussions and Vikranth Dwaracherla, Xiuyuan Lu, Shuhui Ku, and Kuang Xu for pointing out errors and ambiguities in earlier drafts, and more broadly, students who participated in Stanford University's 2017 and 2018 offerings of Reinforcement Learning, for feedback and stimulating discussions on this work.
We also thank John Aslanides, Albin Cassirer, Alex Pritzel, Charles Blundell and the rest of the team at DeepMind for help with experiments, infrastructure and an engaging work environment.


\newpage
\begin{center}
\textbf{\Large APPENDIX}
\end{center}
\appendix
 
\section{Proofs of technical lemmas}
\subsection{Proof of Lemma \ref{lem: planning to Bellman}}
\begin{lemma*}[Planning Error to Bellman Error]
    Let $Q_0, Q_1, Q_2,...,Q_{H}\in \Re^{|\Xc| |\Ac|}$ be any sequence with $Q_{H}=0\in \Re^{|\Xc| |\Ac|}$ and take $\pi=(\pi_0, \pi_1,...)$ to be the policy $\pi_t(x)= \arg\max_{a\in \Ac} Q_t(x,a)$ for all $a,x$. 
    Then for any MDP $\Mc$ and initial state $x\in \Xc$,
    \begin{equation}\label{eq: planning to Bellman}
    Q_0(x, \pi_{0}(x))- V^{\pi}_{\Mc}(x) = \E_{\Mc, \pi}\left[ \sum_{t=0}^{H} \left((Q_{t}-F_{\Mc, t}Q_{t+1})(x_t, a_t)  \right) | x_0=x)  \right]  
    \end{equation}
\end{lemma*}
\begin{proof}
    Define the operator $F_{\Mc, t}^{\pi}$ at time $t$ for the MDP $\Mc$ and policy $\pi$ by  
    \[ 
    F_{\Mc, t}^{\pi} Q(x,a)= \E[r_{t+1} + Q(x_{t+1},\pi_{t+1}(x_{t+1}))| \Mc, x_t = x, a_t =a ]. 
    \]  
    Let $Q^{\pi}_{\Mc,0},...,Q^{\pi}_{\Mc,H} \in \Re^{|\Xc||\Ac|}$ be defined  according to $Q^{\pi}_{\Mc,H}=0$ and  
    \[
    Q^{\pi}_{\Mc,t} = F^{\pi}_{\Mc, t} Q^{\pi}_{\Mc,t+1} \quad  t\in \{0,...,H-1\}.
    \]
    Then $Q^{\pi}_{\Mc,0}(x, \pi_0(x)) = V^{\pi}_{\Mc}(x)$ and, since $\pi_{t}(x)= \arg\max_{a} Q_{t+1}(x,a)$, $F_{\Mc, t}^{\pi} Q_{t+1} =  F_{\Mc, t} Q_{t+1}$ for all $t$. 
    We can therefore rewrite \eqref{eq: planning to Bellman} as 
    \[ 
    Q_{0}(x,\pi_{0}(x)) -  Q^{\pi}_{\Mc,0}(x, \pi_0(x)) =   \E_{\Mc, \pi}\left[ \sum_{t=0}^{H-1} \left((Q_{t}-F_{\Mc, t}^{\pi} Q_{t+1})(x_t, a_t)  \right) | x_0=x)  \right].
    \]
    We have 
    \begin{eqnarray*}
        Q_0 - Q_{\Mc, 0}^{\pi}&=&Q_0 -F_{\Mc, 0}^{\pi}Q_1 +F_{\Mc, 0}^{\pi}Q_1  - Q_{\Mc, 0}^{\pi} \\
        &=&Q_0 -F_{\Mc, 0}^{\pi}Q_1 +F_{\Mc, 0}^{\pi}Q_1  - F_{\Mc, 0}^{\pi}Q_{\Mc, 1}^{\pi}. 
    \end{eqnarray*}
    By definition, this means 
    \[ 
    (Q_0 - Q_{\Mc, 0}^{\pi})(x, \pi_{0}(x)) = \left(Q_0 -F_{\Mc, 0}^{\pi}Q_1 \right)(x, \pi_{0}(x)) + \E_{\Mc, \pi}[ \left(Q_{1}- Q_{\Mc, 1}^{\pi} \right)(x_1, a_1)  | x_0=x ].
    \]
    The result follows by iterating this relation. 
\end{proof}

\subsection{Proof of Lemma \ref{lem: convex operations}}
\begin{lemma*}[Preservation under convex operations]
    For two collections $(X_1,...,X_n)$ and $(Y_1,...,Y_n)$ of independent random variables with $X_i \succeq_{SO} Y_i$ for each $i\in \{1,...n\}$ and any convex increasing function $f:\Re^n \to \Re$, 
    \[ 
    f(X_1,...,X_n) \succeq_{SO} f(Y_1...,Y_n).
    \] 
\end{lemma*}
\begin{proof} The proof proceeds by induction on $n$. First consider the base case $n=1$. Fix any convex increasing function $u: \Re\to \Re$. 
    Then $u\circ f$ is convex increasing and
    \[ 
    \E[u(f(X_1)) ] \geq \E[u(f(Y_1))]. 
    \] 
Now suppose the result holds for any collection of $n-1$ random variables. 
Fix any convex increasing $u: \Re \to \Re$ : Define the convex increasing functions $U_X: \Re \to \Re$  and $U_{Y}: \Re\to \Re$ by 
\begin{eqnarray*}
    U_{X}(z) &\equiv& \E[u(f(z, X_2,...,X_n))] \\
    U_{Y}(z) &\equiv& \E[u(f(z, Y_2,...,Y_n))]. 
\end{eqnarray*}
For each fixed $z\in \mathbb{R}$, $U_{X}(z)\geq U_{Y}(z)$ since 
\[ 
U_{X}(z) = \E[u(f_{z}(X_2,..,X_n) ) ] \geq  \E[u(f_{z}(Y_2,..,Y_n) ) ] = U_{Y}(z)
\]
where $f_{z}: \Re^{n-1} \to \Re$ is the convex increasing function $f_{z}(x_2,...,x_n) =  f(z, x_2,...,x_n)$ and $f_{z}(X_2,...X_n) \succeq_{SO} f_{z}(Y_2,...Y_n)$ by the inductive hypothesis. We conclude
\[ 
\E[u( f(X_1,..., X_n))]=\E[U_{X}(X_1) ] \geq \E[U_{Y}(X_1)] 
\geq  \E[U_{Y}(Y_1)] = \E[u( f(X_1,..., X_n))]
\]
where the first and last equality use the independence of $(X_1,...,X_n)$ and $(Y_1,...,Y_n)$ along with the Fubini--Tonelli theorem. The final inequality uses the definition of stochastic optimism. 
\end{proof}

\subsection{Proof of Corollary \ref{cor: bound on sampled value function}}

\begin{corollary*}
    If RLSVI is applied with parameters $(\lambda, v, \bar{\theta})$ with $v/\lambda=\beta\geq 3$ , $v=3 H^2$ and $\bar{\theta}=H\Ind$,
    \[
    \E[ \max_{\ell\leq L, t< H} \|V_{Q_{\ell,t+1}}\|_{\infty}] \leq 2H+H^2\sqrt{2\log(|\Xc| |\Ac| HL) }
    \]
\end{corollary*}
\begin{proof}
    To begin, we observe a basic fact about the maximum of Gaussian random variables. Fix independent Gaussian random variables $X_0,X_1,...,X_n \sim N(0,1)$. Let $f: (x_0,...,x_n) \mapsto \max_{i} x_i$ be the maximum function, so $\E[f(X_0,...,X_n)]\leq \sqrt{2 \log(n+1)}$ by a standard Gaussian maximum inequality. Then by Jensen's inequality,
    \begin{eqnarray}
    \E\left[\left(\max_{i\in \{1,...,n\} }  X_{i}\right)_{+}\right] = \E[f(0,X_1,...,X_n) ] &=& \nonumber \E[f(\E[ (X_0,X_1,...,X_n)| X_1,...,X_n ] ) ] \\ \nonumber 
    &\leq &  \E[f(X_0,X_1,...,X_n) ] \\ \label{eq: modified maximal inequality}
    &\leq &  \sqrt{2 \log(n+1)}.  
    \end{eqnarray}  
    For every state action value function $Q \in \Re^{|\Xc| |\Ac|}$, 
    $\| V_{Q} \|_{\infty} \leq 1+\| Q \|_{\infty}.$ 
    Therefore, by equation \ref{eq: RLSVI Bellman}, for every episode $\ell$ and period $t$,
    \begin{eqnarray*}
        \| F_{\ell, t} Q  \|_{\infty} &\leq & \max \{ \| \theta \|_{\infty},  \| V_{Q} \|_{\infty}  \} + \max_{x\in \Xc, a\in \Ac} w_{\ell}(t,x,a) \\
        &\leq & \max \{ \| \theta \|_{\infty},  \| Q \|_{\infty}  \} +1+ \max_{x\in \Xc, a\in \Ac} w_{\ell}(t,x,a) \\
        & \leq & \max \{ \| \theta \|_{\infty},  \| Q \|_{\infty}  \} +1+ w_{\max}
    \end{eqnarray*}
    where $w_{\max} \triangleq \left( \max_{t\leq H, \ell \leq L, a\in \Ac, x\in \Xc} w_{\ell}(t,x,a) \} \right)_{+}$. This implies 
    \[
    \| Q_{\ell, H-1} \|_{\infty} = \| F_{\ell, H-1} 0 \|_{\infty}  \leq \| \theta \|_{\infty} +1 + w_{\max}. 
    \]
    Then
    \[
    \| Q_{\ell, H-2} \|_{\infty} = \| F_{\ell, H-2} Q_{\ell, H-1} \|_{\infty} \leq \max \{ \| \theta \|_{\infty},  \| Q_{\ell, H-1} \|_{\infty}  \} +1+ w_{\max} \leq  \| \theta \|_{\infty}+2(1+w_{\max}).
    \]
    Repeating this by backward induction shows, 
    \[
    \max_{t<H} \| \|Q_{\ell,t+1}\|_{\infty} \leq \| \theta \|_{\infty}+(H-1)(1+w_{\max})
    \]

    Therefore
    \[
    \max_{t< H}  \|V_{Q_{\ell,t+1}}\|_{\infty} \leq  1+ \max_{t< H}  \|Q_{\ell,t+1}\|_{\infty} \leq \|\bar{\theta}\|_{\infty} + H\left(1+w_{\max}\right).
    \]
    In addition
    \begin{eqnarray*}
        \E[w_{\max}] = \E\left[ \left(\max_{t\leq H, \ell \leq L, a\in \Ac, x\in \Xc} w_{\ell}(t,x,a) \right)_{+} \right]
        &=& \E\left[ \left(\max_{t< H, \ell \leq L, a\in \Ac, x\in \Xc}\sigma_{\ell}(t,x,a)\frac{w_{\ell}(t,x,a)}{\sigma_{\ell}(t,x,a)} \right)_{+} \right] \\
        &\leq& \sqrt{\frac{v}{\beta}} \E\left[ \left( \max_{t< H, \ell \leq L, a\in \Ac, x\in \Xc}  \frac{w_{\ell}(t,x,a)}{\sigma_{\ell}(t,x,a)} \right)_{+} \right] \\
        &\leq& \sqrt{2 v/\beta  \log(1+|\Xc| |\Ac| HL) }
    \end{eqnarray*}
    where the last step uses equation \eqref{eq: modified maximal inequality}. Combining these results implies,
    \[
    \E[ \max_{\ell\leq L, t< H} \|V_{Q_{\ell,t+1}}\|_{\infty}] \leq \|\bar{\theta}\|_{\infty} +H+ H\E \left[w_{\max} \right]\leq \|\bar{\theta}\|_{\infty}+H+H\sqrt{2 (v/\beta) \log(1+|\Xc| |\Ac| HL) }.
    \]
    The result then follows by plugging in for $\beta \geq 3$, $v=3H^2$, and $\bar{\theta}=H$.
\end{proof}

\subsection{Proof of Lemma \ref{lem: bound on sums}}
\begin{lemma*}
    If $\beta\geq 2$, with probability 1,
    \[
    \sum_{\ell\leq L}\sum_{t\leq H}  \frac{1}{\beta+n_{\ell}(t, x_t, a_t)} \leq H|\Xc||\Ac| \log\left(\frac{1+L}{|\Xc||\Ac|}\right)
    \]
    and
    \[ 
    \sum_{\ell\leq L}\sum_{t\leq H}  \sqrt{\frac{1}{\beta+n_{\ell}(t, x_t, a_t)}} \leq 2H\sqrt{|\Xc||\Ac|L}.
    \]
\end{lemma*}
\begin{proof}
    Set $\Yc = \{0,...,H-1\} \times \Xc \times \Ac$ to be the set of valid period, state, action triples $y=(t,x,a) \in \Yc$. 
    Note that $|\Yc|= H|\Xc||\Ac|$.   
    We have 
    \begin{eqnarray*}
        \sum_{t\leq H, \ell\leq L} \frac{1}{\beta+n_{\ell}(t, x_t, a_t)} = \sum_{y\in \Yc}  \sum_{i=0}^{n_{L}(y)-1}\frac{1}{\beta + i} &\leq& \sum_{y\in \Yc} \intop_{\beta-1}^{n_{L}(y)+\beta-1} \frac{1}{z}dz \\
        &=& \sum_{y\in \Yc} \log\left(\frac{\beta-1+n_{L}(y)}{\beta-1}\right) \\
        &\leq &  \sum_{y\in \Yc} \log\left(1+n_{L}(y)\right)\\
        &\leq&   |\Yc| \log\left(\frac{\sum_{y\in \Yc} (1+ n_{L}(y))}{|\Yc|}\right) \\
        &=& |\Yc| \log\left(1+\frac{LH}{|\Yc|}\right). \\
        &=&  H|\Xc||\Ac| \log\left(1+\frac{L}{|\Xc||\Ac|}\right). \\
    \end{eqnarray*}
    In addition
    \begin{eqnarray*}
        \sum_{t\leq H, \ell\leq L} \sqrt{\frac{1}{\beta+n_{\ell}(t, x_t, a_t)}} = \sum_{y\in \Yc}  \sum_{i=0}^{n_{L}(y)-1}\sqrt{\frac{1}{\beta + i}} &\leq & \sum_{y\in \Yc} \intop_{z=0}^{n_{L}(y)} \frac{1}{(\beta-1+z)^{1/2}}dz \\ 
        &\leq & \sum_{y\in \Yc} \intop_{z=0}^{n_{L}(y)} \frac{1}{z^{1/2}}dz \\
        &=& \sum_{y\in \Yc} 2\sqrt{n_{L}(y)} \\
        &\leq& 2\sqrt{|\Yc| \sum_{y\in \Yc}n_{L}(y)} \\
        &=& 2H\sqrt{|\Xc||\Ac|L}.
    \end{eqnarray*}
\end{proof}

\subsection{Proof of Lemma \ref{lem: Dir Norm}}
This section establishes Lemma \ref{lem: Dir Norm} through a sequence of results. First, Lemma \ref{lem: general stochastic optimism} provides general conditions for stochastic optimism.
Next, we use these properties to show a stochastic dominance relation between Dirichlet and Beta distributions in Lemma \ref{lem: Beta Dirichlet dominance} and between Beta and Gaussian distributions in Lemma \ref{lem: Gaussian Beta dominance}.
The final proof of Lemma \ref{lem: Dir Norm}, given at the end of this section, uses a simple combination of these results.

We begin with Lemma \ref{lem: general stochastic optimism}, which reproduces several sufficient conditions for stochastic optimism.
Conditions (1.) and (2.) are classic results in the theory of second order stochastic dominance (SSD) and we refer to \citep{hadar1969rules, hanoch1969efficiency} for proofs.
\footnote{The literature on SSD generally considers \emph{concave} and increasing utility functions whereas we consider \emph{convex} and increasing utility functions. Results on SSD can be easily translated into results about stochastic optimism through swapping positive and negative signs.}
For a clever and explicit construction of coupled random variables per condition (1.) see \citep{machina1997increasing}.
In fact, both (1.) and (2.) are also necessary conditions for stochastic optimism, although our results do not rely upon this.

\begin{figure}[!h]
    \centering
    \begin{subfigure}{.5\textwidth}
        \centering
        \includegraphics[width=.9\linewidth]{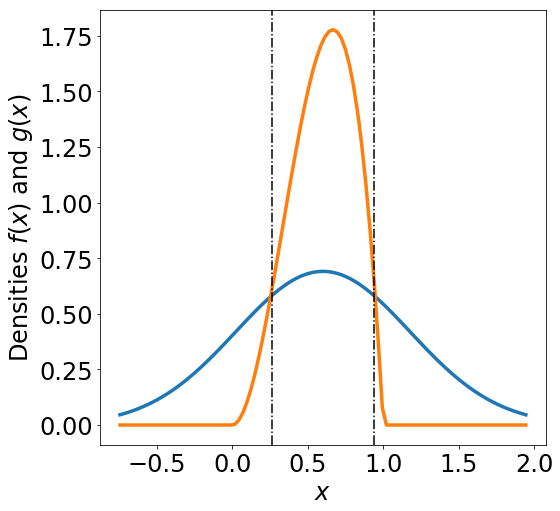}
        \vspace{-2mm}
        \caption{Comparison of PDFs}
        \label{fig:beta_vs_normal_pdfs}
    \end{subfigure}%
    \begin{subfigure}{.5\textwidth}
        \centering
        \includegraphics[width=.9\linewidth]{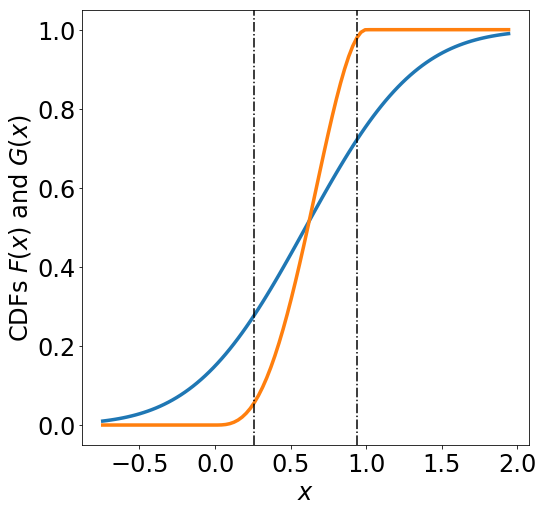}
        \vspace{-2mm}
        \caption{Comparison of CDFs}
        \label{fig:beta_vs_normal_cdfs}
    \end{subfigure}
    \begin{subfigure}{.5\textwidth}
    \centering
    \includegraphics[width=.9\linewidth]{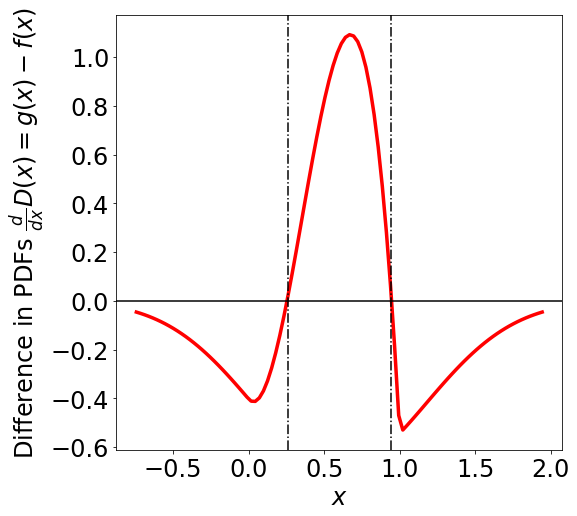}
    \vspace{-2mm}
    \caption{Difference in PDFs $D'(s)=g(s)-f(s)$}
    \label{fig:beta_vs_normal_pdfs_diff}
    \end{subfigure}%
    \begin{subfigure}{.5\textwidth}
    \centering
    \includegraphics[width=.9\linewidth]{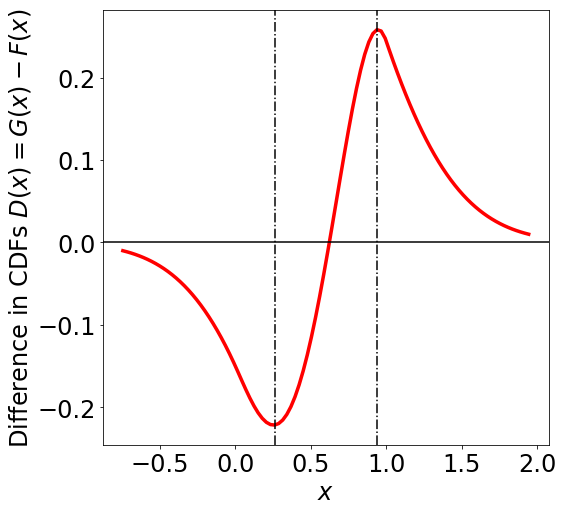}
    \vspace{-2mm}
    \caption{Difference in CDFs $D(s)=G(s)-F(s)$}
    \label{fig:beta_vs_normal_cdfs_diff}
    \end{subfigure}%

    \caption{Comparison of ${\rm Beta}(3,2)$ and $N(3/5, 1/3)$ distributions. The dashed vertical lines indicate the boundaries of set $C$.}
    \label{fig:beta_vs_normal}
\end{figure}

Condition (3.) is less widely stated, but is a simple consequence of (2.) and has been known since \cite{hanoch1969efficiency}. 
To our knowledge, condition (4.) is a new result that provides a sufficient condition for stochastic optimism that is much easier to verify than condition (3.). 
Some intuition for this result is captured by Figure \ref{fig:beta_vs_normal_pdfs}.
We provide a detailed proof that analytically establishes many of the properties observable in Figure \ref{fig:beta_vs_normal}. 

\begin{lemma}[Sufficient Conditions for Stochastic Optimism]\label{lem: general stochastic optimism}
    Consider two integrable random variables $X$ and $Y$ with probability density functions $f$ and $g$. Let $F(s)=\intop^{s}_{-\infty} f(x)dx$ and $G(s)=\intop_{-\infty}^{s} g(x)dx$ denote the corresponding cumulative distribution functions. Then $X\succeq_{SO} Y$ if any of the following properties hold:
    \begin{enumerate} 
    \item One can construct random variables $(\tilde{X}, \tilde{Y},  \tilde{W})$ on a joint probability space such that $\tilde{X}$ has marginal distribution $F$, $\tilde{Y}$ has marginal distribution $G$, $\E[\tilde{W}| \tilde{Y}] \geq 0$ and 
    \[
    \tilde{X} =\tilde{Y} + \tilde{W}. 
    \]
    \item For all $a\in \mathbb{R}$,
    \[ 
    \intop_{a}^{\infty} \left(G(s) - F(s) \right)ds \geq  0.
    \]
    \item $\E[X] \geq \E[Y]$ and there exists $a \in \mathbb{R}$ such that 
    \[ 
    G(s) \geq F(s) \iff s\geq a . 
    \]
    \item $\E[X] \geq \E[Y]$ and  $C:= \{x \in \mathbb{R} : g(x) > f(x) \}$ is convex. 
    \end{enumerate} 
\end{lemma} 
\begin{proof}
    (1.) follows easily by the tower property of conditional expectation and the conditional Jensen inequality. We have $\E[\tilde{X} \mid \tilde{Y}] \geq \tilde{Y}$ and hence for any convex increasing function $u:\mathbb{R}\to \mathbb{R}$, 
    \[ 
    \E[u(\tilde{X})] =\E[ \E[ u(\tilde{X}) \mid \tilde{Y} ]] \geq \E[ u\left(\E[\tilde{X} \mid \tilde{Y}]\right)] \geq \E[ u\left( \tilde{Y} \right) ].
    \]
    (2.) follows from integration by parts and (3.) is a fairly simple consequence of (2.). See \cite{hanoch1969efficiency} for a proof.  The claim (4.) is new and established here. 

Define the function
\[
D(s) = G(s)-F(s) = \intop^{s}_{-\infty} (g(x)-f(x))dx
\]
and note $C= \{x : g(x)-f(x)>0 \}=\{x : D'(x) >0\}$. We have $\lim_{t\to \infty} D(t) = 0$. Two distributions yielding convex $C$ and the corresponding functions $D'(s)$ and $D(s)$ are pictured in Figure \ref{fig:beta_vs_normal}. 

Integration by parts shows \citep[Lemma 1]{hanoch1969efficiency}
\begin{equation}\label{eq: non-negative CDF integral}
\intop_{-\infty}^{\infty} D(s)ds  = \E[X]-\E[Y] \geq 0.  
\end{equation}
 We first show $\sup\{C\}<\infty$. Note that we cannot have $C=(-\infty, \infty )$, as this would imply $\intop g(x) dx > \intop f(x)dx$, contradicting that $g$ and $f$ are probability density functions.
 Now suppose for contradition that $C=(\underline{c}, \infty)$ for some $\underline{c} \in \mathbb{R}$.
 Then, for $t\leq \underline{c}$ we have $D(t) = \intop_{-\infty}^{t} (g(s)-f(s))ds <0$.
 Since $\lim_{t\to \infty} D(t) =0$, we have that for $t\geq \underline{c}$, $D(t) = -\intop_{t}^{\infty}  (g(s)-f(s)) ds <0$.
 Thus we have showed that $D(t)< 0$ for all $t\in \mathbb{R}$ contradicting \eqref{eq: non-negative CDF integral}.

Next, suppose $C=(-\infty, \overline{c})$  for $\overline{c} \in \mathbb{R}$.
Then for $s\in C$, $D'(s)>0$ by definition of $C$, which shows $D(t)=\intop_{-\infty}^t D'(s) ds >0$ for $t\in C$.
But since  $D'(s) <0$ for all $s> \overline{c}$ and $\lim_{t\to \infty} D(t) =0$, we must have $D(t) >0$ also for all $t\geq \overline{c}$. 
Since $D$ is non-negative, we have $\intop_{a}^{\infty} D(s) \geq 0$ for all $a\in \mathbb{R}$ and hence $X\succeq_{SO} Y$.
\footnote{In fact, $X$ first-order stochastically dominates $Y$ in this case.}

Finally, suppose $C=(-\underline{c}, \overline{c})$ and refer to Figures \ref{fig:beta_vs_normal_pdfs_diff} and \ref{fig:beta_vs_normal_cdfs_diff} for visual guidance.
For $s< \underline{c}$,  $D'(s)\leq 0$ and hence $D(t)=\intop_{-\infty}^{t} D'(s) ds \leq 0$ for $t\leq \underline{c}$.
For $s\in C$, $D'(s)>0$.
For  $s> \overline{c}$, $D'(s) \leq 0$, which implies $D(t) \geq 0$ for all $t\geq \overline{c}$  since $\lim_{t\to \infty} D(t) =0$.
Since we have shown $D(\underline{c})\leq 0 \leq D(\overline{c})$ there must exists a single crossing point $a\in (\underline{c}, \overline{c})$ with $D(a)=0$.
For this value of $a$, we have 
\[ 
G(s)-F(s) \iff s\geq a,
\]
implying $X\succeq_{SO} Y$.
This completes the proof of condition (4.).
\end{proof}

Our next result uses several relationships between Gamma, Beta, and Dirichlet random variables to establish a stochastic optimism relationship between specific matched distributions.
First, for two independent Gamma distributed random variables  $\gamma\sim {\rm Gamma}(\alpha, 1)$ and $\gamma'\sim {\rm Gamma}(\beta,1)$, we have that $\frac{\gamma}{\gamma+\gamma'} \sim {\rm Beta}(\alpha, \beta)$.
Next, $\gamma+\gamma' \sim {\rm Gamma}(\alpha+\beta, 1)$ and $\E[\gamma \mid \gamma + \gamma' ] = (\gamma+\gamma')\left( \frac{\alpha}{\alpha+\beta} \right)$.
Finally, for a collection of independent Gamma random variables $\gamma_i \sim {\rm Gamma}(\alpha_i, 1)$, the random probability vector $P= \left(\frac{\gamma_1}{\sum_{j=1}^{n} \gamma_j}, \ldots   \frac{\gamma_n}{\sum_{j=1}^{n} \gamma_j}\right)$ follows a ${\rm Dirichlet}(\alpha)$ distribution.
Lemma \ref{lem: Beta Dirichlet dominance} compares the distribution of the inner product $P^{\top} V$ between a Dirichlet random variable $P$ and a fixed vector $V$ to an appropriate Beta random variable $X$. 
\begin{lemma}[Beta-Dirichlet optimism]\label{lem: Beta Dirichlet dominance}
    If $V\in \mathbb{R}^n$ where $0=V_1\leq \cdots \leq  V_n=1$,  $P \sim {\rm Dirichlet}(\alpha)$ where $\alpha \in \Re^n_+$, and $X\sim {\rm Beta}(\sum_{i=1}^{n} \alpha_i V_i, \, \sum_{i=1}^{n} \alpha_i(1-V_i) )$ then $X\succeq_{SO} P^{\top} V$. 
\end{lemma} 
\begin{proof}
    Our proof constructs coupled Dirichlet and and Beta random variables with the marginal distributions described in the lemma's statement. Consider independent Gamma-distributed random variables $(\gamma_1^0,\gamma_1^1, \ldots, \gamma_n^0, \gamma_n^1)$ where
    \[ 
    \gamma_i^0 \sim {\rm Gamma}(\alpha_i V_i, 1)   \quad \text{and} \quad \gamma_i^1 \sim {\rm Gamma}(\alpha_i (1-V_i), 1)  \qquad i=1,\ldots, n.
    \]
    Set $\gamma_i=\gamma_i^0 + \gamma_i^1$ so that $\gamma_i \sim {\rm Gamma}(\alpha_i,1)$. Then, set 
    \[ 
    P= \frac{(\gamma_1, \ldots, \gamma_n)}{\sum_{i=1}^{n} \gamma_i} \sim {\rm Dirichlet}(\alpha) 
    \]  
    and 
    \[
    X=\frac{\sum_{i=1}^{n} \gamma_i^0 }{\sum_{i=1}^{n} \gamma_i}= \frac{\sum_{i=1}^{n} \gamma_i^0 }{\sum_{i=1}^{n} \gamma_i^0 + \sum_{i=1}^{n} \gamma_i^1} \sim {\rm Beta}\left(\sum_{i=1}^{n} \alpha_i V_i, \, \sum_{i=1}^{n} \alpha_i(1-V_i) \right).
    \]
    Our result follows from the fact that $\sum_{i=1}^{n} \gamma_i^0 \sim {\rm Gamma}(\sum \alpha_i V_i, 1 )$ and $\sum_{i=1}^{n} \gamma_i^{1} \sim {\rm Gamma}(\sum_{i=1}^{n} \alpha_i (1-V_i),1)$ and the ratio of Gamma distributed random variables follows a Beta distribution.
    Now, we have that 
    \[ 
    \E[\gamma_i^{0} | \gamma_i ] = \gamma_i \left(\frac{\E[\gamma_i^{0} ]}{\E[\gamma_i^{0}] +\E[\gamma_i^{1}]}\right) = \gamma_i \left(\frac{\alpha_i V_i}{\alpha_i}\right) = \gamma_i V_i.
    \]
    This implies,
    \begin{eqnarray*}
        \E[X | \gamma]  = \E\left[ \frac{\sum_{i=1}^{n} \gamma_i^0 }{\sum_{i=1}^{n}\gamma_i} \,  \bigg| \, \gamma \right] = \frac{\sum_{i=1}^{n} \E[\gamma_i^0 \mid \gamma_i]}{\sum_{i=1}^{n}\gamma_i}  = \frac{\sum_{i=1}^{n} \gamma_i V_i}{\sum_{i=1}^{n} \gamma_i} = P^{\top} V.
    \end{eqnarray*}
Set $Y=P^{\top} V$ and $W=X-Y$ so that $X=Y+W$. We have shown $\E[W \mid \gamma] =0$ and so $\E[W \mid Y]= \E[ \E[W \mid \gamma] \mid Y]=0$ and the result follows from part 1 of Lemma \ref{lem: general stochastic optimism}. 
\end{proof}

Our next result, Lemma \ref{lem: Gaussian Beta dominance} shows that a Gaussian distribution with sufficiently large mean and varaince stochastically dominates a Beta distribution.
The proof shows this by looking at the ratio of their probability density functions and concluding this is a quasi-concave function. The basic properties of quasi-concave functions used here are in \citep[Section 3.4]{boyd2004convex}.
\begin{lemma}[Gaussian-Beta optimism]\label{lem: Gaussian Beta dominance}
    Let $X\sim N(\mu,\sigma)$ and $Y\sim {\rm Beta}(\alpha, \beta)$.  If $\alpha+\beta \geq 3$, $\mu = \alpha/(\alpha+\beta)$ and $\sigma^2 = (\alpha+\beta -2)^{-1}$, then $X\succeq_{SO}Y$.  
\end{lemma} 
\begin{proof}
    Let $f(x)=\frac{1}{\sqrt{2\pi} \sigma} e^{-(x-\mu)^2 / 2\sigma^2 }$ and $g(x)= \frac{x^{\alpha-1} (1-x)^{\beta-1}}{B(\alpha,\beta)}$ denote the probability density functions corresponding to the $N(\mu,\sigma)$  and ${\rm Beta}(\alpha, \beta)$ distributions. We show $g(x)/f(x)$ is quasi-concave, which implies the super-level set $C= \{x : g(x)/f(x) \geq 1\}$ is convex. The result then follows by Lemma \ref{lem: general stochastic optimism} and condition (4.). 

    Let $\ell(x) = \log\left(g(x)/f(x)\right)$. Since $g(x)/f(x) = e^{\ell(x)}$ is a monotone function of $\ell$, it is quasi-concave a long as $\ell(x)$ is quasi-concave. We show $\ell$ is quasi-concave by considering two cases. First, if $\alpha>1$ and $\beta>1$, we show $\ell$ is concave (and therefore quasi-concave). Differentiation shows 
    \[ 
    \ell''(x) = -\left(\frac{\alpha-1}{x^2}+\frac{\beta-1}{(1-x)^2}\right)+ \frac{1}{\sigma^2} \leq -(\alpha+\beta -2) + \frac{1}{\sigma^2} \leq 0 
    \]
    where the last step follows from our choice of $\sigma^2$. In the case where either $\alpha\in (0,1)$ or $\beta\in (0,1)$, we find that $\ell$ is monotone (and therefore quasi-concave). Assume $\beta \in (0,1)$, in which case $\alpha\geq 2$. Then, 
    \begin{eqnarray*}
        \ell'(x) = \frac{\alpha-1}{x}- \frac{\beta-1}{1-x}+\frac{x-\mu}{\sigma^2}  \geq \frac{\alpha-1}{x}+\frac{x-\mu}{\sigma^2}.
    \end{eqnarray*}
    This is non-negative for $x\geq \mu$. For $x<\mu$, plugging in $\mu=\alpha/(\alpha+\beta)$ and $\sigma^2 \geq 1/(\alpha+\beta)$  gives 
    \[ 
    \ell'(x) \geq  \frac{\alpha-1}{x}+\frac{x-\mu}{\sigma^2} \geq \frac{\alpha-\alpha/2}{x} + x(\alpha+\beta) -\alpha \geq \frac{\alpha}{2x}+\alpha(x-1) \geq 0 
    \]
    where the last inequality can be verified by solving $\min_{x>0} \, \frac{2}{x} + x-1 = \frac{2}{\sqrt{1/2}}+ \sqrt{1/2} -1 \geq 0$. 
\end{proof}

With these results we are now ready to prove Lemma \ref{lem: Dir Norm}, which we restate below.

\begin{lemma*}
Let $Y = P^T V$ for $V \in \Re^n$ fixed and $P \sim {\rm Dirichlet}(\alpha)$ with $\alpha \in \Re^n_+$ and $\sum_{i=1}^n \alpha_i \ge 3$.
Let $X \sim N(\mu, \sigma^2)$ with
$ \mu \geq \frac{\sum_{i=1}^n \alpha_i V_i}{\sum_{i=1}^n \alpha_i},
\ \sigma^2 \geq 3 \left(\sum_{i=1}^n \alpha_i \right)^{-1} {\rm Span}(V)^2$, then $X \succeq_{SO} Y$.
\end{lemma*}

\begin{proof}[Proof of Lemma \ref{lem: Dir Norm}]
    Without loss of generality, assume $V_1 \leq \ldots \leq V_n$. Set $V'_i = (V_i-V_1)/(V_n-V_1)$ for $i=1,\ldots, n$.  If $X\sim {\rm Beta}(\sum_{i=1}^{n} \alpha_i V'_i, \, \sum_{i=1}^{n} \alpha_i(1-V'_i) )$, then $X\succeq_{SO} P^{\top} V'$ by Lemma \ref{lem: Beta Dirichlet dominance}. Let $Z\sim N(\mu, \sigma^2)$ where $\mu = \E[X]= \sum_{i} \alpha_i V'_i / (\sum_{j} \alpha_j )$ and $\sigma^2= (\sum_{i=1}^{n} \alpha_i -2 )^{-1}$. Then $Z\succeq_{SO} X$ by Lemma \ref{lem: Gaussian Beta dominance}. Combining these results gives
    \[
    P^{\top} V = V_1 + (V_n-V_1) P^{\top} V' \preceq_{SO} V_1 + (V_n-V_1)X \preceq_{SO} V_1 + (V_n -V_1) Z. 
    \]
    We have that $V_1 + (V_n - V_1)Z$ is normally distributed with 
    \[ 
    \E[V_1 + (V_n - V_1)Z ] = V_1 + (V_n -V_1) \mu =V_1 +  (V_n -V_1) \left(\frac{\sum_{i} \alpha_i V'_i }{\sum_{j} \alpha_j}\right) = \frac{\sum_{i=1}^{n} \alpha_i V_i }{\sum_{i=1}^{n} \alpha_i }
    \]
    and
    \[
    {\rm Variance}\left(V_1 + (V_n - V_1)Z\right) =   (V_n -V_1)^2 \sigma^2 = {\rm Span}(V)^2 \left(\sum_{i=1}^{n} \alpha_i -2 \right)^{-1} \leq 3 \cdot {\rm Span}(V)^2  \left(\sum_{i=1}^{n} \alpha_i \right)^{-1}. 
    \]
\end{proof}

\bibliography{reference}

\begin{thebibliography}{83}
\providecommand{\natexlab}[1]{#1}
\providecommand{\url}[1]{\texttt{#1}}
\expandafter\ifx\csname urlstyle\endcsname\relax
  \providecommand{\doi}[1]{doi: #1}\else
  \providecommand{\doi}{doi: \begingroup \urlstyle{rm}\Url}\fi

\bibitem[Abbasi-Yadkori and Szepesv{\'a}ri(2011)]{Abbasi-Yadkori2011}
Yasin Abbasi-Yadkori and Csaba Szepesv{\'a}ri.
\newblock Regret bounds for the adaptive control of linear quadratic systems.
\newblock \emph{Journal of Machine Learning Research - Proceedings Track},
  19:\penalty0 1--26, 2011.

\bibitem[Adam et~al.(2012)Adam, Busoniu, and Babuska]{adam2012experience}
Sander Adam, Lucian Busoniu, and Robert Babuska.
\newblock Experience replay for real-time reinforcement learning control.
\newblock \emph{IEEE Transactions on Systems, Man, and Cybernetics, Part C
  (Applications and Reviews)}, 42\penalty0 (2):\penalty0 201--212, 2012.

\bibitem[Agrawal and Goyal(2012)]{agrawal2012analysis}
Shipra Agrawal and Navin Goyal.
\newblock Analysis of {Thompson} sampling for the multi-armed bandit problem.
\newblock In \emph{Conference on Learning Theory}, pages 39--1, 2012.

\bibitem[Agrawal and Goyal(2013{\natexlab{a}})]{agrawal2012further}
Shipra Agrawal and Navin Goyal.
\newblock Further optimal regret bounds for {Thompson} sampling.
\newblock In \emph{Artificial Intelligence and Statistics}, pages 99--107,
  2013{\natexlab{a}}.

\bibitem[Agrawal and Goyal(2013{\natexlab{b}})]{agrawal2013thompson}
Shipra Agrawal and Navin Goyal.
\newblock Thompson sampling for contextual bandits with linear payoffs.
\newblock In \emph{Proceedings of the 30th Annual International Conference on
  Machine Learning}, pages 127--135, 2013{\natexlab{b}}.

\bibitem[Auer and Ortner(2006)]{Auer2006}
Peter Auer and Ronald Ortner.
\newblock Logarithmic online regret bounds for undiscounted reinforcement
  learning.
\newblock In \emph{Advances in Neural Information Processing Systems 19}, pages
  49--56, 2006.

\bibitem[Azar et~al.(2017)Azar, Osband, and Munos]{azar2017minimax}
Mohammad~Gheshlaghi Azar, Ian Osband, and R{\'e}mi Munos.
\newblock Minimax regret bounds for reinforcement learning.
\newblock In \emph{Proceedings of the 34th Annual International Conference on
  Machine Learning}, 2017.

\bibitem[Azizzadenesheli et~al.(2018)Azizzadenesheli, Brunskill, and
  Anandkumar]{azizzadenesheli2018efficient}
Kamyar Azizzadenesheli, Emma Brunskill, and Animashree Anandkumar.
\newblock Efficient exploration through {B}ayesian deep q-networks.
\newblock \emph{arXiv preprint arXiv:1802.04412}, 2018.

\bibitem[Bartlett and Tewari(2009)]{Bartlett2009}
Peter~L. Bartlett and Ambuj Tewari.
\newblock {REGAL}: A regularization based algorithm for reinforcement learning
  in weakly communicating {MDPs}.
\newblock In \emph{Proceedings of the 25th Conference on Uncertainty in
  Artificial Intelligence (UAI2009)}, pages 35--42, June 2009.

\bibitem[Bartlett et~al.(2017)Bartlett, Foster, and
  Telgarsky]{bartlett2017spectrally}
Peter~L Bartlett, Dylan~J Foster, and Matus~J Telgarsky.
\newblock Spectrally-normalized margin bounds for neural networks.
\newblock In \emph{Advances in Neural Information Processing Systems 30}, pages
  6241--6250, 2017.

\bibitem[Bellemare et~al.(2016)Bellemare, Srinivasan, Ostrovski, Schaul,
  Saxton, and Munos]{bellemare2016count}
Marc Bellemare, Sriram Srinivasan, Georg Ostrovski, Tom Schaul, David Saxton,
  and Remi Munos.
\newblock Unifying count-based exploration and intrinsic motivation.
\newblock In \emph{Advances in Neural Information Processing Systems 29}, pages
  1471--1479. 2016.

\bibitem[Bellemare et~al.(2017)Bellemare, Dabney, and
  Munos]{bellemare2017distributional}
Marc~G Bellemare, Will Dabney, and R{\'e}mi Munos.
\newblock A distributional perspective on reinforcement learning.
\newblock In \emph{Advances in Neural Information Processing Systems 30}, 2017.

\bibitem[Bertsekas and Tsitsiklis(1996)]{Bertsekas1996}
Dimitri~P. Bertsekas and John Tsitsiklis.
\newblock \emph{Neuro-Dynamic Programming}.
\newblock Athena Scientific, September 1996.

\bibitem[Bickel and Freedman(1981)]{bickel1981some}
Peter~J Bickel and David~A Freedman.
\newblock Some asymptotic theory for the bootstrap.
\newblock \emph{The Annals of Statistics}, pages 1196--1217, 1981.

\bibitem[Blackwell(1965)]{blackwell1965discounted}
David Blackwell.
\newblock Discounted dynamic programming.
\newblock \emph{The Annals of Mathematical Statistics}, 36\penalty0
  (1):\penalty0 226--235, 1965.

\bibitem[Boyd and Vandenberghe(2004)]{boyd2004convex}
Stephen Boyd and Lieven Vandenberghe.
\newblock \emph{Convex optimization}.
\newblock Cambridge university press, 2004.

\bibitem[Brafman and Tennenholtz(2002)]{Brafman2002}
Ronen~I. Brafman and Moshe Tennenholtz.
\newblock R-max - a general polynomial time algorithm for near-optimal
  reinforcement learning.
\newblock \emph{Journal of Machine Learning Research}, 3:\penalty0 213--231,
  2002.

\bibitem[Dann and Brunskill(2015)]{Dann2015}
Christoph Dann and Emma Brunskill.
\newblock Sample complexity of episodic fixed-horizon reinforcement learning.
\newblock In \emph{Advances in Neural Information Processing Systems 28}, pages
  2818--2826. 2015.

\bibitem[Dearden et~al.(1998)Dearden, Friedman, and Russell]{DeardenFR98}
Richard Dearden, Nir Friedman, and Stuart~J. Russell.
\newblock Bayesian {Q}-learning.
\newblock In \emph{AAAI Conference on Artificial Intelligence}, pages 761--768,
  1998.

\bibitem[Deisenroth et~al.(2013)Deisenroth, Neumann, Peters,
  et~al.]{deisenroth2013survey}
Marc~Peter Deisenroth, Gerhard Neumann, Jan Peters, et~al.
\newblock A survey on policy search for robotics.
\newblock \emph{Foundations and Trends{\textregistered} in Robotics},
  2\penalty0 (1--2):\penalty0 1--142, 2013.

\bibitem[Eckles and Kaptein(2019)]{eckles2019bootstrap}
Dean Eckles and Maurits Kaptein.
\newblock Bootstrap thompson sampling and sequential decision problems in the
  behavioral sciences.
\newblock \emph{SAGE Open}, 9\penalty0 (2):\penalty0 2158244019851675, 2019.

\bibitem[Efron(1982)]{efron1982jackknife}
Bradley Efron.
\newblock \emph{The jackknife, the bootstrap and other resampling plans},
  volume~38.
\newblock SIAM, 1982.

\bibitem[Efron and Tibshirani(1994)]{efron1994introduction}
Bradley Efron and Robert~J Tibshirani.
\newblock \emph{An introduction to the bootstrap}.
\newblock CRC press, 1994.

\bibitem[Fortunato et~al.(2018)Fortunato, Azar, Piot, Menick, Hessel, Osband,
  Graves, Mnih, Munos, Hassabis, Pietquin, Blundell, and
  Legg]{fortunato2017noisy}
Meire Fortunato, Mohammad~Gheshlaghi Azar, Bilal Piot, Jacob Menick, Matteo
  Hessel, Ian Osband, Alex Graves, Volodymyr Mnih, Remi Munos, Demis Hassabis,
  Olivier Pietquin, Charles Blundell, and Shane Legg.
\newblock Noisy networks for exploration.
\newblock In \emph{International Conference on Learning Representations}, 2018.

\bibitem[Fushiki(2005)]{fushiki2005bootstrap}
Tadayoshi Fushiki.
\newblock Bootstrap prediction and bayesian prediction under misspecified
  models.
\newblock \emph{Bernoulli}, pages 747--758, 2005.

\bibitem[Glorot and Bengio(2010)]{glorot2010understanding}
Xavier Glorot and Yoshua Bengio.
\newblock Understanding the difficulty of training deep feedforward neural
  networks.
\newblock In \emph{Proceedings of the 13th international conference on
  artificial intelligence and statistics}, pages 249--256, 2010.

\bibitem[Gopalan and Mannor(2015)]{gopalan2015thompson}
Aditya Gopalan and Shie Mannor.
\newblock Thompson sampling for learning parameterized {Markov} decision
  processes.
\newblock In \emph{Proceedings of the 28th Annual Conference on Learning
  Theory}, 2015.

\bibitem[Hadar and Russell(1969)]{hadar1969rules}
Josef Hadar and William~R Russell.
\newblock Rules for ordering uncertain prospects.
\newblock \emph{The American Economic Review}, pages 25--34, 1969.

\bibitem[Hanoch and Levy(1969)]{hanoch1969efficiency}
G~Hanoch and H~Levy.
\newblock The efficiency analysis of choices involving risk.
\newblock \emph{The Review of Economic Studies}, 36\penalty0 (3):\penalty0
  335--346, 1969.

\bibitem[Ibarra et~al.(2016)Ibarra, Ramos, and Roemheld]{ibarra2016angrier}
Imanol~Arrieta Ibarra, Bernardo Ramos, and Lars Roemheld.
\newblock Angrier birds: Bayesian reinforcement learning.
\newblock \emph{arXiv preprint arXiv:1601.01297}, 2016.

\bibitem[Ibrahimi et~al.(2012)Ibrahimi, Javanmard, and Roy]{Ibrahimi2012}
Morteza Ibrahimi, Adel Javanmard, and Benjamin~V Roy.
\newblock Efficient reinforcement learning for high dimensional linear
  quadratic systems.
\newblock In \emph{Advances in Neural Information Processing Systems 25}, pages
  2636--2644, 2012.

\bibitem[Jaksch et~al.(2010)Jaksch, Ortner, and Auer]{Jaksch2010}
Thomas Jaksch, Ronald Ortner, and Peter Auer.
\newblock Near-optimal regret bounds for reinforcement learning.
\newblock \emph{Journal of Machine Learning Research}, 11:\penalty0 1563--1600,
  2010.

\bibitem[Kakade(2003)]{Kakade2003}
Sham Kakade.
\newblock \emph{On the Sample Complexity of Reinforcement Learning}.
\newblock PhD thesis, University College London, 2003.

\bibitem[Kearns and Koller(1999)]{Kearns1999}
Michael~J. Kearns and Daphne Koller.
\newblock Efficient reinforcement learning in factored {MDP}s.
\newblock In \emph{IJCAI}, pages 740--747, 1999.

\bibitem[Kearns and Singh(2002)]{Kearns2002}
Michael~J. Kearns and Satinder~P. Singh.
\newblock Near-optimal reinforcement learning in polynomial time.
\newblock \emph{Machine Learning}, 49\penalty0 (2-3):\penalty0 209--232, 2002.

\bibitem[Li and Littman(2010)]{Li2010}
Lihong Li and Michael~L Littman.
\newblock Reducing reinforcement learning to kwik online regression.
\newblock \emph{Annals of Mathematics and Artificial Intelligence}, 58\penalty0
  (3-4):\penalty0 217--237, 2010.

\bibitem[Li et~al.(2008)Li, Littman, and Walsh]{LiLW08}
Lihong Li, Michael~L Littman, and Thomas~J Walsh.
\newblock Knows what it knows: a framework for self-aware learning.
\newblock In \emph{Proceedings of the 25th international conference on Machine
  learning}, pages 568--575. ACM, 2008.

\bibitem[Lipton et~al.(2018)Lipton, Li, Gao, Li, Ahmed, and
  Deng]{lipton2016efficient}
Zachary Lipton, Xiujun Li, Jianfeng Gao, Lihong Li, Faisal Ahmed, and Li~Deng.
\newblock {BBQ}-networks: Efficient exploration in deep reinforcement learning
  for task-oriented dialogue systems.
\newblock \emph{AAAI Conference on Artificial Intelligence}, 2018.

\bibitem[Lu and Van~Roy(2017)]{NIPS2017_6918}
Xiuyuan Lu and Benjamin Van~Roy.
\newblock Ensemble sampling.
\newblock In I.~Guyon, U.~V. Luxburg, S.~Bengio, H.~Wallach, R.~Fergus,
  S.~Vishwanathan, and R.~Garnett, editors, \emph{Advances in Neural
  Information Processing Systems 30}, pages 3258--3266. 2017.

\bibitem[Machina and Pratt(1997)]{machina1997increasing}
Mark Machina and John Pratt.
\newblock Increasing risk: some direct constructions.
\newblock \emph{Journal of Risk and Uncertainty}, 14\penalty0 (2):\penalty0
  103--127, 1997.

\bibitem[Mnih et~al.(2015)Mnih, Kavukcuoglu, Silver, Rusu, Veness, Bellemare,
  Graves, Riedmiller, Fidjeland, Ostrovski, et~al.]{mnih2015human}
Volodymyr Mnih, Koray Kavukcuoglu, David Silver, Andrei~A Rusu, Joel Veness,
  Marc~G Bellemare, Alex Graves, Martin Riedmiller, Andreas~K Fidjeland, Georg
  Ostrovski, et~al.
\newblock Human-level control through deep reinforcement learning.
\newblock \emph{Nature}, 518\penalty0 (7540):\penalty0 529--533, 2015.

\bibitem[Munos et~al.(2016)Munos, Stepleton, Harutyunyan, and
  Bellemare]{munos2016safe}
R{\'e}mi Munos, Tom Stepleton, Anna Harutyunyan, and Marc Bellemare.
\newblock Safe and efficient off-policy reinforcement learning.
\newblock In \emph{Advances in Neural Information Processing Systems 29}, pages
  1046--1054, 2016.

\bibitem[O'Donoghue et~al.(2017)O'Donoghue, Osband, Munos, and
  Mnih]{o2017uncertainty}
Brendan O'Donoghue, Ian Osband, Remi Munos, and Volodymyr Mnih.
\newblock The uncertainty {B}ellman equation and exploration.
\newblock In \emph{Proceedings of the 35th Annual International Conference on
  Machine Learning}, 2017.

\bibitem[Ortner and Ryabko(2012)]{Ortner2012}
Ronald Ortner and Daniil Ryabko.
\newblock Online regret bounds for undiscounted continuous reinforcement
  learning.
\newblock In \emph{Advances in Neural Information Processing Systems 25}, pages
  1763--1771, 2012.

\bibitem[Osband(2016)]{osband2016}
Ian Osband.
\newblock \emph{Deep Exploration via Randomized Value Functions}.
\newblock PhD thesis, Stanford University, 2016.

\bibitem[Osband and Van~Roy(2014{\natexlab{a}})]{osband2014model}
Ian Osband and Benjamin Van~Roy.
\newblock Model-based reinforcement learning and the eluder dimension.
\newblock In \emph{Advances in Neural Information Processing Systems 27}, pages
  1466--1474, 2014{\natexlab{a}}.

\bibitem[Osband and Van~Roy(2014{\natexlab{b}})]{osband2014near}
Ian Osband and Benjamin Van~Roy.
\newblock Near-optimal reinforcement learning in factored {MDP}s.
\newblock In \emph{Advances in Neural Information Processing Systems 27}, pages
  604--612, 2014{\natexlab{b}}.

\bibitem[Osband and Van~Roy(2015)]{osband2015bootstrapped}
Ian Osband and Benjamin Van~Roy.
\newblock Bootstrapped {T}hompson sampling and deep exploration.
\newblock \emph{arXiv preprint arXiv:1507.00300}, 2015.

\bibitem[Osband and Van~Roy(2016)]{osband2016lower}
Ian Osband and Benjamin Van~Roy.
\newblock On lower bounds for regret in reinforcement learning.
\newblock \emph{arXiv preprint arXiv:1608.02732}, 2016.

\bibitem[Osband and Van~Roy(2017)]{osband2016posterior}
Ian Osband and Benjamin Van~Roy.
\newblock Why is posterior sampling better than optimism for reinforcement
  learning?
\newblock In \emph{Proceedings of the 34th International Conference on Machine
  Learning}, pages 2701--2710, 2017.

\bibitem[Osband et~al.(2013)Osband, Russo, and Van~Roy]{Osband2013}
Ian Osband, Dnaiel Russo, and Benjamin Van~Roy.
\newblock {(More)} efficient reinforcement learning via posterior sampling.
\newblock In \emph{Advances in Neural Information Processing Systems 26}, pages
  3003--3011. 2013.

\bibitem[Osband et~al.(2016{\natexlab{a}})Osband, Blundell, Pritzel, and
  Van~Roy]{osband2016deep}
Ian Osband, Charles Blundell, Alexander Pritzel, and Benjamin Van~Roy.
\newblock Deep exploration via bootstrapped {DQN}.
\newblock In \emph{Advances In Neural Information Processing Systems 29}, pages
  4026--4034, 2016{\natexlab{a}}.

\bibitem[Osband et~al.(2016{\natexlab{b}})Osband, Van~Roy, and
  Wen]{osband2016rlsvi}
Ian Osband, Benjamin Van~Roy, and Zheng Wen.
\newblock Generalization and exploration via randomized value functions.
\newblock In \emph{Proceedings of The 33rd International Conference on Machine
  Learning}, pages 2377--2386, 2016{\natexlab{b}}.

\bibitem[Osband et~al.(2018)Osband, Aslanides, and Cassirer]{osband2018prior}
Ian Osband, John Aslanides, and Albin Cassirer.
\newblock Randomized prior functions for deep reinforcement learning.
\newblock In S.~Bengio, H.~Wallach, H.~Larochelle, K.~Grauman, N.~Cesa-Bianchi,
  and R.~Garnett, editors, \emph{Advances in Neural Information Processing
  Systems 31}, pages 8625--8637. 2018.

\bibitem[Owen and Eckles(2012)]{owen2012bootstrapping}
Art~B Owen and Dean Eckles.
\newblock Bootstrapping data arrays of arbitrary order.
\newblock \emph{The Annals of Applied Statistics}, pages 895--927, 2012.

\bibitem[Pazis and Parr(2013)]{pazis2013pac}
Jason Pazis and Ronald Parr.
\newblock {PAC} optimal exploration in continuous space {Markov} decision
  processes.
\newblock In \emph{AAAI Conference on Artificial Intelligence}. Citeseer, 2013.

\bibitem[Plappert(2017)]{plappert2017parameter}
Matthias Plappert.
\newblock \emph{Parameter Space Noise for Exploration in Deep Reinforcement
  Learning}.
\newblock PhD thesis, Karlsruhe Institute of Technology, 2017.

\bibitem[Powell and Ryzhov(2011)]{Powell2011}
Warren Powell and Ilya Ryzhov.
\newblock \emph{Optimal Learning}.
\newblock John Wiley and Sons, 2011.

\bibitem[Precup et~al.(2001)Precup, Sutton, and Dasgupta]{precup2001off}
Doina Precup, Richard Sutton, and Sanjoy Dasgupta.
\newblock Off-policy temporal-difference learning with function approximation.
\newblock In \emph{Proceedings of The 18th International Conference on Machine
  Learning}, pages 417--424, 2001.

\bibitem[Russo(2019)]{russo2019worst}
Daniel Russo.
\newblock Worst-case regret bounds for exploration via randomized value
  functions.
\newblock \emph{arXiv preprint arXiv:1906.02870}, 2019.

\bibitem[Russo and Van~Roy(2013)]{Russo2013b}
Daniel Russo and Benjamin Van~Roy.
\newblock Eluder dimension and the sample complexity of optimistic exploration.
\newblock In \emph{Advances in Neural Information Processing Systems 26}, pages
  2256--2264. 2013.

\bibitem[Russo and Van~Roy(2014{\natexlab{a}})]{Russo2014}
Daniel Russo and Benjamin Van~Roy.
\newblock Learning to optimize via posterior sampling.
\newblock \emph{Mathematics of Operations Research}, 39\penalty0 (4):\penalty0
  1221--1243, 2014{\natexlab{a}}.

\bibitem[Russo and Van~Roy(2014{\natexlab{b}})]{russo2014blearning}
Daniel Russo and Benjamin Van~Roy.
\newblock Learning to optimize via information-directed sampling.
\newblock In \emph{Advances in Neural Information Processing Systems 27}, pages
  1583--1591. 2014{\natexlab{b}}.

\bibitem[Russo and Zou(2015)]{russo2015much}
Daniel Russo and James Zou.
\newblock How much does your data exploration overfit? controlling bias via
  information usage.
\newblock \emph{arXiv preprint arXiv:1511.05219}, 2015.

\bibitem[Russo et~al.(2018)Russo, Van~Roy, Kazerouni, Osband, Wen,
  et~al.]{russo2017tutorial}
Daniel~J Russo, Benjamin Van~Roy, Abbas Kazerouni, Ian Osband, Zheng Wen,
  et~al.
\newblock A tutorial on {T}hompson sampling.
\newblock \emph{Foundations and Trends{\textregistered} in Machine Learning},
  11\penalty0 (1):\penalty0 1--96, 2018.

\bibitem[Schaul et~al.(2015)Schaul, Quan, Antonoglou, and
  Silver]{schaul2015prioritized}
Tom Schaul, John Quan, Ioannis Antonoglou, and David Silver.
\newblock Prioritized experience replay.
\newblock \emph{CoRR}, abs/1511.05952, 2015.

\bibitem[Silver et~al.(2016)Silver, Huang, Maddison, Guez, Sifre, Van
  Den~Driessche, Schrittwieser, Antonoglou, Panneershelvam, Lanctot,
  et~al.]{silver2016alphago}
David Silver, Aja Huang, Chris~J Maddison, Arthur Guez, Laurent Sifre, George
  Van Den~Driessche, Julian Schrittwieser, Ioannis Antonoglou, Veda
  Panneershelvam, Marc Lanctot, et~al.
\newblock Mastering the game of go with deep neural networks and tree search.
\newblock \emph{nature}, 529\penalty0 (7587):\penalty0 484--489, 2016.

\bibitem[Silver et~al.(2017)Silver, Schrittwieser, Simonyan, Antonoglou, Huang,
  Guez, Hubert, Baker, Lai, Bolton, et~al.]{silver2017mastering}
David Silver, Julian Schrittwieser, Karen Simonyan, Ioannis Antonoglou, Aja
  Huang, Arthur Guez, Thomas Hubert, Lucas Baker, Matthew Lai, Adrian Bolton,
  et~al.
\newblock Mastering the game of go without human knowledge.
\newblock \emph{Nature}, 550\penalty0 (7676):\penalty0 354, 2017.

\bibitem[Strehl(2007)]{strehl2007probably}
Alexander~L Strehl.
\newblock \emph{Probably approximately correct (PAC) exploration in
  reinforcement learning}.
\newblock PhD thesis, Rutgers University-Graduate School-New Brunswick, 2007.

\bibitem[Strehl et~al.(2006)Strehl, Li, Wiewiora, Langford, and
  Littman]{Strehl2006}
Alexander~L. Strehl, Lihong Li, Eric Wiewiora, John Langford, and Michael~L.
  Littman.
\newblock {PAC} model-free reinforcement learning.
\newblock In \emph{Proceedings of the 23rd Annual International Conference on
  Machine Learning}, pages 881--888, 2006.

\bibitem[Sutton and Barto(2018)]{Sutton2018}
Richard Sutton and Andrew Barto.
\newblock \emph{Reinforcement Learning: An Introduction, Second Edition}.
\newblock MIT Press, 2018.

\bibitem[Sutton et~al.(2009)Sutton, Maei, Precup, Bhatnagar, Silver,
  Szepesv{\'a}ri, and Wiewiora]{sutton2009fast}
Richard Sutton, Hamid~Reza Maei, Doina Precup, Shalabh Bhatnagar, David Silver,
  Csaba Szepesv{\'a}ri, and Eric Wiewiora.
\newblock Fast gradient-descent methods for temporal-difference learning with
  linear function approximation.
\newblock In \emph{Proceedings of the 26th Annual International Conference on
  Machine Learning}, pages 993--1000. ACM, 2009.

\bibitem[Sutton(1988)]{sutton1988learning}
Richard~S Sutton.
\newblock Learning to predict by the methods of temporal differences.
\newblock \emph{Machine learning}, 3\penalty0 (1):\penalty0 9--44, 1988.

\bibitem[Szepesv{\'a}ri(2010)]{2010Szepesvari}
Csaba Szepesv{\'a}ri.
\newblock \emph{Algorithms for Reinforcement Learning}.
\newblock Synthesis Lectures on Artificial Intelligence and Machine Learning.
  Morgan {\&} Claypool Publishers, 2010.

\bibitem[Tang et~al.(2016)Tang, Houthooft, Foote, Stooke, Chen, Duan, Schulman,
  Turck, and Abbeel]{tang2016explore}
Haoran Tang, Rein Houthooft, Davis Foote, Adam Stooke, Xi~Chen, Yan Duan, John
  Schulman, Filip~De Turck, and Pieter Abbeel.
\newblock {\#}{E}xploration: {A} study of count-based exploration for deep
  reinforcement learning.
\newblock \emph{CoRR}, abs/1611.04717, 2016.

\bibitem[Taylor and Stone(2009)]{taylor2009transfer}
Matthew~E Taylor and Peter Stone.
\newblock Transfer learning for reinforcement learning domains: A survey.
\newblock \emph{Journal of Machine Learning Research}, 10\penalty0
  (Jul):\penalty0 1633--1685, 2009.

\bibitem[Tesauro(1995)]{tesauro1995temporal}
Gerald Tesauro.
\newblock Temporal difference learning and {TD}-gammon.
\newblock \emph{Communications of the ACM}, 38\penalty0 (3):\penalty0 58--68,
  1995.

\bibitem[Thompson(1933)]{Thompson1933}
William~R Thompson.
\newblock On the likelihood that one unknown probability exceeds another in
  view of the evidence of two samples.
\newblock \emph{Biometrika}, 25\penalty0 (3/4):\penalty0 285--294, 1933.

\bibitem[Tsitsiklis and Van~Roy(1997)]{tsitsiklis1997analysis}
John~N Tsitsiklis and Benjamin Van~Roy.
\newblock An analysis of temporal-difference learning with function
  approximation.
\newblock \emph{IEEE Transactions on Automatic Control}, 42\penalty0
  (5):\penalty0 674--690, 1997.

\bibitem[Wen(2014)]{wen2014efficient}
Zheng Wen.
\newblock \emph{Efficient reinforcement learning with value function
  generalization}.
\newblock PhD thesis, Stanford University, 2014.

\bibitem[Wen and Van~Roy(2013)]{WenVanroy13}
Zheng Wen and Benjamin Van~Roy.
\newblock Efficient exploration and value function generalization in
  deterministic systems.
\newblock In \emph{Advances in Neural Information Processing Systems 26}, pages
  3021--3029, 2013.

\bibitem[Wierstra et~al.(2008)Wierstra, Schaul, Peters, and
  Schmidhuber]{wierstra2008natural}
Daan Wierstra, Tom Schaul, Jan Peters, and Juergen Schmidhuber.
\newblock Natural evolution strategies.
\newblock In \emph{Evolutionary Computation, 2008. CEC 2008.(IEEE World
  Congress on Computational Intelligence). IEEE Congress on}, pages 3381--3387.
  IEEE, 2008.

\bibitem[Zhang et~al.(2016)Zhang, Bengio, Hardt, Recht, and
  Vinyals]{zhang2016understanding}
Chiyuan Zhang, Samy Bengio, Moritz Hardt, Benjamin Recht, and Oriol Vinyals.
\newblock Understanding deep learning requires rethinking generalization.
\newblock \emph{CoRR}, abs/1611.03530, 2016.

\end{thebibliography}

\end{document}